\pdfoutput=1
\documentclass[11pt]{article}

\usepackage[preprint]{acl}
\usepackage{times}
\usepackage{latexsym}

\usepackage[T1]{fontenc}
\usepackage[utf8]{inputenc}

\usepackage{microtype}

\usepackage{inconsolata}

\usepackage{graphicx}
\usepackage{booktabs} 
\usepackage{hyperref}
\usepackage{array}
\usepackage{amssymb}
\usepackage{multirow}
\usepackage{tikz}
\usepackage{amsthm}
\usepackage{pifont}
\usepackage{tabularray} 
\usepackage{amsmath}
\usepackage[export]{adjustbox}

\newtheorem{definition}{Definition}
\title{Deep Research of Deep Research:\\From Transformer to Agent, From AI to AI for Science}
\author{
	Yipeng Yu\thanks{Email: \texttt{yypzju@163.com}} 
}

\begin{document}
\maketitle

\begin{abstract}
With the advancement of large language models (LLMs) in their knowledge base and reasoning capabilities, their interactive modalities have evolved from pure text to multimodality and further to agentic tool use. Consequently, their applications have broadened from question answering to AI assistants and now to general-purpose agents. Deep research (DR) represents a prototypical vertical application for general-purpose agents, which represents an ideal approach for intelligent information processing and 
%constitutes a significant pathway toward artificial general intelligence (AGI). Centered on LLMs and using tools to interact with the external environment in a multimodal and interactive manner, it 
assisting humans in discovering and solving problems, with the goal of reaching or even surpassing the level of top human scientists. This paper provides a deep research of deep research. We articulate a clear and precise definition of deep research and unify perspectives from industry's deep research and academia's AI for Science (AI4S) within a developmental framework. We position LLMs and Stable Diffusion as the twin pillars of generative AI, and lay out a roadmap evolving from the Transformer to agents. We examine the progress of AI4S across various disciplines. We identify the predominant paradigms of human-AI interaction and prevailing system architectures, and discuss the major challenges and fundamental research issues that remain. AI supports scientific innovation, and science also can contribute to AI growth (Science for AI, S4AI). We hope this paper can help bridge the gap between the AI and AI4S communities.
\end{abstract}
\vspace{0.5em}
\noindent \textbf{Keywords:} Deep Research, AI for Science, AI4S, LLM, Diffusion, Agent, Agentic AI, AI Scientist, Scaling law, GenAI, Generative AI.

\section{Introduction}
\label{sec:introduction}

\begin{quote}
    \textit{``Mind the risks of AI, but fear the halt of its progress more.''}    
    \hfill --- This paper
\end{quote}

Since the emergence of ChatGPT on 30 Nov 2022, nations have gradually become aware of the tremendous advances in AI and have recognized its strategic significance. The European Commission has launched the ``European Strategy for Artificial Intelligence (AI)'' to harness the potential of AI technologies in science and support scientists to adopt them for their research on 8 Oct 2025. The White House of the United States has also launched the ``Genesis Mission'' which aims to win the AI race on 24 Nov 2025. The academic community has increasingly applied LLMs to cutting-edge research areas such as biology~\cite{gao2024empowering,Rao2026}, chemistry \& materials~\cite{tom2024self}, healthcare~\cite{Ong2026}, mathematics~\cite{ju2026aimathematicsprogresschallenges,wang2026horizonmathmeasuringaiprogress}, physics, medicine, meteorology, and other fields~\cite{Gao2024te,ai-science,CHUGUNOVA2026105381}, while industry has begun rolling out next-generation search engines capable of deep research, such as \href{https://gemini.google/overview/deep-research/}{Google DeepMind}, \href{https://openai.com/index/introducing-deep-research/}{OpenAI}, and \href{https://www.perplexity.ai/}{Perplexity}.

Unquestionably, as the capabilities of AI advance and its applications broaden, its integration into the research domain will feature progressively higher levels of automation and intelligence. However, ``deep research'' is a concept that has only emerged within the past two years and currently lacks a unified definition. Its relationship with similar concepts, such as ``deep search'' and ``AI scientist'', also remains ambiguous. Moreover, constrained by disparate resources and environments, industry and academia exhibit divergent motivations, methodologies, and outcomes in their studies of LLMs and deep research. Furthermore, AI researchers often have limited understanding of the research pain points of AI4S researchers, while many AI4S researchers are uncertain about the extent to which AI can contribute to their work. This results in a gap between the two communities.

On the other hand, the majority of current publications on deep research have not undergone rigorous peer review, as many are preprints released on platforms such as arXiv and bioRxiv. Consequently, their quality cannot be guaranteed. Furthermore, the few existing survey papers on the topic often fail to provide a comprehensive overview of deep research. They also lack the clarity and broad applicability necessary to be accessible and useful across different research communities.

In response to these existing issues, we conducted a deep research of deep research. We began by providing a precise definition of the concept and distinguishing it from related notions. Subsequently, we carried out a comprehensive investigation and synthesis of deep research as practiced in both industry and academia. We present the evolution of AI from the Transformer to agents to help AI4S scientists understand core principles. Additionally, we demonstrate how these scientists apply AI within their specific research fields. These practical insights assist AI researchers in refining and iterating deep research agents.

%\section{Related Surveys}
\section{Related Work}
\label{sec:related_surveys}
% consider appendix
As this area is relatively nascent, only a few surveys exist. Wang examined breakthroughs over the past decade that include self-supervised learning and geometric deep learning~\cite{wang2023scientific}. 
Mo conducted a review of conversational search systems which focuses on four modules: query reformulation, search clarification, conversational retrieval, and response generation~\cite{mo2024surveyconversationalsearch}. Lin surveyed the agentic RL foundations of search systems~\cite{lin2025comprehensivesurveyreinforcementlearningbased}. Xi analyzed and categorized the LLM-based search agents from the perspectives of architecture, optimization, application and evaluation~\cite{xi2025surveyllmbaseddeepsearch}. 
Li provided an overview of RL-based agentic search, including methods, evaluation, applications, and challenges~\cite{li2025reinforcementlearningfoundationsdeep}.
Zhang provided a systematic overview of the DR pipeline, which comprises four core stages: planning, question developing, web exploration, and report generation~\cite{zhang2025deepresearchsurveyautonomous}. Ren reviewed the architectures, design, benchmarks, applications, and ethical considerations surrounding LLM-based scientific agents~\cite{ren2025scientificintelligencesurveyllmbased}.
Shi understood DR from three progressive phases (agentic search, integrated search, and AI scientist) and introduced four key components (query planning, information acquisition, memory management, and answer generation)~\cite{Preprints-dr}. Xu defined the scope of DR, and explored the architectural patterns, implementation approaches, and domain-specific adaptations~\cite{xu2025comprehensivesurveydeepresearch}.
Hu reviewed the scientific LLMs from the perspectives of data, model architectures, and agent-based systems~\cite{hu2025surveyscientificlargelanguage}.
Wei offered a domain-oriented review of autonomous scientific discovery across life sciences, chemistry, materials, and physics, synthesizing research progress and advances within each discipline~\cite{wei2025aiscienceagenticscience}. Huang conducted an analysis of the foundational technologies and architectural components that constitute DR agents~\cite{huang2025deepresearchagentssystematic}.

However, existing surveys lack a clear definition of deep research. They often define it narrowly as an agent of search and report generation, and fail to distinguish it from related concepts. Furthermore, they provide little discussion on the collaborative roles of humans and AI, nor do they thoroughly address the gap between industrial applications and academic research. Consequently, many critical questions remain unresolved for the audience. This is especially true for researchers from AI4S. In contrast, our work is not merely another survey. Instead, we offer a deep research of deep research. Specifically, we provide a unified and evolving perspective by comprehensively investigating principles, dataset/benchmarks, models, agents, applications, and challenges. We also articulate promising directions for achieving AGI. Our work can guide and inspire future research for AI and AI4S.

\begin{figure}[t]
  \includegraphics[width=\columnwidth]{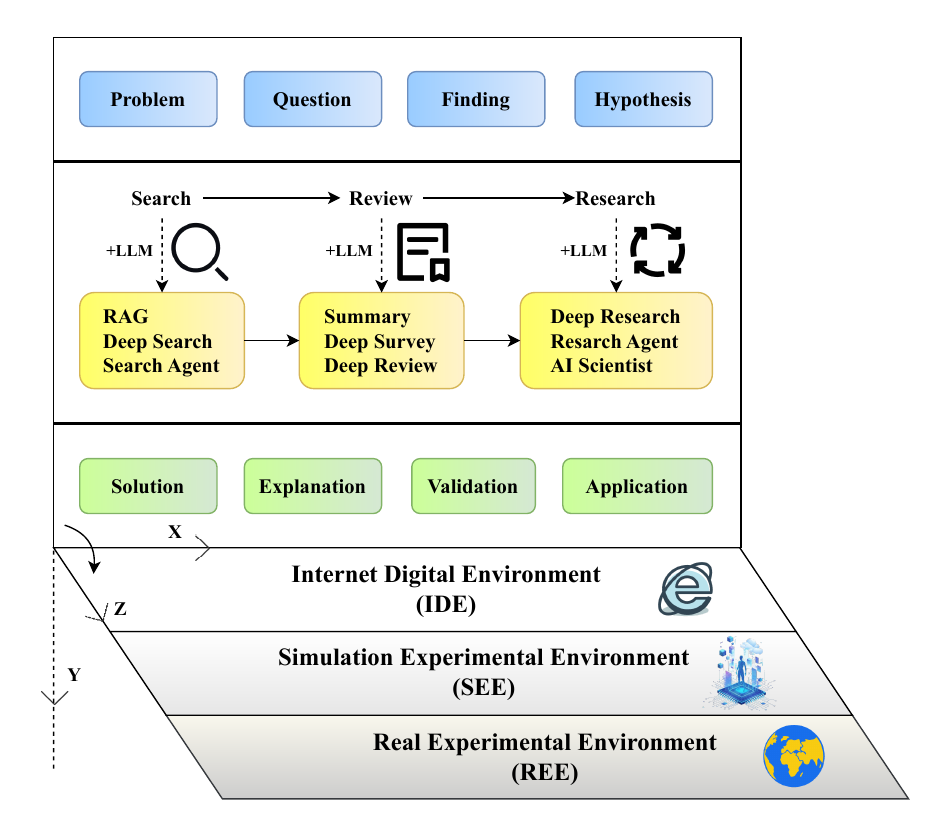}
  \caption{An overview of deep research.}
  \label{fig:overview}
\end{figure}
% insights

\section{Definition and Differences}
% Preliminary
% findings/Discovery, explanation, application
% problems/issues, solutions, Validation
% invention, application

Research is the systematic and diligent inquiry or investigation to discover and interpret facts, generate new knowledge, and gain a deeper understanding of a subject. Research typically comprises basic research and applied research, and it is expected to be reproducible and subjected to peer review. As shown in Figure~\ref{fig:overview}, we provide an overview of deep research from a three-dimensional perspective. The $x$-axis presents the basic stages of research, from search to review and then to research. With the introduction of LLMs, these stages have also been associated with new AI related terms such as ``deep'' and ``agent''. The $y$-axis represents the motivations for research, including the presence of problems, questions, new findings, and hypotheses, as well as the goals, such as providing solutions, answers, validation, and applications. The $z$-axis reflects a gradual progression of experimental settings for research, from the internet digital environment (IDE)  to simulation experimental environment (SEE) and finally to the real experimental environment (REE).
\begin{definition}[\textbf{Deep Research}] Centered on LLM-based AI and using tools to interact with the external environment in a multimodal and interactive manner with feedback, deep research assists humans in discovering and solving problems at different levels of automation, with the goal of reaching or even surpassing the level of top human scientists.
\end{definition}

To clearly delineate the boundaries of deep research, we distinguish it from adjacent terminologies as follows:
\begin{itemize}
    \item Differentiating from Search/RAG: Search constitutes a key step within deep research.
    \item Differentiating from Research: The addition of ``deep'' in deep research implies that the research process becomes more automated, more efficient, and more intelligent.
    \item Differentiating from Deep Review/Survey/Summarization: Review constitutes a key step within deep research.
    \item Differentiating from AI4S / AI for Science: Similar to ``AI assistant'', AI4S emphasizes the use of AI as a tool to support scientific research in various domains.
     \item Differentiating from Vibe Research: Vibe research can be regarded as a stage of deep research. It involves partial automation, but it still requires human intervention.
    \item Differentiating from AI Scientist: AI Scientist is closely related to deep research, but in industry, deep research places greater emphasis on the development of next-generation information processing engines endowed with research capabilities.
\end{itemize}

\section{Foundation: From Transformer to Agent} 
%On the origin of algorithmic progress in AI~\cite{gundlach2025originalgorithmicprogressai}

\subsection{Machine Learning}
AI is a system and a goal. Machine learning (ML)~\cite{mitchell1997machine} is an effective approach to realize AI. Generative AI (GenAI) is an effective route to AI. A generative algorithm is a kind of ML algorithm. From an application standpoint, ML is typically categorized into classification, regression, and clustering. Based on research methodology, the field is further mainly divided into statistical machine learning, deep neural networks~\cite{LeCun2015,Goodfellowetal2016}, reinforcement learning (RL), and evolutionary computation. Based on their approaches to modeling probability distributions, models are also classified as generative or discriminative. According to the use of labeled data, learning is categorized as supervised or unsupervised. Prior to the recent boom in GenAI, ML involved significantly smaller datasets, compute clusters, and model sizes.

\begin{figure*}[t]
	\includegraphics[width=\textwidth]{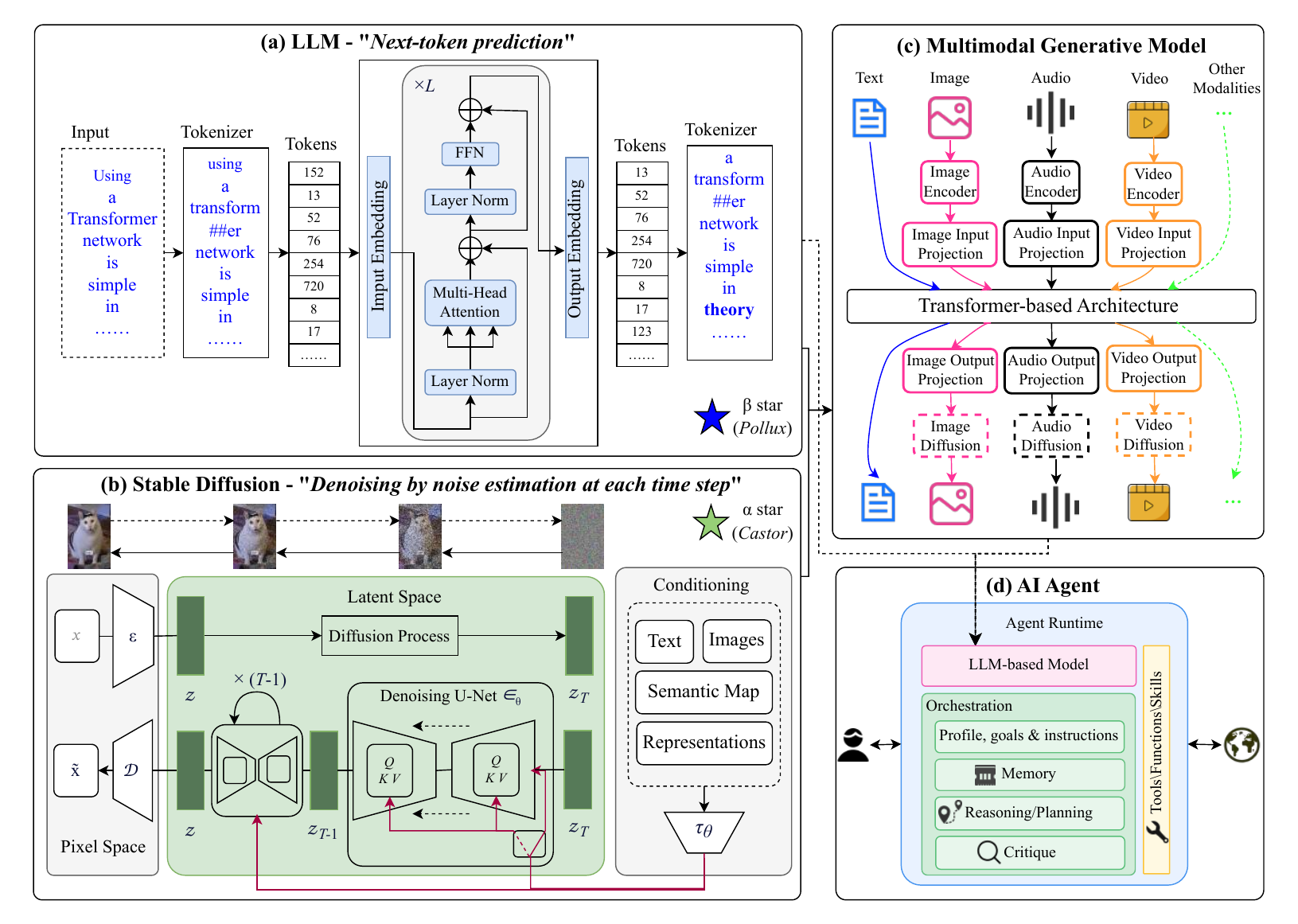}
	\caption{The Gemini of generative AI.}
	\label{fig:gemini}
\end{figure*}
\subsection{Gemini of Generative AI}
As illustrated in Figure~\ref{fig:gemini}, we regard LLMs (\textit{Pollux}) and Stable Diffusion (\textit{Castor}) as the Gemini of GenAI.
\subsubsection{\textit{Pollux}: LLM}
The time from the adoption of deep neural network for natural language processing (NLP) to the emergence of LLMs is actually only about ten years. Word2Vec introduced efficient neural methods to learn dense distributed word representations that capture semantic and syntactic relationships~\cite{word2veciclr13,nipsword2vec13}. Better subword tokenization techniques like BPE~\cite{sennrich-etal-2016-neural, kudo-2018-subword}, WordPiece and SentencePiece, further help to reduce vocabulary size and handle rare words by encoding text into subword units (tokens). Subsequently, the Transformer model was proposed~\cite{NIPS2017_3f5ee243}. Transformers work because they use self-attention to dynamically weigh the importance of different words in a sequence, enabling parallel processing and capturing long-range dependencies more effectively than recurrent or convolutional models. Following this, different architectures based on the Transformer demonstrated superior performance on NLP tasks. These included the decoder-only GPT-1~\cite{radford2018improving}, the encoder-only BERT~\cite{devlin-etal-2019-bert}, and the encoder-decoder T5~\cite{t520}. OpenAI continued to refine the decoder-only models (see Figure~\ref{fig:gemini}(a)) and published GPT-2~\cite{radford2019language} and GPT-3~\cite{brown2020language}. Note that GPT-3 ushered in the era of prompt engineering. In 2022, OpenAI released ChatGPT based on GPT-3.5. This system marked the first human-like conversation and attracted widespread public attention. In the next year, Meta's open-weight model Llama 2 accelerated the trend of LLMs moving from ``close'' to ``open''~\cite{touvron2023llama2openfoundation}. Later, more effective positional embedding approaches were proposed to integrate the positional information among tokens~\cite{RoFormer24,dape24,liu-etal-2025-vrope},  LoRA~\cite{hu2022lora} and RL~\cite{gptrlhf,JMLRDLRL} were used to fine-tune LLMs. It can be argued that LLMs are a success of brute force in computation. They demonstrate that larger models, more data, and stronger computational resources can lead to a qualitative improvement in neural network performance. This established the foundation for the later idea summarized as ``scaling law''~\cite{kaplan2020scalinglawsneurallanguage,yan2026what}, ``intelligence emerging''~\cite{wei2022emergent}, ``grokking''~\cite{power2022grokking}, ``LLM is intelligence compression''~\cite{deletang2024language}, or ``Aha moment''.
%positional embedding~\cite{liu-etal-2025-vrope,ge2025v2pe,dape24}, 
Above all, next-token prediction (NTP) in the pretraining is the core and foundational training paradigm for autoregressive language modeling, its training objective can be formulated as follows:
\begin{align}
	\mathcal{L}(\theta) = -\sum_{t=1}^{T-1} \log P_\theta(x_{t+1} \mid x_1, x_2, \dots, x_t)
\end{align}
$P_\theta(x_{t+1}\mid x_{1:t})$ is the probability assigned by the model (parameterized by $\theta$) to the true next token $x_{t+1}$, given the context $x_{1:t}$. The sum runs from $t=1$ to $T-1$, since there is no ``next token'' after the final token.

\subsubsection{\textit{Castor}: Stable Diffusion}
A diffusion probabilistic model is a parameterized Markov chain trained using variational inference to produce samples matching the data after finite time~\cite{ddpm20}. The forward process (diffusion/noising) gradually adds Gaussian noise to an image over many steps until it becomes pure noise, while the reverse process (denoising/generation) involves a neural network (U-Net~\cite{ronneberger2015u}) learning to predict and subtract that noise step-by-step, allowing it to generate new, realistic data from random noise by reversing the corruption. The U-Net learns to reverse the noising, taking a noisy image and a timestep, and predicting the noise component to subtract, effectively moving from \(x_{t}\) to \(x_{t-1}\). Stable Diffusion (SD)~\cite{Rombach_2022_CVPR} performs these processes not on pixel data, but in a compressed ``latent space'' using an autoencoder. This makes the process much faster and less computationally intensive than applying diffusion directly to high-resolution images. Moreover, by introducing cross-attention layers into the model architecture, SD turns diffusion models into powerful and flexible generators for general conditioning inputs (see Figure~\ref{fig:gemini}(b)). In the same year, \href{https://stability.ai/}{Stability AI} and \href{midjourney.com}{Midjourney} released their image generation products and sparked a wave of mass participation in image creation. The corresponding objective can be simplified as follows:
\begin{align}
	L_{\mathrm{SD}}
	&:= \mathbb{E}_{\mathcal{E}(x),\, y,\, \epsilon \sim \mathcal{N}(0,1),\, t}
	\left[ \left\| \epsilon - \epsilon_\theta(z_t, t, \tau_\theta(y)) \right\|_2^2 \right]
\end{align}
The noise is drawn from a standard normal distribution $\epsilon\sim\mathcal{N}(0,1)$, the encoder $\mathcal{E}$ encodes an image $x$ into a latent representation $z$, the encoder $\tau_\theta$ projects the condition $y$ to an intermediate representation $\tau_\theta(y))$, the denoising U-Net $\epsilon_\theta$ estimates the noise at each time step. Later, the Transformer replaced U-Net~\cite{peebles2023scalable}, and Rectified flow~\cite{sd324}, a new generative model formula for finding a transport map between two empirically observed distributions by learning an ordinary differential equation, demonstrated superior performance compared to the diffusion formulations. Thus flow and diffusion architectures based on the Transformer became the dominant approach for image generation~\cite{ma2024sit}. Concurrently, ControlNet~\cite{zhang2023adding} and InstantID~\cite{wang2024instantid} were proposed to provide finer grained control over image generation using conditional inputs, and AnimateDiff~\cite{guo2023animatediff} were developed to generate temporally consistent images, enabling video synthesis.

\begin{table*}[!ht]
	\centering
	\begin{tblr}
		{
			colspec={X[c]X[c]X[c]X[c]X[c]X[c]}, 
			colsep = 1mm,
			row{1,2} = {font=\bfseries},
			rows={valign=m},
			hline{2} = {2-4}{solid},
			column{1} = {wd=0.9cm},
			column{2} = {wd=2.6cm},
			column{3} = {wd=3.84cm},
			column{5} = {wd=2.2cm},
			column{6} = {wd=3.35cm},
			row{3-Z} = {halign=l},
			hline{1}  = {1pt},  
			hline{3}  = {1pt},  
			hline{Z}  = {1pt},
			rowsep = 0.5pt,
			stretch = 1.0
		}
		\SetCell[r=2]{c} Year & \SetCell[c=3]{c} GPU & & & \SetCell[r=2]{c} Framework & \SetCell[r=2]{c} Application  \\
		& Model & Hardware & Software & & \\
		2007 & G80, Tesla C870 &  & CUDA & Theano &  \\
		\hline
		\textcolor{red}{2012} & GTX 580 &  &  &  & \textcolor{red}{AlexNet} \\
		\hline
		2013 &  & &  & Caffe & Word2Vec \\
		\hline
		2014 & GTX 980, 1080 &  & cuDNN &  & GAN \\
		\hline
		2015 &  &  &  & Keras, Tensorflow & ResNet \\
		\hline
		2016 & Tesla P100 & Double/single-precision, HBM2, NVLink & TensorRT &
		\textcolor{red}{PyTorch}, MXNet & AlphaGo \\
		\hline
		2017 & Tesla V100 &
		Mixed-precision training (FP16),
		Tensor Cores &  &  & Transformer \\
		\hline
		2018 &  &  &  &  & BERT, GPT-1 \\
		\hline
		2019 &  &  &  & Megatron-LM & GPT-2 \\
		\hline
		2020 & \textcolor{red}{A100} &
		TF32, BF16, 40/80GB HBM,
		Sparse Computing &  &
		DeepSpeed & GPT-3 \\
		\hline
		\textcolor{red}{2022} & H100 &
		FP8, Transformer Engine, DPX &
		Triton &  &
		\textcolor{red}{ChatGPT (GPT-3.5), Stable Diffusion} \\
	\end{tblr}
	\caption{The GPU-driven golden decade of AI development from 2012 to 2022.}
	\label{tab:gpu}
\end{table*}

\subsection{Multimodal Generative Model}
As shown in Figure~\ref{fig:gemini}(c), multimodal generative models aim to integrate autoregressive language models and diffusion/flow models into a single framework, thereby extending model capabilities from a single modality to multiple modalities~\cite{Chen2024MultiModalGA}. One approach is to unify understanding and generation across multiple modalities within the same NTP used by LLMs~\cite{wu2025janus,chen2025janusprounifiedmultimodalunderstanding,Wang2026}, another approach cascades external diffusion models after the output of an MLLM to generate visual and audio modalities~\cite{wang2024emu3,ge2025seedxmultimodalmodelsunified}, and a third approach combines next-token prediciton with mask token prediction~\cite{xie2025showo} or rectified flow~\cite{ma2025janusflow} in one single LLM that can handle different modalities in distinct ways. The exploration of multimodal generative models is still in its early research stage. The leading products right now are Google's \href{https://gemini.google/overview/image-generation/}{Nano Banana} and Bytedance's \href{https://seed.bytedance.com/en/seedance2_0}{Seedance2.0}.
%block diffusion~\cite{arriola2025block}
%~\cite{gu2026thinkmorph}

\subsection{Agent}
\begin{quote}
	\textit{``Token is cheap, show me your agent.''}    
	\hfill --- This paper
\end{quote}

The term ``agent'' is not new. In earlier work, it typically referred to humans~\cite{Jiang_Ma_Lu_Yu_Yu_Li_2019,yu-etal-2020-conversation}. After the emergence of ChatGPT, it has attracted renewed attention. Its meaning has also shifted from referring to humans to referring to AI. With the discovery of ``Test-time Scaling''~\cite{wang2023selfconsistency,lightman2023letsverifystepstep} and advances in interaction methods and reasoning capabilities of LLMs~\cite{cot22,tot23,yao2023reactsynergizingreasoningacting,openai2024openaio1card,Guo2025}, researchers have begun integrating LLMs as the cognitive core into existing agents, while also equipping LLMs with tools, memory, and feedback~\cite{nakano2022webgpt,toolformer23,reflexion23}. Real-time and factual information from tools can help mitigate the hallucination issue and transcend the NTP paradigm in LLMs. LLMs are static, stateless, and passive, whereas agents are dynamic, stateful, and proactive. After pretraining, fine-tuning, or prompt engineering, an LLM can generate tokens that specify tool invocation. The interaction between an agent and an LLM typically carries out the ``Thought$\rightarrow$Action$\rightarrow$Observation'' loop. First, a user submits a query and the agent calls the LLM to obtain an initial token sequence. Second, if the agent detects that the output requires a tool call, it executes the tool, obtains the result, concatenates the result with the context, and calls the LLM again. Third, the agent repeats these steps until a termination condition is met, and then returns the answer to the user. The core modules of an agent are typically shown in Figure~\ref{fig:gemini}(d). Different agents feature distinct orchestration structures and interact with the LLM through different processes. Notable open-source agent frameworks include \textit{\href{https://github.com/langchain-ai/langchain}{LangChain}}, \textit{\href{https://github.com/significant-gravitas/autogpt}{AutoGPT}}, \textit{\href{https://github.com/run-llama/llama_index}{LlamaIndex}}, \textit{\href{https://github.com/microsoft/autogen}{AutoGen}}, \textit{\href{https://github.com/FoundationAgents/MetaGPT}{MetaGPT}}, \href{https://github.com/openclaw/openclaw}{\textit{OpenClaw}}, and \textit{\href{https://github.com/crewaiinc/crewai}{CrewAI}}, and popular applications mainly include the coding agent \textit{\href{https://cursor.com/}{Cursor}}, the search agent \textit{\href{https://www.perplexity.ai/}{Perplexity}}, and deep research agents introduced in this paper.

\subsection{GPU}
As shown in Table~\ref{tab:gpu}, the development of AI has relied heavily on advances in NVIDIA's GPUs and their associated software infrastructure. The introduction of CUDA in 2007 enabled GPU computing power to be used not only for graphics rendering but also for general-purpose computation. In the same year, Theano laid the groundwork for modern deep learning frameworks. However, it was the AlexNet work using only \textbf{two} GTX 580 GPUs in 2012 that truly established GPUs as essential hardware for deep learning, triggering what is often called the ``Cambrian explosion'' of AI. This paper refers to the period from 2012, when AlexNet was published, to 2022, when ChatGPT was released, as the golden decade of AI. During this time, GPUs evolved rapidly. CUDA core counts continued to increase. Memory capacity grew larger, and memory bandwidth became higher. As a result, two key metrics, compute density (FLOPs) and interconnect bandwidth, improved significantly. The application domain of GPUs expanded from consumer gaming to AI data centers. Concurrently, deep learning frameworks matured, giving rise to two dominant platforms: TensorFlow and PyTorch. Among the GPU generations, the NVIDIA A100 made large-scale distributed training of large models practically feasible. ChatGPT can be regarded as a product of GPT-3.5 trained on A100/V100 GPUs with PyTorch. Leading LLMs today are typically trained on GPU clusters with more than \textbf{10,000} GPUs.
%model on GPU \href{https://chatjimmy.ai/}{Taalas}

\begin{figure*}[!th]
	\includegraphics[width=\textwidth]{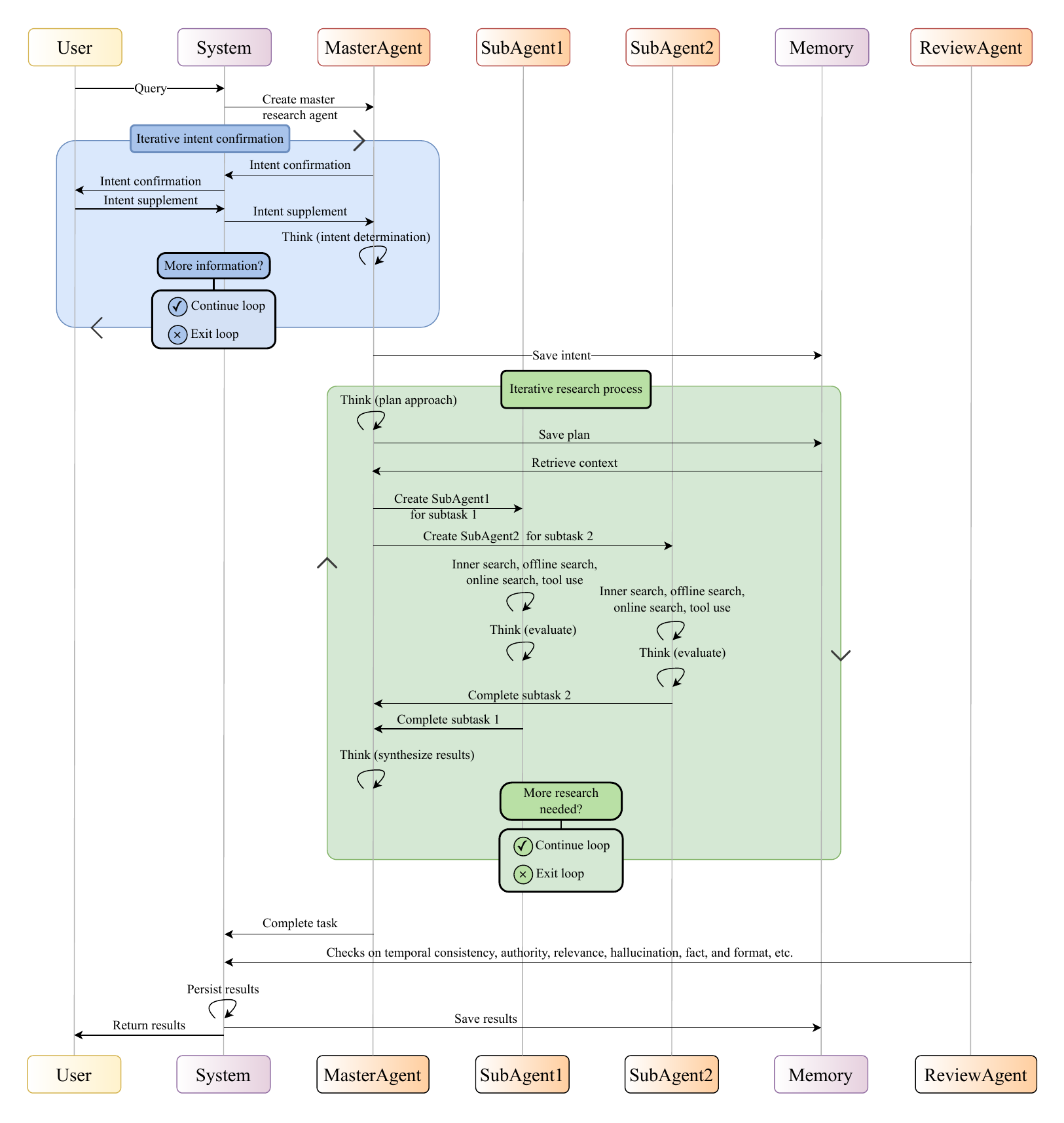}
	\caption{Iterative deep research.}
	\label{fig:itr-dr}
\end{figure*}
\section{AI Perspective}
\begin{quote}
	\textit{``Train AI for the real world, not just for the leaderboard.''}
	\hfill --- This paper
\end{quote}
This section investigates deep research from an AI perspective, focusing on why and how AI developers design and build such agents based on LLMs. The motivation of  DR is to automate complex, multi-step research by using LLMs to plan, search the web, analyze hundreds of sources, and synthesize information into detailed, cited reports, drastically cutting down research time from days to minutes for tasks like market analysis and legal reviews. It goes beyond simple Q\&A by tackling intricate queries that require reasoning and gathering data from vast online sources, providing actionable insights and plans. 
%DRs tackle complex, multi-step queries by autonomously planning, searching hundreds of sources, analyzing, and synthesizing information into detailed, cited reports, saving users significant time compared to manual research, making it ideal for in-depth market analysis, scientific queries, and others, functioning like an automated research analyst for Pro users and developers. 
%more intelligent information processing

\begin{table*}[!th]
	\centering
	\begin{tblr}
		{
			colspec={Q[2.05cm,l]Q[2.42cm,l]X[c]X[c]X[c]X[c]X[c]X[c]X[c]}, 
			colsep = 1mm,
			row{1,2} = {font=\bfseries},
			rows={valign=m},
			hline{2} = {5-7}{solid},
			rowsep = 0.5pt,
			stretch = 1.0
		}
		\hline
		\SetCell[r=2]{c} DR Agent & \SetCell[r=2]{c} Backbone  & \SetCell[r=2]{c} Logo &\SetCell[r=2]{c} Open or Closed & \SetCell[c=3]{c}Capabilities & & & \SetCell[r=2]{c} ENV & \SetCell[r=2]{c} Autonomy  \\
		& & & & Search & Review & Research & & \\
		\hline                  
		\href{https://gemini.google.com/app?utm_source=sem&utm_source=google&utm_medium=paid-media&utm_medium=cpc&utm_campaign=deepresearch_bkws&utm_campaign=2024enUS_gemfeb&gad_source=1&gad_campaignid=22908443171&gbraid=0AAAAApk5BhkgBSBInl-CMsKwqJQ2jak5f&eps=AHas8cB8iqQyQW-GZIpphZXOVyKq-hl7fufLyBWEJVjiVz-9u77INPBb7v-lUGqEg0tFWwlcZjCwmUFmAObfLhWkfGLlqnTiLd725uYHNTdQv7_EC7_V&gclid=Cj0KCQiAxJXJBhD_ARIsAH_JGjj2MP9D874nmHlsH3NlnnEZUge3y5oxwp7bys4RjHhtAq12fh4GoHMaAm80EALw_wcB&gclsrc=aw.ds}{Gemini} &Gemini &\includegraphics[width=0.45cm, valign=m]{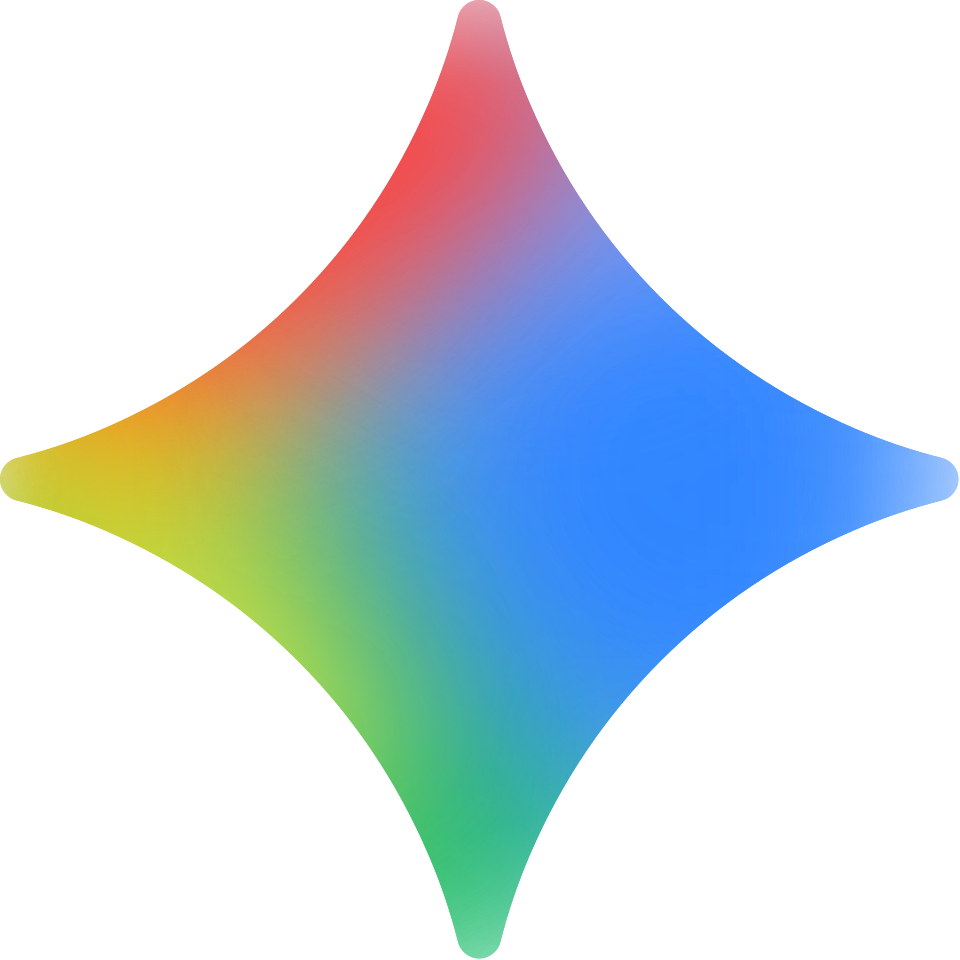} & Closed & $\checkmark$ & $\checkmark$ & \ding{73}\ding{73}\ding{72}\ding{72}\ding{72} & IDE & $\circ\circ\bullet\bullet\bullet$ \\
		\href{https://openai.com/zh-Hans-CN/}{ChatGPT}  &GPT & \includegraphics[width=0.55cm, valign=m]{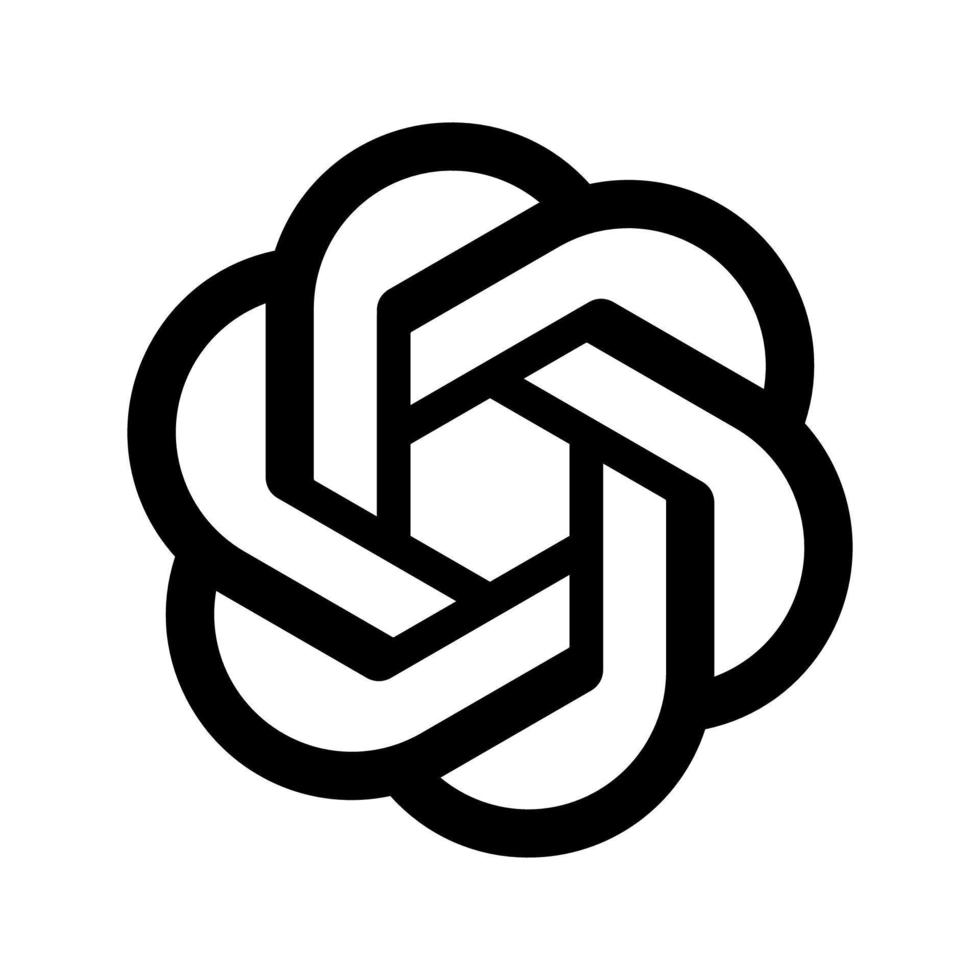}& Closed & $\checkmark$ & $\checkmark$ & \ding{73}\ding{73}\ding{72}\ding{72}\ding{72} & IDE & $\circ\circ\bullet\bullet\bullet$ \\
		\href{https://claude.com/product/overview}{Claude} &Claude&\includegraphics[width=0.4cm, valign=m]{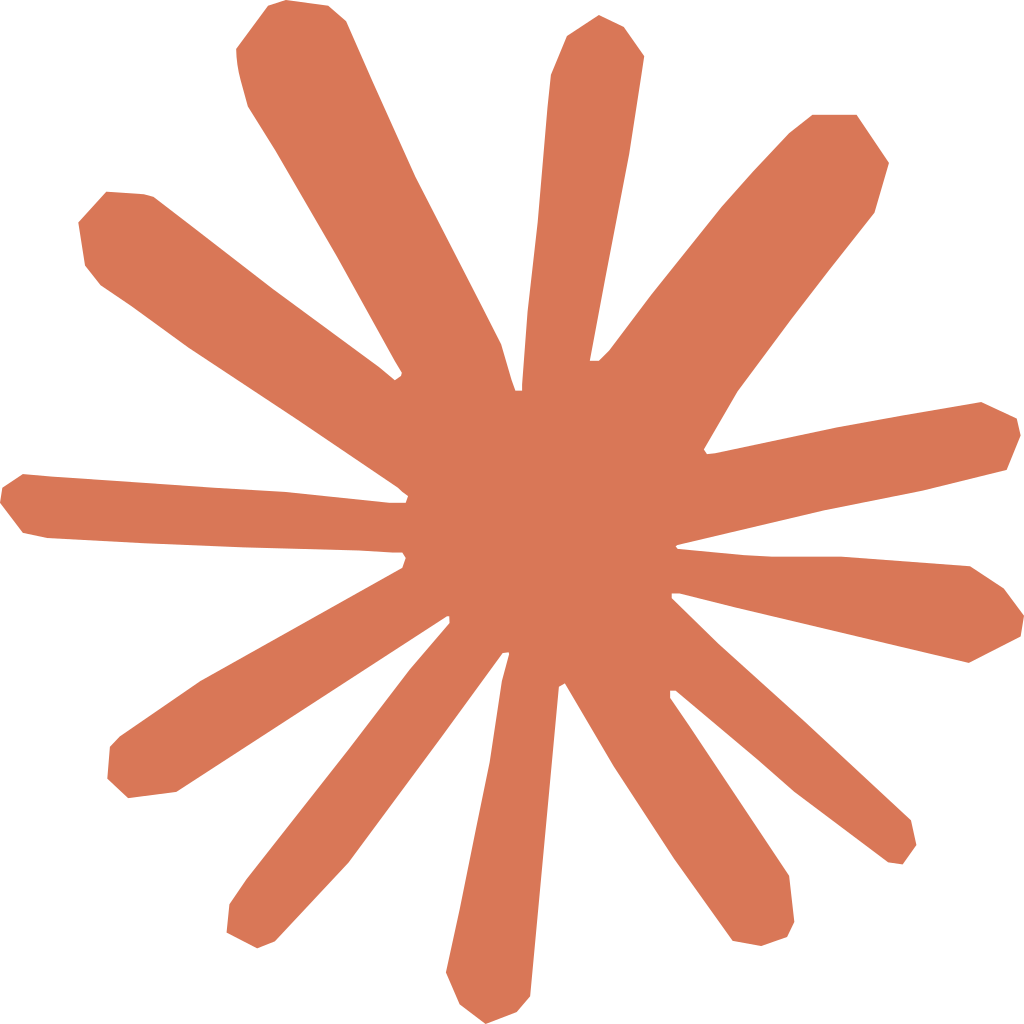}&Closed& $\checkmark$& $\checkmark$&\ding{73}\ding{73}\ding{72}\ding{72}\ding{72}&IDE&$\circ\circ\bullet\bullet\bullet$\\
		\href{https://grok.com/}{Grok} &Grok&\includegraphics[width=0.45cm, valign=m]{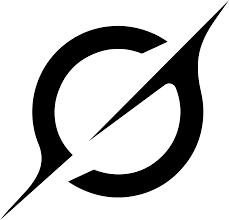}& Closed & $\checkmark$ & $\checkmark$ & \ding{73}\ding{73}\ding{73}\ding{72}\ding{72} & IDE & $\circ\circ\bullet\bullet\bullet$ \\
		\href{https://www.kimi.com/researcher}{Kimi} &Kimi&\includegraphics[width=0.45cm, valign=m]{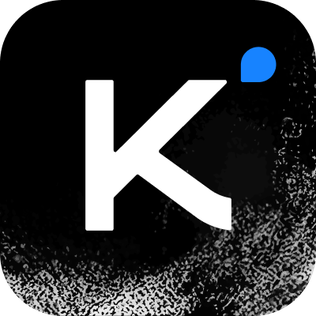}& Closed & $\checkmark$&$\checkmark$&\ding{73}\ding{73}\ding{73}\ding{72}\ding{72}&IDE&$\circ\circ\bullet\bullet\bullet$\\
		\href{https://www.doubao.com/}{Doubao} &Seed&\includegraphics[width=0.6cm, valign=m]{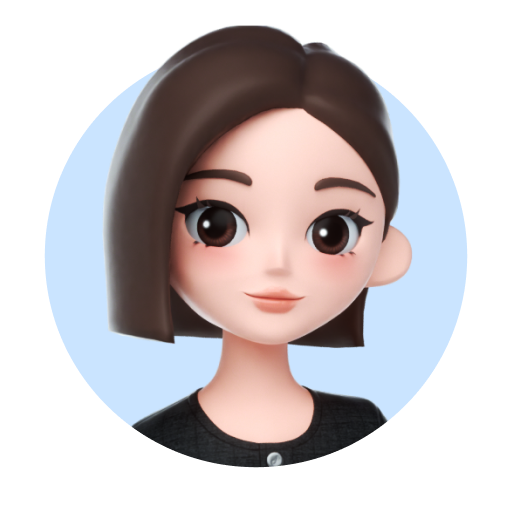}&Closed& $\checkmark$& $\checkmark$&\ding{73}\ding{73}\ding{73}\ding{72}\ding{72}&IDE&$\circ\circ\bullet\bullet\bullet$\\
		\href{https://agent.minimaxi.com/}{MiniMax} &MiniMax&\includegraphics[width=0.45cm, valign=m]{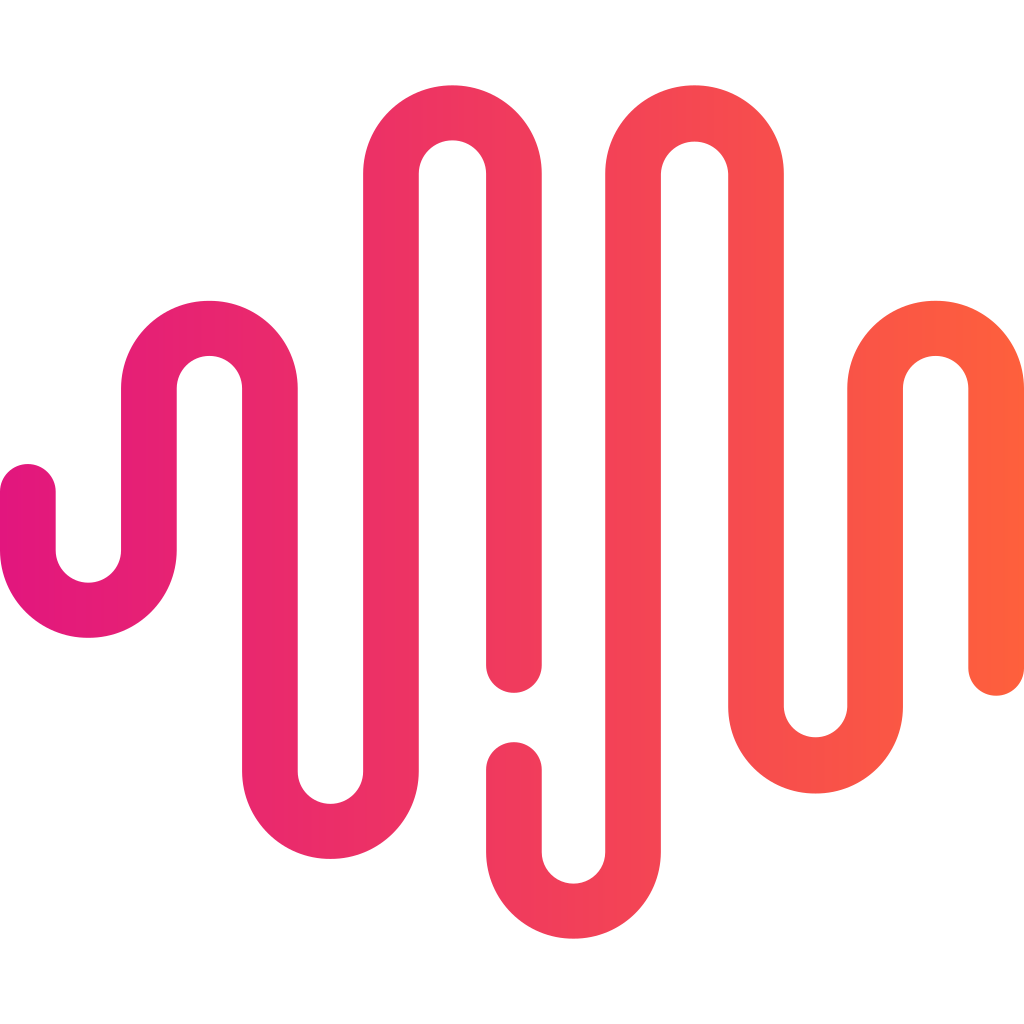}&Closed& $\checkmark$& $\checkmark$&\ding{73}\ding{73}\ding{73}\ding{72}\ding{72}&IDE&$\circ\circ\bullet\bullet\bullet$\\
		\href{https://ernie.baidu.com/}{Ernie} & Ernie &\includegraphics[width=0.45cm, valign=m]{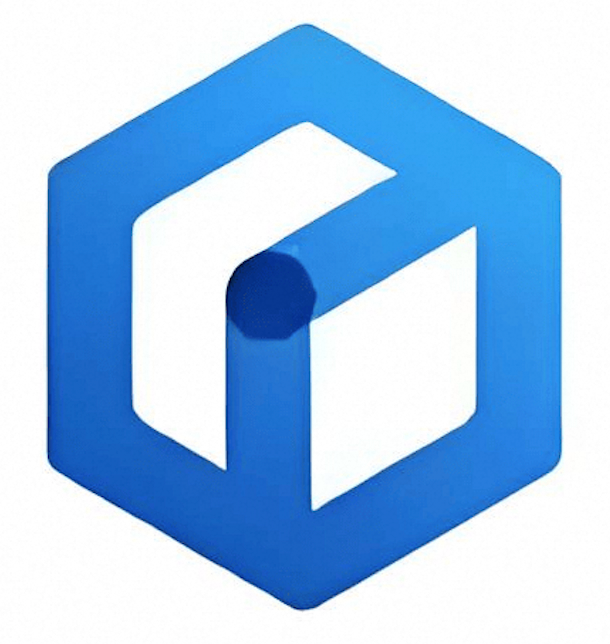}&Closed& $\checkmark$& $\checkmark$&\ding{73}\ding{73}\ding{73}\ding{72}\ding{72}&IDE&$\circ\circ\bullet\bullet\bullet$\\
		\href{https://www.stepfun.com/chats/new}{StepFun} &Step&\includegraphics[width=0.45cm, valign=m]{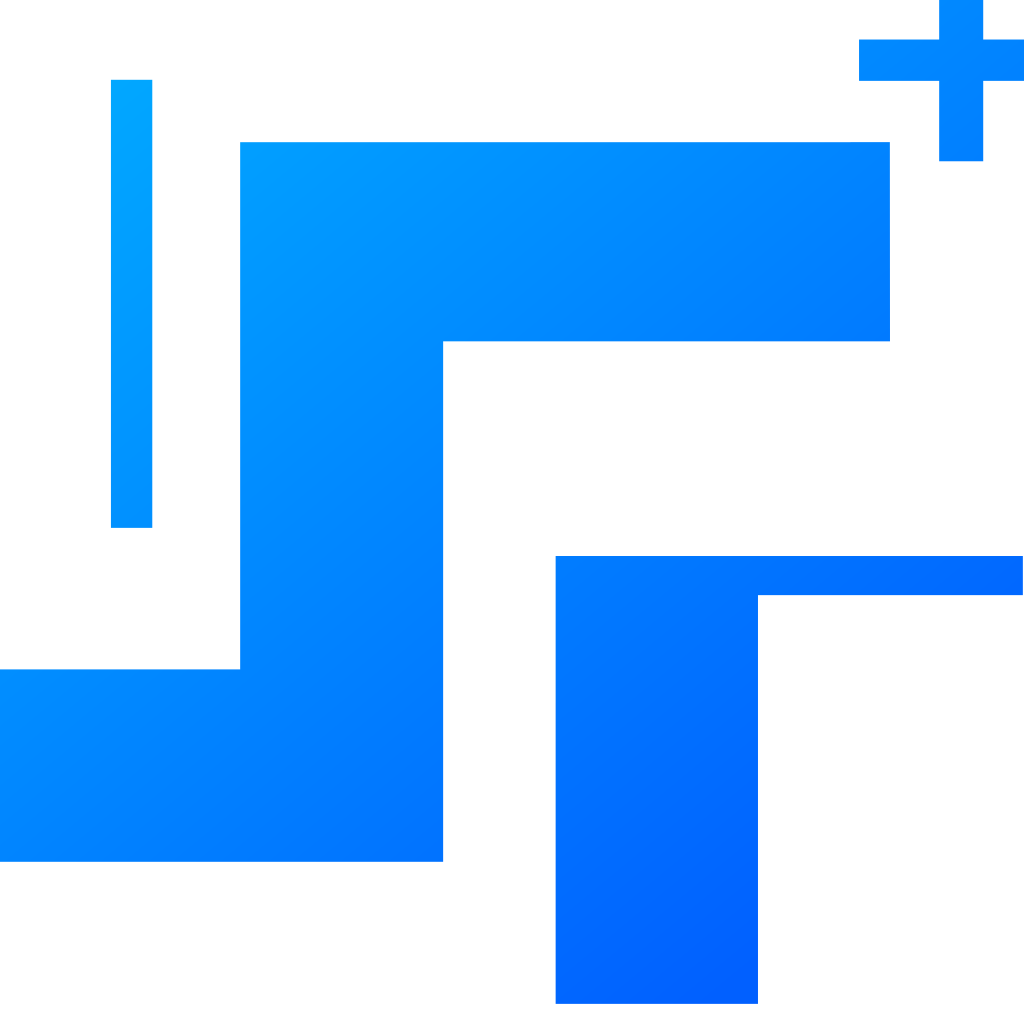}&Closed& $\checkmark$& $\checkmark$&\ding{73}\ding{73}\ding{73}\ding{72}\ding{72}&IDE&$\circ\circ\bullet\bullet\bullet$\\
		\hline[dashed]
		%		ERNIE &Close& $\checkmark$& $\checkmark$&\ding{73}\ding{73}\ding{73}\ding{72}\ding{72}&IDE&$\bullet\circ\circ$\\
		\href{https://chat.qwen.ai/?inputFeature=deep_research}{Qwen}&Qwen&\includegraphics[width=0.5cm, valign=m]{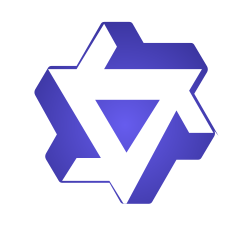} &\href{https://github.com/Alibaba-NLP/DeepResearch}{Open}& $\checkmark$& $\checkmark$&\ding{73}\ding{73}\ding{72}\ding{72}\ding{72}&IDE&$\circ\circ\bullet\bullet\bullet$\\
		\href{https://chat.deepseek.com/}{DeepSeek}&DeepSeek&\includegraphics[width=0.55cm, valign=m]{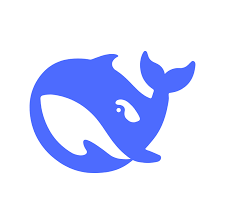} &\href{https://github.com/deepseek-ai}{Open}& $\checkmark$& $\checkmark$&\ding{73}\ding{73}\ding{73}\ding{72}\ding{72}&IDE&$\circ\circ\bullet\bullet\bullet$\\
		\href{https://chat.z.ai/}{GLM}&GLM &\includegraphics[width=0.35cm, valign=m]{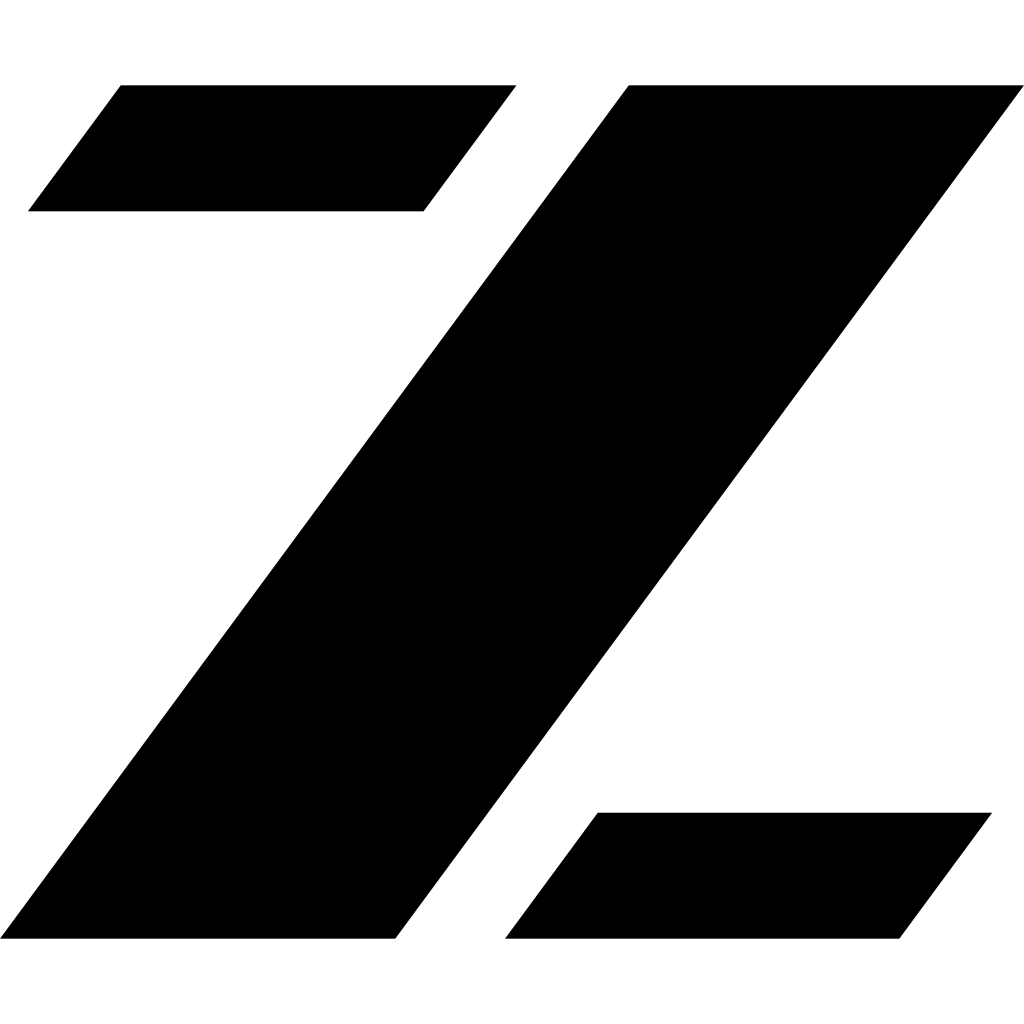} &\href{https://github.com/zai-org/GLM-4.5}{Open}& $\checkmark$& $\checkmark$&\ding{73}\ding{73}\ding{73}\ding{72}\ding{72}&IDE&$\circ\circ\bullet\bullet\bullet$\\
		\hline[dashed]
		\href{https://www.perplexity.ai/}{Perplexity} &DeepSeek & \includegraphics[width=0.45cm, valign=m]{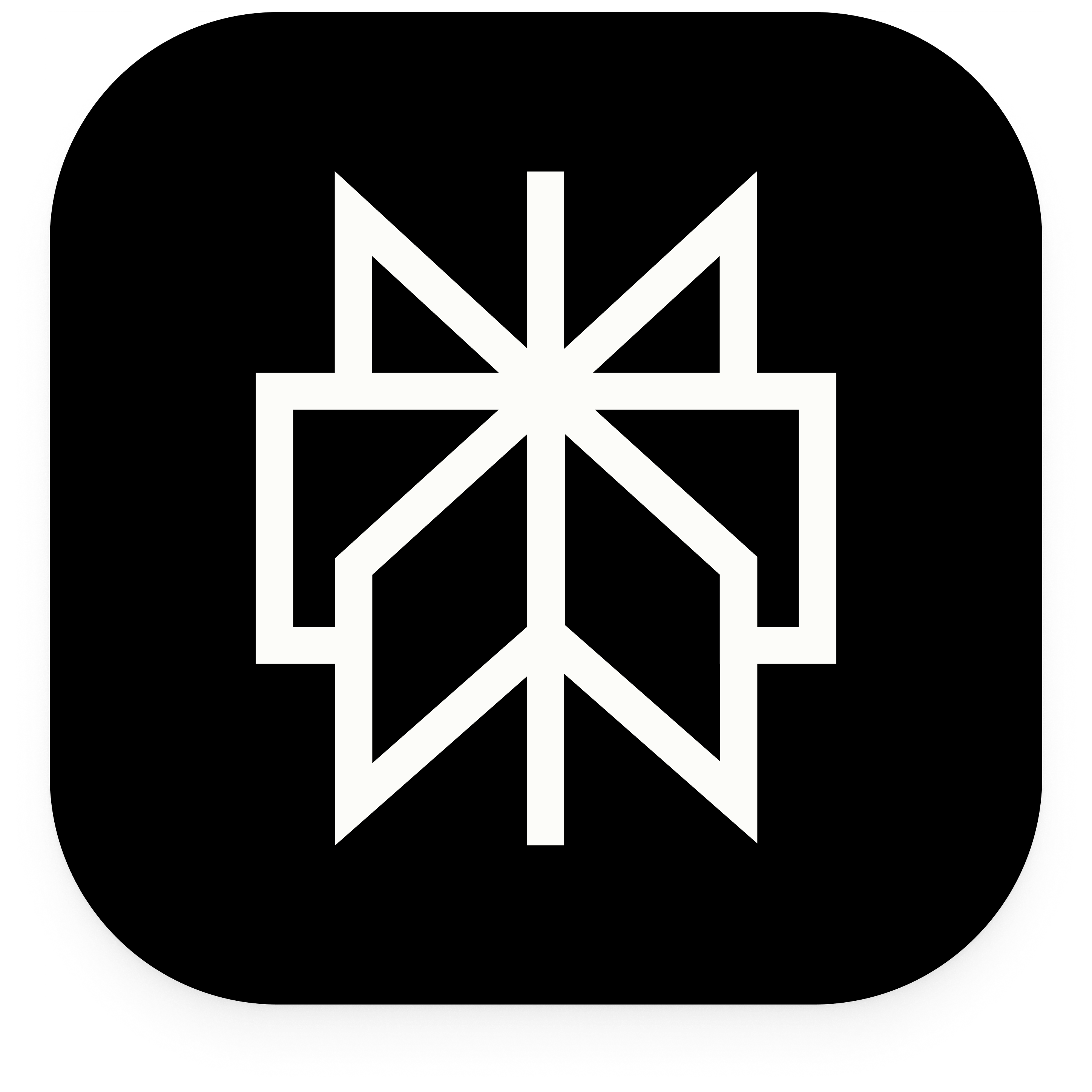}& Closed & $\checkmark$ & $\checkmark$ & \ding{73}\ding{73}\ding{72}\ding{72}\ding{72} & IDE & $\circ\circ\bullet\bullet\bullet$ \\
		\href{https://dr.miromind.ai/}{MiroThinker}& Qwen & \includegraphics[width=0.45cm, valign=m]{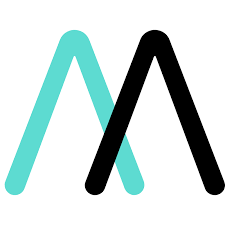} &\href{https://github.com/MiroMindAI/MiroThinker}{Open}& $\checkmark$& $\checkmark$&\ding{73}\ding{73}\ding{73}\ding{72}\ding{72}&IDE&$\circ\circ\bullet\bullet\bullet$\\
		\href{https://scimaster.bohrium.com/}{SciMaster}& DeepSeek & \includegraphics[width=0.45cm, valign=m]{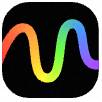} &\href{https://github.com/sjtu-sai-agents/X-Master}{Open}& $\checkmark$& $\checkmark$&\ding{73}\ding{73}\ding{73}\ding{72}\ding{72}&IDE&$\circ\circ\bullet\bullet\bullet$\\
		\href{https://deerflow.net/chat}{DeerFlow}& Model-agnostic & \includegraphics[width=0.45cm, valign=m]{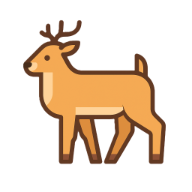} &\href{https://github.com/bytedance/deer-flow}{Open}& $\checkmark$& $\checkmark$&\ding{73}\ding{73}\ding{73}\ding{72}\ding{72}&IDE&$\circ\circ\bullet\bullet\bullet$\\

		\hline
	\end{tblr}
	\caption{Pioneering deep research agents (proprietary closed, proprietary open, external base) from the AI perspective. The five levels of automation correspond to L1-L5 in Figure~\ref{fig:paradigm}. Note that the performance of these agents varies over time. AI4S researchers may also consider conducting studies based on open-weight models such as Llama 3~\cite{grattafiori2024llama3herdmodels}, MiMO~\cite{coreteam2026mimov2flashtechnicalreport}, Mistral~\cite{liu2026ministral3}, and LongCat~\cite{longcatflashthinking2601technicalreport}.}
	\label{tab:industry}
\end{table*}
%Mistral, Llama, MiMo, LongCat

\subsection{Benchmark}
\label{sec:bench}
A benchmark is typically built on one or more datasets. It defines evaluation rubrics, metrics, parameters, and procedures to assess and rank agent performance. Benchmarks can be either public or internally proprietary. The datasets used in the public benchmarks may be fully public, have hidden test sets, or be entirely private. Agent performance on these benchmarks can guide version iteration, regression testing, and deployment decisions. Common public benchmarks for deep research agents include 
GAIA~\cite{mialon2023gaia}, 
GPQA~\cite{rein2024gpqa},
FRAMES~\cite{krishna2025factfetchreasonunified},
BrowseComp~\cite{wei2025browsecompsimplechallengingbenchmark,zhou2025browsecompzhbenchmarkingwebbrowsing}, WebWalkerQA~\cite{wu2025webwalker},
\href{https://github.com/youdotcom-oss/ydc-deep-research-evals}{DeepConsult}, DeepResearchGym~\cite{coelho2025deepresearchgymfreetransparentreproducible}, xbench-DeepSearch~\cite{chen2025xbenchtrackingagentsproductivity},
DeepResearch Bench~\cite{du2025deepresearchbenchcomprehensivebenchmark}, ScholarQABench~\cite{Asai2026-mp},
and Humanity’s Last Exam~\cite{phan2025humanitysexam}. 
Newly proposed public benchmarks that still require validation include
SuperCLUE~\cite{xu2023supercluecomprehensivechineselarge},
WebAggregatorQA~\cite{wang2025exploreevolvescalingevolved},
Mind2Web 2~\cite{gou2025mindweb}, MLR-Bench~\cite{chen2025mlrbenchevaluatingaiagents}, ArXivBench~\cite{li2025arxivbenchllmsassistresearchers}, PaperBench~\cite{starace2025paperbenchevaluatingaisability}, ResearchBench~\cite{liu2025researchbenchbenchmarkingllmsscientific},
ResearcherBench~\cite{xu2025researcherbenchevaluatingdeepai},
DeepScholar-Bench~\cite{patel2025deepscholarbenchlivebenchmarkautomated},
ResearchRubrics~\cite{sharma2025researchrubricsbenchmarkpromptsrubrics},
ReportBench~\cite{li2025reportbenchevaluatingdeepresearch},
AcademicBrowse~\cite{zhou2025scholarsearchbenchmarkingscholarsearching},
AstaBench~\cite{bragg2025astabenchrigorousbenchmarkingai},
LiveDRBench~\cite{java2025characterizingdeepresearchbenchmark},
LiveSearchBench~\cite{zhou2025livesearchbenchautomaticallyconstructedbenchmark},
ExpertLongBench~\cite{ruan2025expertlongbenchbenchmarkinglanguagemodels},
DeepResearch Arena~\cite{wan2025deepresearcharenaexamllms},
SPOT~\cite{son2025aicoscientistsfailspota},
DatasetResearch~\cite{li2025datasetresearchbenchmarkingagentsystems},
RigorousBench~\cite{yao2025rigorousbenchmarkmultidimensionalevaluation},
DRBench~\cite{abaskohi2025drbenchrealisticbenchmarkenterprise},
DeepShop~\cite{lyu2025deepshopbenchmarkdeepresearch},
FINDER~\cite{zhang2025fargenuinelyusefuldeep},
PHYBench~\cite{qiu2025phybenchholisticevaluationphysical},
ScienceAgentBench~\cite{chen2025scienceagentbench}, MicroVQA~\cite{burgess2025microvqa},
PDR-Bench~\cite{liang2025personalizeddeepresearchbenchmarks},
LiveResearchBench~\cite{wang2025liveresearchbenchlivebenchmarkusercentric},
LiveNewsBench~\cite{anonymous2025livenewsbench},
SealQA~\cite{pham2025sealqaraisingbarreasoning},
DR-Arena~\cite{gao2026drarenaautomatedevaluationframework}, DeepSearchQA~\cite{gupta2026deepsearchqabridgingcomprehensivenessgap}, P2P~\cite{sun2025p2pautomatedpapertopostergeneration},
and DeepResearch Bench II~\cite{li2026deepresearchbenchiidiagnosing}. 
These cover domains such as finance, science, policy, and engineering. Research questions are typically provided in the form of text, images, audio, videos, or PDF documents. Most benchmarks supply standard answers, though a few rely on expert human evaluation.
% \textcolor{red}{Reviewed}

\subsection{Architecture}
%Tongyi deep research~\cite{tongyideepresearchteam2025tongyideepresearchtechnicalreport}
Inspired by Anthropic’s Claude Research Agent and informed by published papers and open-source projects, the architecture of a deep research agent is generally as shown in Figure~\ref{fig:itr-dr}. 
%When a user submits a query, the lead agent analyzes it, develops a strategy, and spawns subagents to explore differents aspects simultaneously. The subagents act as intelligenct filters by iteratively using search tools to gather information, and then returning a list of companies to the lead agent so it can compile a final answer.
When a user submits a query, the system first confirms user intents through a simple interactive procedure, and then creates a MasterAgent that enters an iterative research process. The MasterAgent begins by thinking through the approach and saving its plan to Memory to persist the context. It then creates specialized SubAgents with specific research tasks. Each SubAgent independently performs searches, evaluates tool results using interleaved thinking, and returns findings to the MasterAgent. The MasterAgent synthesizes these results and decides whether more research is needed. Once sufficient information is gathered, the system exits the research loop and passes all findings to a ReviewAgent, which ensures all claims are properly attributed to their sources. The final research results, complete with citations, are then returned to the user.

\subsection{Pioneering Agents}
Pioneering deep research agents from AI perspective that provide users with accessible links are listed in Table~\ref{tab:industry}. We can see that, 1) Backbone large models, after agentic training, all gain the ability to perform tool-augmented deep research; 2) These agents consistently demonstrate search and review capabilities, \textbf{but their research capability remains comparatively weak, leaving a substantial gap between current systems and the vision of AI4S}; 3) Open-weight agents lag behind closed source systems in deep research performance, although the disparity is not significant; 4) DR agents built on open-weight models can outperform their backbones; 5) Current studies evaluate these agents primarily in IDE rather than in REE or SEE; 6) In terms of automation, DR systems are generally at level three. In addition, other notable open-source DR projects include \href{https://github.com/assafelovic/gpt-researcher}{GPT Researcher}, langchain ``\href{https://github.com/langchain-ai/open_deep_research}{open\_deep\_research}'' and ``\href{https://github.com/langchain-ai/local-deep-researcher}{local-deep-researcher}'', \href{https://github.com/jina-ai/node-DeepResearch}{node-DeepResearch}, ``\href{https://github.com/nickscamara/open-deep-research}{Open Deep Research}'', \href{https://github.com/dzhng/deep-research}{deep-research}, \href{https://github.com/mshumer/OpenDeepResearcher}{OpenDeepResearcher},
\href{https://github.com/TIGER-AI-Lab/OpenResearcher}{OpenResearcher},
\href{https://analemma.ai/blog/introducing-fars/}{FARS}, \href{https://github.com/karpathy/autoresearch}{autoresearch}, \href{https://github.com/bytedance/deer-flow}{DeerFlow}, and \href{https://github.com/RUC-NLPIR/WebThinker}{WebThinker}~\cite{li2025webthinkerempoweringlargereasoning}.
%paperdebugger~\cite{hou2025paperdebuggerpluginbasedmultiagentineditor}
%WisPaper~\cite{ju2025wispaperaischolarsearch}: \url{https://www.wispaper.ai/en}

%\subsection{Key Aspects}
% \section{Key Components}
% \label{sec:challenges}
% core components
% Modeling Challenges and Issues
%\subsection{Core Techniques}
\subsection{Key Aspects}
%agent issue, general agents
\paragraph{Datasets \& Benchmarks} As shown in Section~\ref{sec:bench}, many datasets and benchmarks have not yet been widely adopted. General DR agents can be trained on a broader range of high quality datasets to improve the generalizability of their research capability and intelligence. Domain researchers can develop dedicated datasets for their fields and then build their own DR agents on either open-weight or proprietary foundation models.

%~\cite{heidt2025ai}
\paragraph{Tools} ToolkenGPT represented each tool as a token and learned an embedding for it, enabling tool calls in the same way as generating a regular word token~\cite{hao2023toolkengpt}. SciMaster utilized a tool-augmented reasoning agent designed to emulate human researchers by interacting flexibly with external tools during its reasoning process~\cite{chai2025scimastergeneralpurposescientificai}. AutoTools was a framework that enables LLMs to act as automated tool learners, automating the tool-use workflow~\cite{Wildtool25}. Wu proposed an agentic reasoning framework by integrating mind-map, search, and code tools~\cite{wu2025agenticreasoningreasoningllms}. WebDancer provides the agent with search and click tools~\cite{wu2025webdancerautonomousinformationseeking}. FlashRAG is an efficient and modular open-source toolkit designed to assist researchers~\cite{FlashRAG_WWW}. TTE enables agents to synthesize, verify, and evolve executable tools during inference for scientific reasoning~\cite{lu2026statictoolstesttimetool}. \href{https://www.futuretools.io/}{FutureTools} and \href{https://www.bohrium.com/sciencepedia/agent-tools}{SciencePedia} provide tools for science in multiple fields.

\paragraph{Agent Framework} 
ResearStudio is a human-intervenable framework for building controllable 
DR Agents~\cite{yang2025researstudiohumanintervenableframeworkbuilding}. SFR-DeepResearch was a native single-agent featuring minimal web crawling and Python tool integration~\cite{nguyen2025sfrdeepresearcheffectivereinforcementlearning}. SciAgent operationalized scientific problem solving under a hierarchical Coordinator–Worker–Subagents framework~\cite{li2025sciagentunifiedmultiagentgeneralistic}. DeepResearcher  implemented a  multi-agent architecture where browsing agents extract relevant information from various webpage structures~\cite{zheng-etal-2025-deepresearcher}. TTD-DR conceptualized research report generation as a diffusion process~\cite{han2025deepresearchertesttimediffusion}. FlowSearch was a multi-agent framework that actively constructs and evolves a dynamic structured knowledge flow to drive subtask execution and reasoning~\cite{hu2025flowsearchadvancingdeepresearch}. MARS was a multi-agent system that seamlessly integrates System 1's fast, intuitive thinking with System 2's deliberate reasoning~\cite{chen2025marsoptimizingdualsystemdeep}. MiroFlow was a three-tier hierarchical agent framework for general deep research tasks~\cite{2025mirothinker}. WebWeaver was a dual-agent framework with a planner and a writer for open-ended deep research~\cite{li2025webweaverstructuringwebscaleevidence}. WebWatcher combined vision-language reasoning and multi-tool interaction~\cite{geng2025webwatcherbreakingnewfrontier}. O-Researcher was an open ended DR model via multi-agent distillation and agentic RL~\cite{yao2026oresearcheropenendeddeep}. Vision-DeepResearch performed multi-turn, multi-entity and multiscale visual and textual search to robustly hit real-world search engines under heavy noise~\cite{huang2026visiondeepresearchincentivizingdeepresearchcapability}. FS-Researcher was a file-system-based and dual-agent framework that scales deep research beyond the context window via a persistent workspace~\cite{zhu2026fsresearchertesttimescalinglonghorizon}.

%~\cite{behrouz2024titanslearningmemorizetest}
%llm or agent learning: \cite{song2025r1searcherincentivizingsearchcapability,jin2025empiricalstudyreinforcementlearning,wang2025stepsearchignitingllmssearch}
\paragraph{Agentic Learning} Atom-Searcher provided atomic thought rewards for fine-grained guidance to address conflicting gradients and reward sparsity in RL learning~\cite{deng2025atomsearcherenhancingagenticdeep}. DeepDive designed a redundancy penalty that discourages repeated similar queries in multi-turn RL~\cite{lu2025deepdiveadvancingdeepsearch}. Hong designed M-GRPO RL training methods for vertical multi-agent DR systems~\cite{hong2025multiagentdeepresearchtraining}. DeepPlanner trained the DR agent by GRPO with advantage shaping to scale its planning capability~\cite{fan2025deepplannerscalingplanningcapability}. DR Tulu used RL with evolving rubrics for learning in long-form tasks~\cite{drtulu}. PokeeResearch-7B is trained by an annotation-free RLAIF framework to optimize policies using LLM-based reward signals that capture factual accuracy, citation faithfulness, and instruction adherence~\cite{wan2025pokeeresearcheffectivedeepresearch}. IterResearch introduced efficiency-aware rewards and adaptive downsampling into the RL learning framework~\cite{anonymous2025iterresearch}. DeepSearch overcame the bottleneck of RL with verifiable rewards via Monte Carlo Tree Search~\cite{wu2025deepsearchovercomebottleneckreinforcement}. Search-R1++ was a strong DR agent adopting fast thinking templates and trained via REINFORCE with F1+ reward~\cite{xu2026traindeepresearchagent}.
%~\cite{scinips22}

% https://github.com/zechenzhangAGI/AI-research-SKILLs
\paragraph{Context \& Memory} 
The enterprise Dingtalk-DeepResearch was able to evolve via an entropy-guided, memory-aware online learning mechanism, retrieving high-value prior cases from an episodic memory bank and exploring diverse historical contexts~\cite{chen2025dingtalkdeepresearchunifiedmulti}. WebResearcher~\cite{qiao2025webresearcherunleashingunboundedreasoning} and IterResearch~\cite{anonymous2025iterresearch} introduced a Markovian structure to build effective context and memory. EvoFSM proposed a self-evolving mechanism for DR with finite state machines~\cite{zhang2026evofsmcontrollableselfevolutiondeep}. K-Dense BYOK is a free, open-source AI co-scientist that runs on your desktop, powered by Claude Scientific Skills~\cite{claudescientificskills2026}. PantheonOS implemented an extensible skill system encoding domain expertise as markdown templates with structured workflows for automatic genomics discovery~\cite{Xu2026bioskill}.

%\paragraph{Planning \& Reasoning} xxx
% main issues
% research topics
% Taxonomy of Deep research

\section{AI4S Perspective}
%\textit{``Training an AI system with a knowledge cutoff of 1911 and seeing if it could come up with general relativity like Einstein did in 1915. That’s the true test of whether we have a full AGI system.''}    
%\hfill --- Demis Hassabis
\textit{``Training an AI system with a knowledge cutoff of 1911 and seeing if it could come up with general relativity like Einstein did in 1915.''}\hfill ---Demis Hassabis
\subsection{Related Summaries and Platforms}
Tom provided an overview of self-driving laboratories for chemistry and materials science~\cite{tom2024self}.  Gao and Wang found that the use of AI in research is
widespread throughout the sciences, growing especially rapidly since 2015 ~\cite{Gao2024te}. Messeri and Crockett were concerned that the proliferation of AI tools in science risks introducing a phase of scientific enquiry in which we produce more but understand less~\cite{Messeri2024}.
OpenAI and researchers presented a collection of short case studies in which GPT-5 produced new, concrete steps in ongoing research across mathematics, physics, astronomy, computer science, biology, and materials science~\cite{bubeck2025earlyscienceaccelerationexperiments}. Si found that LLM-generated ideas are judged as more novel than human expert ideas while being judged slightly weaker on feasibility~\cite{Si2025Gap,Si2025Can}. Ren provided a review of the architectures, design, benchmarks, applications, and ethical considerations surrounding LLM-based scientific agents~\cite{ren2025scientificintelligencesurveyllmbased}. Wei offered a domain-oriented review of autonomous scientific discovery across life sciences, chemistry, materials, and physics~\cite{wei2025aiscienceagenticscience}. Hu reviewed recent Sci-LLMs, from general-purpose foundations to specialized models across diverse scientific disciplines, alongside an extensive analysis of training datasets~\cite{hu2025surveyscientificlargelanguage}.
Zheng provided a conceptual architecture and strategic foresight to navigate and shape the future of AI-driven scientific discovery~\cite{zhengetal2025automation}.
 Trehan and Chopra reported lessons from four autonomous research attempts using LLMs~\cite{trehan2026llmsarentscientistsyet}. Hao stated that AI tools expand scientists' impact but contract science's focus~\cite{Hao2026}.

At present, the accessible AI4S platforms include \href{https://www.futurehouse.org/?utm_source=chatgpt.com}{FutureHouse}, 
\href{https://platform.edisonscientific.com/login}{Edison}, 
\href{https://app.researchrabbit.ai/}{ResearchRabbit},
\href{https://scispace.com/}{SciSpace},
\href{https://scite.ai/}{scite\_},
\href{https://sider.ai/}{Sider},
\href{https://elicit.com/}{Elicit},
\href{https://www.autoscience.ai/}{Autoscience}, 
\href{https://www.deepprinciple.com/cn/product.html}{Deep Principle},
\href{https://hypogenic.ai/}{hypogenic.ai}, and \href{https://discovery.intern-ai.org.cn/}{Intern-Discovery}. 
The open-source AI4S platforms include \href{https://github.com/ymx10086/ResearchClaw}{ResearchClaw}, \href{https://github.com/karpathy/autoresearch}{autoresearch}, \href{https://github.com/wanshuiyin/Auto-claude-code-research-in-sleep}{Auto-claude-code-research-in-sleep}, \href{https://github.com/aiming-lab/AutoResearchClaw}{AutoResearchClaw}, \href{https://github.com/beita6969/ScienceClaw}{ScienceClaw}, \href{https://github.com/OpenRaiser/NanoResearch}{NanoResearch}, \href{https://github.com/openlair/dr-claw}{dr-claw}, 
\href{https://github.com/gair-nlp/asi-evolve}{ASI-Evolve},
and \href{https://github.com/SakanaAI/AI-Scientist-v2}{AI-Scientist}~\cite{Lu2026}.

%\subsection{Dataset and Benchmark}
% task

\begin{figure}[t]
	\includegraphics[width=\columnwidth]{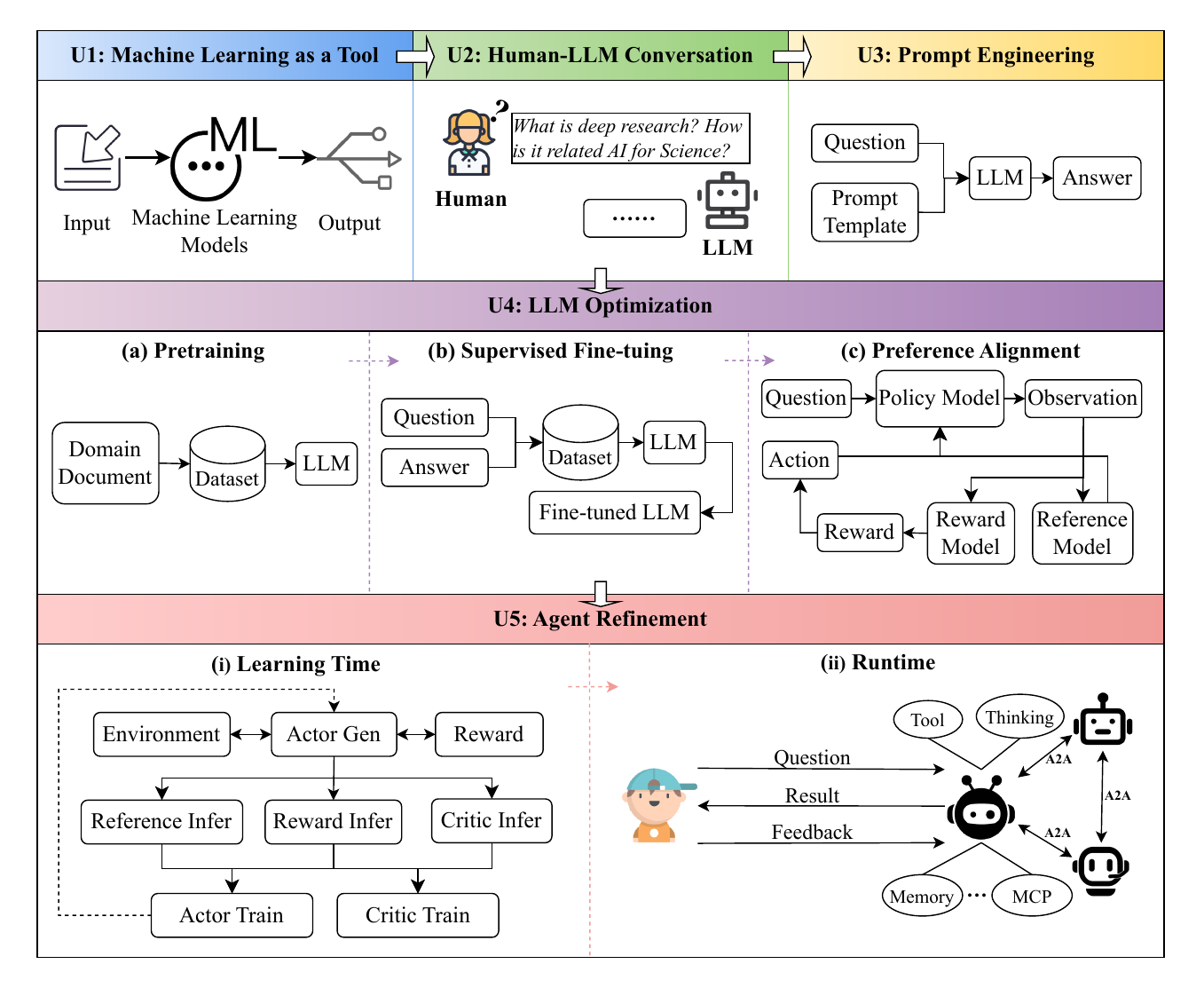}
	\caption{Five interaction paradigms in AI4S.}
	\label{fig:use}
\end{figure}

\subsection{Paradigm}
%transfer learning phase, scaling phase, instruction-following phase, agentic science
As shown in Figure~\ref{fig:use}, we categorize the interaction paradigms between researchers and AI in AI4S into five types. The fourth type, $U4$, can be further divided into three subtypes, and the fifth type, $U5$, into two subtypes.
\paragraph{\textbf{\textit{U1}}: Machine Learning as a Tool} Before ChatGPT appeared, researchers commonly referred to AI precursors as machine learning algorithms. These algorithms were primarily used to process and model experimental data. Within this paradigm, machine learning functioned as a tool.
\paragraph{\textbf{\textit{U2}}: Human-LLM Conversation} After LLMs like ChatGPT gained acceptance, researchers began treating LLMs as more advanced search engines or smarter information processing systems. In this paradigm, researchers typically interact with these models through conversation.
\paragraph{\textbf{\textit{U3}}: Prompt Engineering} Compared with $U2$, $U3$ involves more complex conversation. Researchers use prompt engineering to encourage LLMs to produce more effective responses.
\paragraph{\textbf{\textit{U4}}: LLM Optimization} In $U4$, researchers build their own LLMs. They use methods such as pretraining (a), fine-tuning (b), and alignment preference (c). These models may be based on open-weight LLMs or developed entirely from scratch.
\paragraph{\textbf{\textit{U5}}: Agent Refinement} The use of tools drives the transition from $U4$ to $U5$. Researchers can train models using agentic RL (i). Alternatively, they can build agents with LLMs that possess tool capabilities (ii).
%include context engineerning and harness engineering
%three phases: scientific data, scientific assistant, scientific agent
%dialog-prompt-rl-agent~\cite{liu2025promptr1collaborativeautomaticprompting}
\subsection{Fields}
% domain, disciplines
For each field of scientific research, we first present the relevant datasets/benchmarks ($DBs$) and $Tools$, then describe approaches following the paradigm in the order of $U1\rightarrow U2\rightarrow U3\rightarrow U4\rightarrow U5$. 
%\textbf{Note that for each specific field (non-general), we only consider representative research works that have already undergone rigorous review and been formally accepted.}
Note that if a study uses a Transformer model with a relatively small number of parameters, we classify it as $U1$ rather than $U3$.
\subsubsection{Task-Agnostic \& Multi-Task}
%disciplines
%Multi-Tasks
%SCIMON idea generation~\cite{wangetal2024scimon}, ~\cite{NEURIPS2024}
\textbf{\textit{[DBs]}}: QASA benchmark consists of 1,798 novel question answering pairs that require full-stack reasoning on scientific articles in AI and ML fields~\cite{QASA23}. PIQA is a large-scale QA dataset specifically designed to interpret complex figures and tables within the context of scientific research articles across various domains of computer science~\cite{pramanick2024spiqa}. Multimodal ArXiv is a dataset for improving scientific comprehension of large vision-language models~\cite{li-etal-2024-multimodal-arxiv}. SciEval is a multi-level LLM evaluation benchmark for scientific research~\cite{SunYu2024}. SciBench contains carefully curated dataset featuring a range of collegiate-level scientific problems from mathematics, chemistry, and physics domains~\cite{wang2024scibench}. OlympiadBench is an Olympiad-level bilingual multimodal scientific benchmark, featuring 8,476 problems from Olympiad-level mathematics and physics competitions, including the Chinese college entrance exam~\cite{he2024olympiadbench}. Song introduced a scenario grounded benchmark that evaluates LLMs across biology, chemistry, materials, and physics~\cite{song2025evaluatinglargelanguagemodels} . Scientist-Bench is a comprehensive benchmark comprising state-of-the-art papers across diverse AI research domains~\cite{airesearcher}. Liu introduced ATLAS benchmark, which is a cross-disciplinary evaluation suite composed of approximately 800 original problems~\cite{liu2025atlashighdifficultymultidisciplinarybenchmark}. LIMITGEN is a benchmark for evaluating LLMs' capability to support early-stage feedback and complement human peer review~\cite{xuetal2025llmsidentify}. ScienceAgentBench evaluated language agents for data-driven scientific discovery, extracting 102 tasks from 44 peer-reviewed publications in four disciplines~\cite{chen2025scienceagentbench}. SciArena is an open and collaborative platform for evaluating foundation models on scientific literature-grounded tasks~\cite{zhao2025sciarena}. SCIVER is a benchmark designed to evaluate the ability of foundation models to verify claims within a multimodal scientific context~\cite{wangetal2025sciver}. Humanity’s Last Exam is a multi-modal benchmark at the frontier of human knowledge, designed to be an expert-level closed-ended academic benchmark with broad subject coverage~\cite{phan2025humanitysexam}.
%\href{https://github.com/GAIR-NLP/Data-Darwinism?tab=readme-ov-file#data-darwinism----part1-unlocking-the-value-of-scientific-data-for-pre-training}{Data Darwinism} 
\textbf{\textit{[Tools]}}: SciToolAgent leveraged a scientific tool knowledge graph across biology, chemistry, and materials science that enables intelligent tool selection and execution through graph-based RAG~\cite{ding2025scitoolagent}. 
\textbf{\textit{[U3]}}: PersLEARN is a tool designed to facilitate the cultivation of scientific perspectives, starting from a basic seed idea and progressing to a well-articulated framework~\cite{shi2023perslearn}. \textbf{\textit{[U4]}}:
OpenResearcher was built based on RAG to integrate LLMs with up-to-date, domain-specific knowledge~\cite{zheng2024openresearcherunleashingaiaccelerated}.
GraphEval is a lightweight graph-based LLM framework for idea evaluation~\cite{feng2025graphevallightweightgraphbasedllm}. Goel leveraged the vast corpus of existing research papers to train LLMs that generate better research plans~\cite{goel2025trainingaicoscientistsusing}. Intern-S1 is a scientific multimodal foundation model with 28 billion activated parameters and 241 billion total parameters~\cite{bai2025interns1scientificmultimodalfoundation}. SciReasoner introduced a scientific language foundation model that bridges general-purpose large language modeling with the heterogeneous data and reasoning workflows of the natural sciences~\cite{wang2025scireasonerlayingscientificreasoning}. TTT-Discover performed RL at test time, so the LLM can continue to train, but with experience specific to the test scientific problem~\cite{tttdiscover2026}. Innovator-VL is a scientific MLLM designed to advance multimodal understanding and reasoning across diverse scientific domains while still maintaining excellent performance on general vision tasks~\cite{wen2026innovatorvlmultimodallargelanguage}. \textbf{\textit{[U5]}}: Le proposed a multi-agent deep research MLLMs system for multimedia verification~\cite{le2025multimediaverificationmultiagentdeep}. SciSciGPT is a multi-agent designed to serve as a research collaborator for science of science researchers and practitioners~\cite{shao2025sciscigptadvancinghumanaicollaboration}. aiXiv is a multi-agent ecosystem, which allows research proposals and papers to be submitted, reviewed, and iteratively refined by both human and AI scientists~\cite{zhang2025aixivnextgenerationopenaccess}. VIRSCI organized a team of agents to collaboratively generate, evaluate, and refine research ideas~\cite{su-etal-2025-many}. PiFlow is an information-theoretical multi-agent framework, treating automated scientific discovery as a structured uncertainty reduction problem guided by principles~\cite{pu2025piflowprincipleawarescientificdiscovery}. Denario is a multi-agent system designed to serve as a research assistant for scientific discovery~\cite{villaescusanavarro2025denarioprojectdeepknowledge}. 
Kosmos is a multi-agent AI scientist that automates data-driven discovery~\cite{mitchener2025kosmosaiscientistautonomous}. AI co-scientist is a multi-agent system to help uncover new, original knowledge and formulate demonstrably novel research hypotheses and proposals~\cite{gottweis2025aicoscientist}. Li proposed a multi-agent framework to decompress scientific reasoning and construct a verifiable long CoT knowledge base~\cite{li2025inverseknowledgesearchverifiable}.
SAGA is a bi-level agent to accelerate scientific discovery for antibiotic design, inorganic materials design, functional DNA sequence design, and chemical process design~\cite{du2025acceleratingscientificdiscoveryautonomous}. InternAgent is a unified closed-loop multi-agent framework to conduct autonomous scientific research across various scientific research fields~\cite{internagentteam2025}. AgentExpt is a framework for baseline and dataset recommendation~\cite{li2025agentexptautomatingaiexperiment}. Deep Ideation integrated LLMs with scientific networks to generate novel and scientifically grounded research ideas~\cite{deep_ideation_2025}. AI-Researcher is a multi-agent orchestrating literature review, idea generation, algorithm implementation, experimental validation, and paper writing~\cite{airesearcher}. Chain of Ideas agent offered a promising and concise solution by organizing ideas into a chain structure, effectively mirroring the progressive development
 within a given research domain~\cite{li2024chainideasrevolutionizingresearch}.
 URSA is a scientific agent ecosystem for accelerating research tasks, which consists of a set of modular agents and tools~\cite{grosskopf2025ursauniversalresearchscientific}. RDR is a generalizable pipeline capable of systematically analyzing AI, robotics and beyond: identifying emerging trends, uncovering cross-domain opportunities, and offering concrete starting points for new inquiry~\cite{zou2025realdeepresearchai}.
 DEPLOY-MASTER constructed reproducible runtime environments for 50,112 scientific tools, and each successful tool is validated by a minimal executable command and registered in \href{https://www.bohrium.com/en/sciencepedia}{SCIENCEPEDIA} for search and reuse~\cite{wang2026deploymaster}. MARVEL is a locally deployable, open-source framework for domain-aware question answering and assisted scientific research~\cite{mukund2026marvelmultiagentbasedresearch}. EvoScientist is an evolving multi-agent AI scientist framework that continuously improves its research strategies through persistent memory and self-evolution~\cite{lyu2026evoscientistmultiagentevolvingai}. AI Scientist used existing foundation models to perform ideation, literature search, experiment planning and implementation, result analysis, manuscript writing, and peer review to produce complete, new papers of machine learning science~\cite{Lu2026}.

\subsubsection{Biology}
%Gao presents a perspective summary of biomedical discovery with AI agents~\cite{gao2024empowering}.
\textbf{\textit{[U1]}}: AlphaFold used a Transformer-like neural network to predict the three-dimensional structure that a protein will adopt based solely on its amino acid sequence~\cite{Jumper2021}. Wang used deep learning approaches for scaffolding protein functional sites without needing to prespecify the fold or secondary structure of the scaffold~\cite{scienceabn2100}. AlphaMissense is an adaptation of AlphaFold fine-tuned on human and primate variant population frequency databases to predict missense variant pathogenicity~\cite{AlphaMissense}. CLEAN is a contrastive learning algorithm for enzyme annotation~\cite{Enzyme23}. Lutz used Monte Carlo tree search with RL to design protein architectures~\cite{proteinscience23}. LinearDesign algorithm found an optimal mRNA design for the spike protein in just 11 minutes, and concurrently optimized stability and codon usage~\cite{Zhang2023}. RoseTTAFold was proposed to design protein structure and function, which used diffusion architecture to model protein backbone geometry and sequence–structure relationships~\cite{Watson2023}. Chroma is a diffusion model for proteins and protein complexes that can directly sample novel protein structures and sequences~\cite{Ingraham2023}. NAErnie is an RNA-focused pretrained model built upon the transformer architecture~\cite{wang2024multi}. AlphaFold 3 is a diffusion-based architecture that is capable of predicting the joint structure of complexes including proteins, nucleic acids, small molecules, ions, and modified residues~\cite{Abramson2024}. MxDNA is a framework developed to autonomously learn effective DNA tokenization strategies solely through gradient descent~\cite{MxDNA24}. scGPT is a foundation model for single-cell biology, which is based on a generative pretrained transformer across a repository of over 33 million cells~\cite{cui2024scgpt}. GEMORNA is a transformer encoder-decoder capable of designing mRNA sequences with unprecedented translational capacity and durability~\cite{mRNA-science}.
\textbf{\textit{[U4]}}: Lin demonstrated direct inference of full atomic-level protein structure from primary sequence using a LLM~\cite{lin2023evolutionary}. scFoundation is a large pretrained model with 100 million parameters covering about 20,000 genes, pretrained on over 50 million human single-cell transcriptomic profiles~\cite{hao2024large}. BiomedGPT is an open-source and lightweight vision–language foundation model for diverse biomedical tasks~\cite{zhang2024generalist}. UniFMIR is a pre-trained foundation models for universal fluorescence microscopy image restoration ~\cite{Ma2024}. scTranslator utilized an encoder-decoder Transformer-based architecture for translating single-cell transcriptome to proteome~\cite{Liu2025}. ESM3 is a multimodal generative language model that reasons over the sequence, structure, and function of proteins~\cite{hayes2025simulating}. Omni-DNA is a family of models spanning 20M to 1.1B parameters that supports sequence understanding, long-context genomic reasoning, and natural language annotation~\cite{liomni}. 
%Evo is a genomic language model, which learns a distributional semantics over genes that enables the function-guided design of new sequences reflecting prokaryotic functional relationships~\cite{}. 
LucaOne is a pre-trained foundation model trained on nucleic acid and protein sequences from 169,861 species~\cite{he2025generalized}. AlphaGenome used a U-Net-inspired backbone with transformer blocks to analyze the regulatory genome for predicting molecular functions and variant effects from DNA~\cite{Avsec2026}. EnzymeCAGE is a catalytic-specific geometric foundation model trained on approximately 1.5 million structure-informed enzyme–reaction pairs spanning over 3,000 species~\cite{Liu2026dk}. Evo is a biological foundation model trained on 9 trillion DNA base pairs from a highly curated genomic atlas spanning all domains of life~\cite{nguyen2024sequence,Merchant2025at,Brixi2026}.
\textbf{\textit{[U5]}}:
\href{https://github.com/snap-stanford/Biomni}{Biomni} integrated LLM reasoning with RAG and code-based execution to help scientists dramatically enhance research productivity and generate testable hypotheses~\cite{huang2025biomni}. K-Dense Analyst is a hierarchical multi-agent system that achieves autonomous bioinformatics analysis through a dual-loop architecture~\cite{li2025kdenseanalystfullyautomated}. LabOS AI is a co-scientist for the biomedical domain that unites computational reasoning with physical experimentation through multimodal perception, self-evolving agents, and XR-enabled human-AI collaboration~\cite{cong2025labosaixrcoscientistsees}. Virtual Lab, an AI–human research collaboration multi-agent, was used to design nanobody binders to recent variants of SARS-CoV-2~\cite{swanson2025virtual}. ChatNT is a multimodal conversational agent to bridge the gap between biology foundation models and conversational agents~\cite{de2025multimodal}. CellWhisperer established a user-friendly approach for exploring scRNA-seq data, driven by chat-based analysis with natural language~\cite{schaefer2025multimodal}.

\subsubsection{Materials}
%\paragraph{Dataset/Bench}
\textbf{\textit{[U1]}}: Raccuglia used ML algorithms trained on reaction data to predict reaction outcomes for the crystallization of templated vanadium selenites~\cite{Raccuglia2016}. Generative models were used for inverse molecular design in matter engineering~\cite{molecularscience}. Tshitoyan captured latent knowledge from materials science literature through unsupervised word embeddings~\cite{Tshitoyan2019}. Burger used a mobile robot to search for improved photocatalysts for hydrogen production from water with a batched Bayesian search algorithm~\cite{Burger2020}. GNoME is graph neural networks for efficient discovery of inorganic materials~\cite{Merchant2023}. A-Lab is an autonomous laboratory for the solid-state synthesis of inorganic powders with machine learning and active learning~\cite{Szymanski2023}. InvDesFlow-AL is an active learning-based diffusion model for designing target functional inorganic crystal materials across the periodic table~\cite{han2025invdesflow}. MatterGen is a diffusion-based generative model that generates stable, diverse inorganic materials across the periodic table and can be fine-tuned towards a wide range of downstream tasks for inverse materials design~\cite{Zeni2025}. MatRIS leveraged attention to model three-body interactions for quantum mechanism calculation in materials~\cite{zhou2026matris}.
\textbf{\textit{[U4]}}: CrystaLLM is  an autoregressive LLM for the versatile generation of crystal structures~\cite{antunes2024crystal}.
\textbf{\textit{[U5]}}: SciAgents is a multi-agent designed to autonomously generate and refine research hypotheses by
leveraging LLMs and a comprehensive ontological knowledge graph~\cite{ghafarollahi2024sciagents,buehler2024graphreasoning}. ChatMOF is an AI system for predicting and generating metal-organic frameworks using LLMs, tools, and evaluators~\cite{kang2024chatmof}.

\subsubsection{Healthcare}
\textbf{\textit{[DBs]}}: MultiMedQA is a benchmark combining six existing medical question answering datasets spanning professional medicine, research and consumer queries, and HealthSearchQA, a new dataset of medical questions searched online~\cite{Singhal2023}. 
 OmniMedVQA is a comprehensive medical visual question answering (VQA) benchmark, including 12 different modalities and covering more than 20 distinct anatomical regions~\cite{hu2024omnimedvqa}.
 GMAI-MMBench is a general medical benchmark with 284 datasets across 38 medical image modalities, 18 clinical-related tasks, 18 departments, and 4 perceptual granularities in a VQA format~\cite{GMAIMMBench}.
\textbf{\textit{[U1]}}: Esteva demonstrated classification of skin lesions using a single CNN, trained end-to-end from images directly, using only pixels and disease labels as inputs~\cite{Esteva2017}. Barata utilized a RL model for AI-based decision support in skin cancer~\cite{Barata2023}. Steyaert fused multimodal data for cancer biomarker discovery with deep learning~\cite{steyaert2023multimodal}. RadDiag is a transformer-based foundational model for large-scale long-tailed disease diagnosis on radiology images~\cite{Zheng2024}. OISA is a post-training framework based on a pre-trained CLIP model for radiology report generation with self-generation, self-evaluation, self-alignment, and self-iteration~\cite{xiao-etal-2025-online}. MAOSS is multi-modal and transformer-based AI for opportunistic screening, staging and progression risk stratification of steatotic liver disease~\cite{Gao2026}. \textbf{\textit{[U2]}}: Bean conducted a randomized study testing the effects of using LLMs to support medical self-assessment, and highlights the challenges of public deployments of LLMs for direct patient care~\cite{Bean2026}.
\textbf{\textit{[U4]}}: Flan-PaLM and Med-PaLM are instruction-tuned variants of PaLM on clinical data~\cite{Singhal2023,singhal2025toward}. GMAI is a foundation models proposed for generalist medical artificial intelligence~\cite{Moor2023}. Zhongjing is a Chinese medical LLaMA-based LLM for Chinese medicine~\cite{Yangmedical2024}. Delphi-2M is a GPT-based architecture to predict the rates of more than 1,000 diseases, conditional on each individual's past disease history diseases~\cite{Shmatko2025}. SlideChat is a large vision-language assistant for whole-slide pathology image understanding~\cite{chen2025slidechat}. CSFM is a multimodal foundation model pretrained on data from 1.7 million individuals for cardiac health assessment across scenarios and devices~\cite{Gu2026ax}.
\textbf{\textit{[U5]}}: PathChat is a vision-language generalist AI assistant for human pathology~\cite{lu2024multimodal}. LLM-RDF is a chemical synthesis development platform powered by GPT-4, which comprises six specialized LLM-based agents, including Literature Scouter, Experiment Designer, Hardware Executor, Spectrum Analyzer, Separation Instructor, and Result Interpreter~\cite{ruan2024automatic}. AMIE is an LLM-based AI system optimized for diagnostic dialogue~\cite{Tu2025,McDuff2025}. DeepRare is a multi-agent system with 40 specialized tools and up-to-date knowledge sources for rare disease differential diagnosis decision support~\cite{Zhao2026uu}.

\subsubsection{Medicine}
%~\cite{Sadybekov2023}
%\paragraph{Dataset/Bench}
\textbf{\textit{[U1]}}: Similarity-based machine learning approaches were used to predict new molecular targets for known drugs~\cite{Keiser2009}. Deep neural networks were used to predict molecules with antibacterial activity~\cite{stokes2020deep}. RosettaVS is a structure-based virtual screen method based on active learning to predict docking poses and binding affinities for drug discovery~\cite{Zhou2024}. DrugCLIP combined contrastive learning and dense retrieval based on a transformer architecture to achieve rapid and accurate genome-wide virtual screening~\cite{DrugCLIP26}.
\textbf{\textit{[U4]}}: MMed-Llama 3 is an 8B multilingual language model for medicine~\cite{qiu2024towards}. InstructMol is a multi-modal LLM which effectively aligns molecular structures with natural language via an instruction-tuning approach~\cite{caoetal2025instructmol}.
\textbf{\textit{[U5]}}: MolRL-MGPT used a RL algorithm with multiple GPT agents for drug molecular generation~\cite{NEURIPS202}.
%~\cite{modulatorsscienceadi8577}

\subsubsection{Chemistry}
\textbf{\textit{[DBs]}}: ChemBench is an automated framework for evaluating the chemical knowledge and reasoning abilities of LLMs against the expertise of chemists~\cite{Mirza2025}.
\textbf{\textit{[U1]}}: MCTS combined deep neural networks and symbolic rules to perform chemical synthesis planning~\cite{Segler2018}. 
Reac-Discovery used ML models for process optimization and reactor geometry refinement~\cite{tinajero2025reac}.
\textbf{\textit{[U4]}}: Chemma is a fully fine-tuned LLM with 1.28 million pairs of Q\&A about reactions, as an assistant to accelerate organic chemistry synthesis~\cite{zhang2025large}. ChemVLM is an open-source chemical MLLM, which is trained on a bilingual multimodal dataset including molecular structures, reactions, and chemistry examination questions~\cite{ChemVLM2025}. QFANG is a scientific reasoning model for organic synthesis procedure generation~\cite{liu2025scientificreasoningmodelorganic}. MOSAIC is a computational framework that fine-tunes the open-weight Llama 3.1-8B-instruct model into 2,498 specialized chemistry experts~\cite{Li2026}.
\textbf{\textit{[U5]}}: Coscientist is a system driven by GPT-4 that autonomously designs, plans, and performs complex chemical experiments with tools such as internet and documentation search, code execution and experimental automation~\cite{boiko2023autonomous}. ChemCrow is a chemistry agent with 18 expert-designed tools designed to accomplish tasks across organic synthesis, drug discovery and materials design~\cite{m2024augmenting}.

\subsubsection{Mathematics}
%~\cite{Fawzi2022}
\textbf{\textit{[DBs]}}: MATHVISTA is a benchmark designed to combine challenges from diverse mathematical and visual tasks~\cite{lu2024mathvista}.
\textbf{\textit{[U1]}}: Davies demonstrated a method by which machine learning can aid mathematicians in discovering new conjectures and theorems~\cite{Davies2021}. Alfarano trained sequence-to-sequence transformers to discover a Lyapunov function that ensures the global stability of a dynamical system~\cite{alfarano2024global}. 
\textbf{\textit{[U3]}}: DSP+ is an improved version of the Draft, Sketch, and Prove framework for advanced theorem proving using LLMs, featuring a fine-grained and integrated neurosymbolic enhancement~\cite{cao2025revivingdspadvancedtheorem}.
\textbf{\textit{[U4]}}: LLEMMA is an open language model for mathematics by pretraining Code Llama on Proof-Pile-2~\cite{azerbayev2024llemma}. POSEIDON is a foundation model based on a multiscale operator transformer for learning the solution operators of PDEs~\cite{PDE24}. Math-Shepherd is a process-oriented math verifier, which assigns a reward score to each step of the LLM's outputs on math problems~\cite{wangetal2024math}. FunSearch is an evolutionary procedure based on pairing a pretrained LLM with a systematic evaluator for mathematical discoveries~\cite{RomeraParedes2024}. STP is a self-play LLM theorem prover with iterative conjecturing and proving~\cite{dong2025stp}.
\textbf{\textit{[U5]}}: AlphaGeometry is a neuro-symbolic system that uses a neural language model to prove theorems in Euclidean plane geometry by synthesizing millions of theorems and proofs across varying complexity levels~\cite{Trinh2024}. TORA is a series of novel tool-integrated reasoning agents that synergistically combines natural language rationale with program-based tool-use for mathematical problem solving~\cite{gou2024tora}. AlphaProof is an AI agent that learns to find formal proofs through RL by training on millions of auto-formalized problems~\cite{Hubert2025}. AlphaEvolve is an LLM-based code-mutation agent that helps researchers make advances in complexity theory~\cite{nagda2026reinforcedgenerationcombinatorialstructures}. Aletheia is a math research agent that iteratively generates, verifies, and revises
solutions end-to-end in natural language, leveraging a novel inference-time scaling law based upon Gemini Deep Think~\cite{feng2026aletheiatacklesfirstproofautonomously,feng2026autonomousmathematicsresearch}.

\subsubsection{Physics}
%~\cite{Wu2021}, ~\cite{wu2026geoptscalingphysicssimulation}, ~\cite{Ruan2026}
\textbf{\textit{[DBs]}}: NewtonBench is a benchmark comprising 324 scientific law discovery tasks across 12 physics domains~\cite{zheng2026newtonbench}.
\textbf{\textit{[U1]}}: Canabarro employed both unsupervised and supervised machine learning to identify quantum phase transitions~\cite{PhysRevB100}. Wu investigated an ``AI Physicis'' learning agent for unsupervised learning~\cite{PhysRevE100}. Seif used machine learning models to infer the direction of time's arrow identifies entropy production as the relevant physical quantity in its decision-making process~\cite{Seif2021}. Degrave presented a paradigm for plasma magnetic confinement on tokamaks through deep RL in nuclear fusion~\cite{Degrave2022}. TQS is a transformer-based model for quantum many-body problems~\cite{PhysRevB107075147}. Reinschmidt introduced RL to cold atom experiments and demonstrated a flexible and adaptive approach to control a magneto-optical trap~\cite{Reinschmidt2024}. Belis used an unsupervised kernel machine and two clustering algorithms to detect quantum anomaly detection~\cite{Belis2024hb}. Zhang also provided a technical and unified review of AI for quantum, atomistic, and continuum systems~\cite{zhang2023artificial}.
\textbf{\textit{[U3]}}: Pan carried out the quantum many-body physics calculations using LLMs by multistep prompt templates~\cite{Pan2025}.
\textbf{\textit{[U5]}}: SGA is a scientific generative agent in which LLMs act as knowledgeable and adaptable reasoners to propose scientific solutions such as physics equations or molecular structures, while simulations serve as experimental platforms that provide observational feedback and optimize continuous components like physical parameters~\cite{ma2024llm}. AI-Newton is a concept-driven discovery workflow capable of autonomously deriving physical laws from raw data~\cite{fang2025ainewtonconceptdrivenphysicallaw}.

\subsubsection{Meteorology}
%\paragraph{Dataset/Bench}
\textbf{\textit{[U1]}}: Ham used a CNN model to produce skilful ENSO forecasts for lead times of up to one and a half years~\cite{Ham2019}. GraphCast is a machine learning–based method trained from reanalysis data to learn skillful medium-range global weather forecasting~\cite{weatherskill23}. Pangu-Weather used a 3D transformer-based encoder-decoder model for fast and accurate global weather forecast~\cite{Bi2023}. NeuralGCM is a differentiable hybrid atmospheric model that combines the strengths of traditional general circulation models with machine learning for weather forecasting and climate simulation~\cite{Kochkov2024}. GenCast is a conditional diffusion model for probabilistic weather forecasting~\cite{Price2025}. FuXi-CFD is a machine learning-based framework designed to generate detailed 3D near-surface wind fields at 30-meter horizontal resolution, using only coarse-resolution atmospheric inputs and high-resolution terrain information~\cite{Lin2026}.
%To model the different sources of uncertainty in weather forecasts and capture the joint spatial structure of forecast, FGN generates ensembles via learned model-perturbations with an ensemble of
%appropriately constrained models distributions~\cite{alet2025skillfuljointprobabilisticweather}.
%HealDA is a Transformer-based encoder-decoder architecture for initializing global weather forecasts with
%only raw point cloud data as input~\cite{gupta2026healdahighlightingimportanceinitial}.
%Stormscope is a family of Transformer-based generative diffusion models trained on high-resolution, multi-band geostationary satellite imagery and ground-based weather radar~\cite{pathak2026learningaccuratestormscaleevolution}.
%The ATLAS framework enables high-accuracy weather prediction for medium-range forecasts based on a latent transformer and a local projector~\cite{kossaifi2026demystifyingdatadrivenprobabilisticmediumrange}.

\begin{figure*}[!th]
	\includegraphics[width=\textwidth]{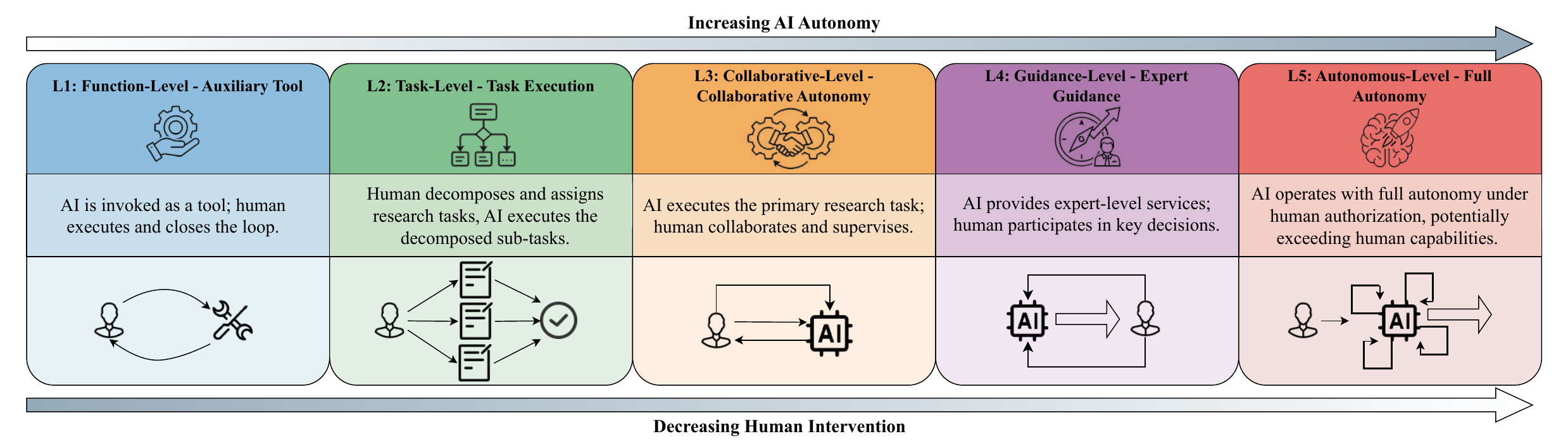}
	\caption{Human-AI collaboration in science.}
	\label{fig:paradigm}
\end{figure*}

%\subsubsection{Machine Learning}
%~\cite{zhu2026ultralonghorizonagenticsciencecognitive},~\cite{oh2025discovering}
%Agent Laboratory is an open-source LLM agent framework for accelerating the individual's ability to perform research in machine learning~\cite{schmidgall2025agentlaboratoryusingllm}.

%\subsubsection{Data Science}
%~\cite{nam2025dsstardatascienceagent}, ~\cite{anonymous2026srscientist}
%Kosmos~\cite{mitchener2025kosmosaiscientistautonomous,peng2023kosmos2groundingmultimodallarge}

% \section{Key Components}
% \label{sec:challenges}
% core components
% Modeling Challenges and Issues
% Core Techniques
% main issues
% research topics

\subsection{Present and Perspective of AI4S}
%\begin{quote}
%	\textit{``The dawn of vibe research may make this my last truly handwritten paper..''}    
%	\hfill --- This Paper
%\end{quote}
\begin{quote}
\textit{``AI for Science is the new lens of discovery, much like cryo-electron microscopes for proteins, particle accelerators for physics, and telescopes for astronomy.''}\hfill --- This paper
\end{quote}
%Anthropology, life science
Current research demonstrates substantial progress for AI4S in fields such as biology, materials, healthcare, medicine, chemistry, mathematics, physics, and meteorology. AI is also playing an increasing role in other fields such as robotics~\cite{Kaufmann2023,Ilijahumanoid,Tuomassoccer,Lu2025}, neuroscience~\cite{scienceaav9436,Luo2025}, aerospace~\cite{Reichstein2019}, agriculture~\cite{ying-etal-2025-seedbench}, operations research~\cite{wang-etal-2025-ormind}, nuclear reactions~\cite{scienceadm8201}, geography~\cite{brown2025alphaearthfoundationsembeddingfield}, and finance~\cite{jin2025finrptdatasetevaluationllmbased}. 
No single interaction paradigm is inherently superior to another. The choice of paradigm depends on the specific research problem. Even simple multi-turn dialogues (\textbf{\textit{U2}}) can provide valuable assistance to scientists. However, several limitations persist. 
%Automation levels remain inadequate for fully autonomous research. While agent-based paradigms enhance efficiency, scientists must still perform numerous manual operations (work for agents).
Automation levels remain insufficient for fully autonomous research. While agent-based paradigms enhance efficiency, they often shift the manual burden rather than eliminating it, leaving scientists to perform the essential ``scaffold work'' for the agents.
 Furthermore, AI4S researchers often lack direct collaboration with researchers of AI foundation models. Although scientists frequently release datasets and benchmarks, it remains unclear whether these resources are directly applicable for training AI foundation models.

%in this work, we introduce a ten-level taxonomy (L0–L9) to organize data transformations. 
Based on the degree of automation, as shown in Figure~\ref{fig:paradigm}, we introduce a five-level taxonomy (L1-L5) to organize human-AI collaboration in AI4S.
%classify AI4S into five levels, from L1 to L5. 
L1 is ``Function-Level'', in which AI is invoked as a tool, and the human executes and closes the loop. L2 is ``Task-Level'', in which human decomposes and assigns research tasks, AI executes the decomposed sub-tasks. L3 is ``Collaborative-Level'', in which AI executes the primary research task, human collaborates and supervises. L4 is ``Guidance-Level'', in which AI provides expert-level services, human participates in key decisions. L5 is ``Autonomous-Level'', in which AI operates with full autonomy under human authorization, potentially exceeding human capabilities. 
%It is can be observed that the highest automation level achieved in most current research is L3, while only a very small number reach L4. 
It can be observed that most current research reaches at most L3 automation, with only a small fraction attaining L4. The current stage of AI4S can also be described as vibe research~\cite{zhang2026viberesearchingwolfcoming}.

\section{Discussion and Future Directions}
%\begin{quote}
%	\textit{``Practice is the sole criterion for testing AI.''\footnote{https://simple.wikipedia.org/wiki/Practice\_is\_the\_Sole\_Criterion\_for\_Testing\_Truth}}    
%	\hfill --- This Paper
%\end{quote}
%\begin{quote}
%	\textit{``Strive for progress, not perfection.''}    
%	\hfill --- David McCallum
%\end{quote}
%\begin{quote}
%	\textit{``Deep research may be the next scaling law of AI.''}    
%	\hfill --- Yipeng Yu
%\end{quote}
%\begin{quote}
%	\textit{``The best is the enemy of the good.''}    
%	\hfill --- Voltaire
%\end{quote}
\subsection{Key Challenges}
%porcupine
\paragraph{LLM and Harness} The knowledge and reasoning capacities of LLMs constitute the core basis of DR agents. However, LLMs can do some very complex things extremely well, yet fail at other tasks that seem simpler or closely related. ``Jagged Intelligence'' is a concept used to describe their uneven and unpredictable capabilities. Most relative research focuses on improving a DR agent's executive
capability, while enhancing its LLM's scientific taste remains largely underexplored~\cite{tong2026ailearnscientifictaste}. In addition, the architecture of DR agents warrants further investigation. Harness is an agentic architecture that allows multiple agents to work with shared context across different sessions and context windows. Building reliable harnesses for DR sometimes matters more than the LLMs.

\paragraph{Self-evolution} The cost-effectiveness of pre-training is facing diminishing returns now. Approaches like prompt engineering, context engineering, test-time scaling, SFT, and RL offer a limited performance ceiling for specific tasks. Consequently, current LLMs and agents lack the capacity for robust life-long learning~\cite{dupoux2026aisystemsdontlearn}. Unlike these models and agents, humans can continually learn from the environment through observation, interaction and feedback. Therefore, DR agents must be capable of self-evolution after deployment and able to learn autonomously or online within research environments.

\paragraph{From IDE to REE}
% insights
% Open Problems and Challenges
% Prospects
% Opportunities
% Vision
% relations to other hot research directions
A significant disparity exists between the research environments of human scientists and DR agents. Humans primarily conduct research in REE while utilizing IDE and SEE (see Figure~\ref{fig:overview}). In contrast, DR agents operate mostly in the IDE and occasionally in the SEE. These environmental differences present three primary challenges. First, foundation models for DR must develop better perception and reasoning for physical properties such as olfactory sensing and spatial cognition. Second, these agents require a broader set of tools to interact with REE. Third, DR agents need physical embodiment to move and perceive in the real world. This embodiment could be manifested as robotic systems or human proxies~\cite{geminiroboticsteam2025geminiroboticsbringingai}.

\subsection{Promising Avenues}
\begin{quote}
	\textit{``AI's next scaling law is AI itself.''}    
	\hfill --- This paper
\end{quote}
%find the right problems is more important than find the solution
%``Perhaps human knowledge is already sufficient to build a superhuman AI; it's just a matter of cleverly assembling the right pieces together.''
%history, bubble, false prosperity
%interaction is data, interaction is intelligence, next scaling law is research or coding? or interaction with the real world, or From carbon intelligence to silicon intelligence.
AI and AI4S are mutually reinforcing. Advances in AI can improve the quality and efficiency of scientific research (AI for Science). Progress in science can also advance AI (Science for AI). For example, DNNs were inspired by the human brain. Advances in physics and materials science can improve semiconductor manufacturing and chip design. In the context of Science for AI, researchers shift from passive data collection to active generation. This approach provides specific data designed for model training and structural reasoning. Data production thus prioritizes the needs of the model over the display of single experimental results.

The emergence of ChatGPT and subsequent advancements in LLMs suggest that AI has effectively passed the Turing Test~\cite{turing2007computing}. Current efforts in the field focus on reaching artificial general intelligence (AGI)~\cite{hendrycks2025definitionagi} or artificial superintelligence (ASI or SI). Beyond DR agents, here we briefly introduce three promising directions for AGI.

\paragraph{Agentic AI} DR agents and coding agents (such as \href{https://cursor.com/}{Cursor}, \href{https://openai.com/codex/}{Codex}, \href{https://github.com/anomalyco/opencode}{OpenCode}) are built for specific research and programming tasks. General agents serve a broader purpose. They aim to execute various tasks in unfamiliar settings. This capability does not require extensive domain-specific engineering. At present, there are two primary pathways to achieve agentic AI. The first is to incorporate agentic capabilities into LLMs, such as Claude-4, Gemini-3, GPT-5, and Kimi-2.5. The second is to build agent swarms on top of proprietary or open-weight LLMs, such as \href{https://github.com/openclaw/openclaw}{OpenClaw} and \href{https://manus.im/}{Manus}.
%\paragraph{Hyper Network} Information topological structure, manifold
%Agentic Coding: ~\cite{li2025deepcodeopenagenticcoding}
%\paragraph{Energh is Intelligence} xxx

\paragraph{Embodied AI} Embodied AI can be viewed as an advanced form of agentic AI. Its core components are ``world model'' and ``embodied agent''. World models, such as \href{https://marble.worldlabs.ai/}{Marble} and \href{https://3dgen.io/}{Genie 3}, can be regarded as a multimodal and multidimensional extension of LLMs~\cite{Hafner2025-kx,wang2025worldgentexttraversableinteractive,tong2026languagemodelingexplorationmultimodal,maes2026leworldmodelstableendtoendjointembedding}. Current world models typically use NTP for language and diffusion for vision, as shown in Figure~\ref{fig:gemini}(c). Their training data go beyond text and include videos, image-text pairs, and even action-conditioned videos. The model evolutionary path is roughly ``NTP (1D)$\rightarrow$Diffusion (2D)$\rightarrow$NeRF / Video Model (3D)$\rightarrow$World Model (4D)''. The embodied agent extends agents from the IDE to the REE. It is often built on top of the world models. The agent can explore and interact in a physics-based simulation (SEE)~\cite{simateam2025sima2generalistembodied}. It can also interact with the real world through live cameras and voice interfaces, smart glasses, autonomous cars, IP camera surveillance, or robots like \href{https://bostondynamics.com/products/atlas/}{Boston Atlas} and \href{https://www.unitree.com/}{Unitree robotics}~\cite{geminiroboticsteam2025geminiroboticsbringingai,Li2026vla}. This allows the embodied agents to learn directly through continuous interaction.
%World model, 
%~\cite{ha2018recurrent,ha2018world,lecun2022path,wang2025worldgentexttraversableinteractive}, Sora\footnote{https://openai.com/index/video-generation-models-as-world-simulators/}, Meta V-JEPA 2~\cite{assran2025vjepa2selfsupervisedvideo}, DeepMind Genie 3\footnote{https://deepmind.google/blog/genie-3-a-new-frontier-for-world-models/}, DeepMind SIMA 2~\cite{simateam2025sima2generalistembodied}
%\href{https://www.worldlabs.ai/blog/marble-world-model}{marble}
%vl-jepa~\cite{chen2025vljepajointembeddingpredictive}
%JEPA~\cite{huang2025llmjepalargelanguagemodels,balestriero2025lejepaprovablescalableselfsupervised}
%Embodied AI: ~\cite{wang2023voyageropenendedembodiedagent,driess2023palmeembodiedmultimodallanguage,simateam2025sima2generalistembodied}

\paragraph{Neuromorphic Intelligence} The term ``Intelligence'' is used rather than ``AI'' because biological brains play the primary role in this direction. There are two main branches of neuromorphic intelligence. One branch focuses on brain-mimetic models~\cite{1997Networks} and hardware~\cite{Pei2019bm}. These technologies derive intelligence from simulating biological neural architectures. The other branch is  ``Cyborg Intelligence''~\cite{mazesolving,cyborgyipeng}. This approach uses brain-computer interfaces (BCIs) to establish direct communication between biological brains and machines. This integration facilitates the fusion of biological and artificial intelligence. Within this framework, machines may handle rapid System 1 tasks while biological brains manage deliberate System 2 decision-making. Their roles are also interchangeable depending on the context.

\section{Conclusions}
\label{sec:conclusions}
This paper first provides a definition of deep research. We differentiate this concept from related terms for clarity. To help non-experts understand the core of AI, we track the technical evolution of deep research from the Transformer to agents. We then analyze AI4S across multiple disciplines. These include biology, materials, chemistry, medicine, mathematics, physics, meteorology, and other disciplines. The specific roles and impacts of AI in each field are clarified. AI can advance science, and science can in turn inform AI. Finally, this paper summarizes the core challenges facing deep research. We also propose three promising research directions for achieving AGI.

\section*{Limitations}
Open-weight models and commercial closed-source LLMs are continuously and rapidly evolving. Therefore, this paper reflects only the state of these models at the time of publication. 
%In addition, the author is an AI researcher and may not fully grasp the current key research questions and progress in all other scientific disciplines. As a result, the investigation of AI for science in this paper may be not fully comprehensive.
The authors carefully reviewed all included papers. This applies to preprints from arXiv and bioRxiv as well. However, the authors primarily specialize in AI research. Consequently, the scope of the investigation into various AI for Science fields might be limited.
%the investigation of diverse AI for Science domains might not be fully comprehensive.
%\section*{Ethical Considerations}
%
%xxx

% \section*{Acknowledgments}
% This document has been adapted
% by Steven Bethard, Ryan Cotterell and Rui Yan
% from the instructions for earlier ACL and NAACL proceedings, including those for
% A

\clearpage

\bibliography{ref}

@article{ghafarollahi2024sciagents,
  title={SciAgents: Automating Scientific Discovery Through Bioinspired Multi-Agent Intelligent Graph Reasoning},
  author={Ghafarollahi, Alireza and Buehler, Markus J},
  journal={Advanced Materials},
  pages={2413523},
  year={2024}
}

@article{buehler2024graphreasoning,
	author={Markus J. Buehler},
	title={Accelerating Scientific Discovery with Generative Knowledge Extraction, Graph-Based Representation, and Multimodal Intelligent Graph Reasoning},
	journal={Machine Learning: Science and Technology},
	year={2024}
}

@inproceedings{zheng2024openresearcherunleashingaiaccelerated,
	title = "{O}pen{R}esearcher: Unleashing {AI} for Accelerated Scientific Research",
	author = "Zheng, Yuxiang  and
	Sun, Shichao  and
	Qiu, Lin  and
	Ru, Dongyu  and
	Jiayang, Cheng  and
	Li, Xuefeng  and
	Lin, Jifan  and
	Wang, Binjie  and
	Luo, Yun  and
	Pan, Renjie  and
	Xu, Yang  and
	Min, Qingkai  and
	Zhang, Zizhao  and
	Wang, Yiwen  and
	Li, Wenjie  and
	Liu, Pengfei",
	booktitle = "Proceedings of the 2024 Conference on Empirical Methods in Natural Language Processing: System Demonstrations",
	month = nov,
	year = "2024",
	pages = "209--218"
}

@inproceedings{li2024chainideasrevolutionizingresearch,
	title = "Chain of Ideas: Revolutionizing Research Via Novel Idea Development with {LLM} Agents",
	author = "Li, Long  and
	Xu, Weiwen  and
	Guo, Jiayan  and
	Zhao, Ruochen  and
	Li, Xingxuan  and
	Yuan, Yuqian  and
	Zhang, Boqiang  and
	Jiang, Yuming  and
	Xin, Yifei  and
	Dang, Ronghao  and
	Rong, Yu  and
	Zhao, Deli  and
	Feng, Tian  and
	Bing, Lidong",
	booktitle = "Findings of the Association for Computational Linguistics: EMNLP 2025",
	month = nov,
	year = "2025",
	pages = "8971--9004"
}

@article{mo2024surveyconversationalsearch,
	author = {Mo, Fengran and Mao, Kelong and Zhao, Ziliang and Qian, Hongjin and Chen, Haonan and Cheng, Yiruo and Li, Xiaoxi and Zhu, Yutao and Dou, Zhicheng and Nie, Jian-Yun},
	title = {A Survey of Conversational Search},
	year = {2025},
	issue_date = {November 2025},
	volume = {43},
	number = {6},
	journal = {ACM Trans. Inf. Syst.},
	month = sep,
	articleno = {167},
	numpages = {50}
}

@inproceedings{wu2025agenticreasoningreasoningllms,
	title = "Agentic Reasoning: A Streamlined Framework for Enhancing {LLM} Reasoning with Agentic Tools",
	author = "Wu, Junde  and
	Zhu, Jiayuan  and
	Liu, Yuyuan  and
	Xu, Min  and
	Jin, Yueming",
	booktitle = "Proceedings of the 63rd Annual Meeting of the Association for Computational Linguistics (Volume 1: Long Papers)",
	month = jul,
	year = "2025",
	pages = "28489--28503"
}

@misc{gottweis2025aicoscientist,
      title={Towards an AI co-scientist}, 
      author={Juraj Gottweis and Wei-Hung Weng and Alexander Daryin and Tao Tu and Anil Palepu and Petar Sirkovic and Artiom Myaskovsky and Felix Weissenberger and Keran Rong and Ryutaro Tanno and Khaled Saab and Dan Popovici and Jacob Blum and Fan Zhang and Katherine Chou and Avinatan Hassidim and Burak Gokturk and Amin Vahdat and Pushmeet Kohli and Yossi Matias and Andrew Carroll and Kavita Kulkarni and Nenad Tomasev and Yuan Guan and Vikram Dhillon and Eeshit Dhaval Vaishnav and Byron Lee and Tiago R D Costa and José R Penadés and Gary Peltz and Yunhan Xu and Annalisa Pawlosky and Alan Karthikesalingam and Vivek Natarajan},
      year={2025},
      eprint={2502.18864},
      archivePrefix={arXiv},
      primaryClass={cs.AI},
      url={https://arxiv.org/abs/2502.18864}, 
}

@misc{liu2025researchbenchbenchmarkingllmsscientific,
      title={ResearchBench: Benchmarking LLMs in Scientific Discovery via Inspiration-Based Task Decomposition}, 
      author={Yujie Liu and Zonglin Yang and Tong Xie and Jinjie Ni and Ben Gao and Yuqiang Li and Shixiang Tang and Wanli Ouyang and Erik Cambria and Dongzhan Zhou},
      year={2025},
      eprint={2503.21248},
      archivePrefix={arXiv},
      primaryClass={cs.CL},
      url={https://arxiv.org/abs/2503.21248}, 
}

@misc{ren2025scientificintelligencesurveyllmbased,
      title={Towards Scientific Intelligence: A Survey of LLM-based Scientific Agents}, 
      author={Shuo Ren and Pu Jian and Zhenjiang Ren and Chunlin Leng and Can Xie and Jiajun Zhang},
      year={2025},
      eprint={2503.24047},
      archivePrefix={arXiv},
      primaryClass={cs.AI}
}

@inproceedings{
	starace2025paperbenchevaluatingaisability,
	title={PaperBench: Evaluating {AI}{\textquoteright}s Ability to Replicate {AI} Research},
	author={Giulio Starace and Oliver Jaffe and Dane Sherburn and James Aung and Jun Shern Chan and Leon Maksin and Rachel Dias and Evan Mays and Benjamin Kinsella and Wyatt Thompson and Johannes Heidecke and Amelia Glaese and Tejal Patwardhan},
	booktitle={Forty-second International Conference on Machine Learning},
	year={2025}
}

@inproceedings{zheng-etal-2025-deepresearcher,
	title = "{D}eep{R}esearcher: Scaling Deep Research via Reinforcement Learning in Real-world Environments",
	author = "Zheng, Yuxiang  and
	Fu, Dayuan  and
	Hu, Xiangkun  and
	Cai, Xiaojie  and
	Ye, Lyumanshan  and
	Lu, Pengrui  and
	Liu, Pengfei",
	booktitle = "Proceedings of the 2025 Conference on Empirical Methods in Natural Language Processing",
	month = nov,
	year = "2025",
	pages = "414--431"
}

@misc{li2025arxivbenchllmsassistresearchers,
      title={ArxivBench: Can LLMs Assist Researchers in Conducting Research?}, 
      author={Ning Li and Jingran Zhang and Justin Cui},
      year={2025},
      eprint={2504.10496},
      archivePrefix={arXiv},
      primaryClass={cs.IR},
      url={https://arxiv.org/abs/2504.10496}, 
}

@misc{fang2025ainewtonconceptdrivenphysicallaw,
      title={AI-Newton: A Concept-Driven Physical Law Discovery System without Prior Physical Knowledge}, 
      author={You-Le Fang and Dong-Shan Jian and Xiang Li and Yan-Qing Ma},
      year={2025},
      eprint={2504.01538},
      archivePrefix={arXiv},
      primaryClass={cs.AI}
}

@inproceedings{
	qiu2025phybenchholisticevaluationphysical,
	title={{PHYB}ench: Holistic Evaluation of Physical Perception and Reasoning in Large Language Models},
	author={Shi Qiu and Shaoyang Guo and Zhuo-Yang Song and Yunbo Sun and Zeyu Cai and Jiashen Wei and Tianyu Luo and Yixuan Yin and Zhang Haoxu and Yi Hu and Chenyang Wang and Chencheng Tang and Haoling Chang and Qi Liu and Ziheng Zhou and Tianyu Zhang and Jingtian Zhang and Zhangyi Liu and Minghao Li and Yuku Zhang and Boxuan Jing and Xianqi Yin and Yutong Ren and Zizhuo Fu and Jiaming Ji and Weike Wang and Xudong Tian and Anqi Lv and Laifu Man and Jianxiang Li and Feiyu Tao and Qihua Sun and Zhou Liang and Yushu Mu and Zhongxuan Li and Jing-Jun Zhang and Shutao Zhang and Xiaotian Li and Xingqi Xia and Jiawei Lin and Zheyu Shen and Jiahang Chen and Qiuhao Xiong and Binran Wang and Fengyuan Wang and Niziyang and Bohan Zhang and Fan Cui and shaochangkun and Qing-Hong Cao and Ming-xing Luo and Muhan Zhang and Hua Xing Zhu},
	booktitle={The Thirty-ninth Annual Conference on Neural Information Processing Systems Datasets and Benchmarks Track},
	year={2025}
}

@inproceedings{
	li2025webthinkerempoweringlargereasoning,
	title={WebThinker: Empowering Large Reasoning Models with Deep Research Capability},
	author={Xiaoxi Li and Jiajie Jin and Guanting Dong and Hongjin Qian and Yongkang Wu and Ji-Rong Wen and Yutao Zhu and Zhicheng Dou},
	booktitle={The Thirty-ninth Annual Conference on Neural Information Processing Systems},
	year={2025}
}

@inproceedings{
	sun2025p2pautomatedpapertopostergeneration,
	title={P2P: Automated Paper-to-Poster Generation and Fine-Grained Benchmark},
	author={Tao Sun and Enhao Pan and Zhengkai Yang and Kaixin Sui and Jiajun Shi and Xianfu Cheng and Tongliang Li and Ge Zhang and Wenhao Huang and Jian Yang and Zhoujun Li},
	booktitle={The Fourteenth International Conference on Learning Representations},
	year={2026}
}

@inproceedings{
	airesearcher,
	title={{AI}-Researcher: Autonomous Scientific Innovation},
	author={Jiabin Tang and Lianghao Xia and Zhonghang Li and Chao Huang},
	booktitle={The Thirty-ninth Annual Conference on Neural Information Processing Systems},
	year={2025},
	url={https://openreview.net/forum?id=kQWyOYUAC4}
}

@inproceedings{
	chen2025mlrbenchevaluatingaiagents,
	title={{MLR}-Bench: Evaluating {AI} Agents on Open-Ended Machine Learning Research},
	author={Hui Chen and Miao Xiong and Yujie Lu and Wei Han and Ailin Deng and Yufei He and Jiaying Wu and Yibo Li and Yue Liu and Bryan Hooi},
	booktitle={The Thirty-ninth Annual Conference on Neural Information Processing Systems Datasets and Benchmarks Track},
	year={2025}
}

@inproceedings{
	wu2025webdancerautonomousinformationseeking,
	title={WebDancer: Towards Autonomous Information Seeking Agency},
	author={Jialong Wu and Baixuan Li and Runnan Fang and Wenbiao Yin and Liwen Zhang and Zhenglin Wang and Zhengwei Tao and Ding-Chu Zhang and Zekun Xi and Xiangru Tang and Yong Jiang and Pengjun Xie and Fei Huang and Jingren Zhou},
	booktitle={The Thirty-ninth Annual Conference on Neural Information Processing Systems},
	year={2025}
}

@inproceedings{FlashRAG_WWW,
author = {Jin, Jiajie and Zhu, Yutao and Dou, Zhicheng and Dong, Guanting and Yang, Xinyu and Zhang, Chenghao and Zhao, Tong and Yang, Zhao and Wen, Ji-Rong},
title = {FlashRAG: A Modular Toolkit for Efficient Retrieval-Augmented Generation Research},
year = {2025},
booktitle = {Companion Proceedings of the ACM on Web Conference 2025},
pages = {737–740},
numpages = {4},
series = {WWW '25}
}

@misc{lyu2025deepshopbenchmarkdeepresearch,
      title={DeepShop: A Benchmark for Deep Research Shopping Agents}, 
      author={Yougang Lyu and Xiaoyu Zhang and Lingyong Yan and Maarten de Rijke and Zhaochun Ren and Xiuying Chen},
      year={2025},
      eprint={2506.02839},
      archivePrefix={arXiv},
      primaryClass={cs.IR}
}

@misc{xu2025comprehensivesurveydeepresearch,
      title={A Comprehensive Survey of Deep Research: Systems, Methodologies, and Applications}, 
      author={Renjun Xu and Jingwen Peng},
      year={2025},
      eprint={2506.12594},
      archivePrefix={arXiv},
      primaryClass={cs.AI},
      url={https://arxiv.org/abs/2506.12594}, 
}

@misc{huang2025deepresearchagentssystematic,
      title={Deep Research Agents: A Systematic Examination And Roadmap}, 
      author={Yuxuan Huang and Yihang Chen and Haozheng Zhang and Kang Li and Meng Fang and Linyi Yang and Xiaoguang Li and Lifeng Shang and Songcen Xu and Jianye Hao and Kun Shao and Jun Wang},
      year={2025},
      eprint={2506.18096},
      archivePrefix={arXiv},
      primaryClass={cs.AI}
}

@misc{chai2025scimastergeneralpurposescientificai,
      title={SciMaster: Towards General-Purpose Scientific AI Agents, Part I. X-Master as Foundation: Can We Lead on Humanity's Last Exam?}, 
      author={Jingyi Chai and Shuo Tang and Rui Ye and Yuwen Du and Xinyu Zhu and Mengcheng Zhou and Yanfeng Wang and Weinan E and Yuzhi Zhang and Linfeng Zhang and Siheng Chen},
      year={2025},
      eprint={2507.05241},
      archivePrefix={arXiv},
      primaryClass={cs.AI}
}

@misc{han2025deepresearchertesttimediffusion,
      title={Deep Researcher with Test-Time Diffusion}, 
      author={Rujun Han and Yanfei Chen and Zoey CuiZhu and Lesly Miculicich and Guan Sun and Yuanjun Bi and Weiming Wen and Hui Wan and Chunfeng Wen and Solène Maître and George Lee and Vishy Tirumalashetty and Emily Xue and Zizhao Zhang and Salem Haykal and Burak Gokturk and Tomas Pfister and Chen-Yu Lee},
      year={2025},
      eprint={2507.16075},
      archivePrefix={arXiv},
      primaryClass={cs.CL}
}

@article{swanson2025virtual,
  title={The Virtual Lab of AI agents designs new SARS-CoV-2 nanobodies},
  author={Swanson, Kyle and Wu, Wesley and Bulaong, Nash L and Pak, John E and Zou, James},
  journal={Nature},
  volume = {646},
  pages={716--723},
  year={2025}
}

@inproceedings{
	java2025characterizingdeepresearchbenchmark,
	title={Characterizing Deep Research: A Benchmark and Formal Definition},
	author={Abhinav Java and Ashmit Khandelwal and Sukruta Prakash Midigeshi and Aaron Halfaker and Amit Deshpande and Navin Goyal and Ankur Gupta and Nagarajan Natarajan and Amit Sharma},
	booktitle={The Fourteenth International Conference on Learning Representations},
	year={2026},
	url={https://openreview.net/forum?id=5EmpOCq1Ql}
}

@misc{xi2025surveyllmbaseddeepsearch,
      title={A Survey of LLM-based Deep Search Agents: Paradigm, Optimization, Evaluation, and Challenges}, 
      author={Yunjia Xi and Jianghao Lin and Yongzhao Xiao and Zheli Zhou and Rong Shan and Te Gao and Jiachen Zhu and Weiwen Liu and Yong Yu and Weinan Zhang},
      year={2025},
      eprint={2508.05668},
      archivePrefix={arXiv},
      primaryClass={cs.IR},
      url={https://arxiv.org/abs/2508.05668}, 
}

@inproceedings{
	geng2025webwatcherbreakingnewfrontier,
	title={WebWatcher: Breaking New Frontiers of Vision-Language Deep Research Agent},
	author={Xinyu Geng and Peng Xia and Zhen Zhang and Xinyu Wang and Qiuchen Wang and Ruixue Ding and Chenxi Wang and Jialong Wu and Kuan Li and Yida Zhao and Huifeng Yin and Yong Jiang and Pengjun Xie and Fei Huang and Huaxiu Yao and Yi R. Fung and Jingren Zhou},
	booktitle={The Fourteenth International Conference on Learning Representations},
	year={2026}
}

@misc{zhang2025deepresearchsurveyautonomous,
      title={Deep Research: A Survey of Autonomous Research Agents}, 
      author={Wenlin Zhang and Xiaopeng Li and Yingyi Zhang and Pengyue Jia and Yichao Wang and Huifeng Guo and Yong Liu and Xiangyu Zhao},
      year={2025},
      eprint={2508.12752},
      archivePrefix={arXiv},
      primaryClass={cs.IR}
}

@misc{deng2025atomsearcherenhancingagenticdeep,
      title={Atom-Searcher: Enhancing Agentic Deep Research via Fine-Grained Atomic Thought Reward}, 
      author={Yong Deng and Guoqing Wang and Zhenzhe Ying and Xiaofeng Wu and Jinzhen Lin and Wenwen Xiong and Yuqin Dai and Shuo Yang and Zhanwei Zhang and Qiwen Wang and Yang Qin and Yuan Wang and Quanxing Zha and Sunhao Dai and Changhua Meng},
      year={2025},
      eprint={2508.12800},
      archivePrefix={arXiv},
      primaryClass={cs.CL}
}

@misc{hu2025surveyscientificlargelanguage,
      title={A Survey of Scientific Large Language Models: From Data Foundations to Agent Frontiers}, 
      author={Ming Hu and Chenglong Ma and Wei Li and Wanghan Xu and Jiamin Wu and Jucheng Hu and Tianbin Li and Guohang Zhuang and Jiaqi Liu and Yingzhou Lu and Ying Chen and Chaoyang Zhang and Cheng Tan and Jie Ying and Guocheng Wu and Shujian Gao and Pengcheng Chen and Jiashi Lin and Haitao Wu and Lulu Chen and Fengxiang Wang and Yuanyuan Zhang and Xiangyu Zhao and Feilong Tang and Encheng Su and Junzhi Ning and Xinyao Liu and Ye Du and Changkai Ji and Cheng Tang and Huihui Xu and Ziyang Chen and Ziyan Huang and Jiyao Liu and Pengfei Jiang and Yizhou Wang and Chen Tang and Jianyu Wu and Yuchen Ren and Siyuan Yan and Zhonghua Wang and Zhongxing Xu and Shiyan Su and Shangquan Sun and Runkai Zhao and Zhisheng Zhang and Yu Liu and Fudi Wang and Yuanfeng Ji and Yanzhou Su and Hongming Shan and Chunmei Feng and Jiahao Xu and Jiangtao Yan and Wenhao Tang and Diping Song and Lihao Liu and Yanyan Huang and Lequan Yu and Bin Fu and Shujun Wang and Xiaomeng Li and Xiaowei Hu and Yun Gu and Ben Fei and Zhongying Deng and Benyou Wang and Yuewen Cao and Minjie Shen and Haodong Duan and Jie Xu and Yirong Chen and Fang Yan and Hongxia Hao and Jielan Li and Jiajun Du and Yanbo Wang and Imran Razzak and Chi Zhang and Lijun Wu and Conghui He and Zhaohui Lu and Jinhai Huang and Yihao Liu and Fenghua Ling and Yuqiang Li and Aoran Wang and Qihao Zheng and Nanqing Dong and Tianfan Fu and Dongzhan Zhou and Yan Lu and Wenlong Zhang and Jin Ye and Jianfei Cai and Wanli Ouyang and Yu Qiao and Zongyuan Ge and Shixiang Tang and Junjun He and Chunfeng Song and Lei Bai and Bowen Zhou},
      year={2025},
      eprint={2508.21148},
      archivePrefix={arXiv},
      primaryClass={cs.CL},
      url={https://arxiv.org/abs/2508.21148}, 
}

@misc{nguyen2025sfrdeepresearcheffectivereinforcementlearning,
      title={SFR-DeepResearch: Towards Effective Reinforcement Learning for Autonomously Reasoning Single Agents}, 
      author={Xuan-Phi Nguyen and Shrey Pandit and Revanth Gangi Reddy and Austin Xu and Silvio Savarese and Caiming Xiong and Shafiq Joty},
      year={2025},
      eprint={2509.06283},
      archivePrefix={arXiv},
      primaryClass={cs.AI}
}

@misc{li2025reinforcementlearningfoundationsdeep,
      title={Reinforcement Learning Foundations for Deep Research Systems: A Survey}, 
      author={Wenjun Li and Zhi Chen and Jingru Lin and Hannan Cao and Wei Han and Sheng Liang and Zhi Zhang and Kuicai Dong and Dexun Li and Chen Zhang and Yong Liu},
      year={2025},
      eprint={2509.06733},
      archivePrefix={arXiv},
      primaryClass={cs.AI}
}

@misc{lu2025deepdiveadvancingdeepsearch,
      title={DeepDive: Advancing Deep Search Agents with Knowledge Graphs and Multi-Turn RL}, 
      author={Rui Lu and Zhenyu Hou and Zihan Wang and Hanchen Zhang and Xiao Liu and Yujiang Li and Shi Feng and Jie Tang and Yuxiao Dong},
      year={2025},
      eprint={2509.10446},
      archivePrefix={arXiv},
      primaryClass={cs.CL}
}

@misc{qiao2025webresearcherunleashingunboundedreasoning,
      title={WebResearcher: Unleashing unbounded reasoning capability in Long-Horizon Agents}, 
      author={Zile Qiao and Guoxin Chen and Xuanzhong Chen and Donglei Yu and Wenbiao Yin and Xinyu Wang and Zhen Zhang and Baixuan Li and Huifeng Yin and Kuan Li and Rui Min and Minpeng Liao and Yong Jiang and Pengjun Xie and Fei Huang and Jingren Zhou},
      year={2025},
      eprint={2509.13309},
      archivePrefix={arXiv},
      primaryClass={cs.CL}
}

@inproceedings{
	li2025webweaverstructuringwebscaleevidence,
	title={WebWeaver: Structuring Web-Scale Evidence with Dynamic Outlines for Open-Ended Deep Research},
	author={Zijian Li and Xin Guan and Bo Zhang and Shen Huang and Houquan Zhou and Shaopeng Lai and Ming Yan and Yong Jiang and Pengjun Xie and Fei Huang and Jun Zhang and Jingren Zhou},
	booktitle={The Fourteenth International Conference on Learning Representations},
	year={2026}
}

@inproceedings{
	wu2025deepsearchovercomebottleneckreinforcement,
	title={DeepSearch: Overcome the Bottleneck of Reinforcement Learning with Verifiable Rewards via Monte Carlo Tree Search},
	author={Fang Wu and Weihao Xuan and Heli Qi and Ximing Lu and Aaron Tu and Li Erran Li and Yejin Choi},
	booktitle={The Fourteenth International Conference on Learning Representations},
	year={2026}
}

@misc{yao2025rigorousbenchmarkmultidimensionalevaluation,
      title={A Rigorous Benchmark with Multidimensional Evaluation for Deep Research Agents: From Answers to Reports}, 
      author={Yang Yao and Yixu Wang and Yuxuan Zhang and Yi Lu and Tianle Gu and Lingyu Li and Dingyi Zhao and Keming Wu and Haozhe Wang and Ping Nie and Yan Teng and Yingchun Wang},
      year={2025},
      eprint={2510.02190},
      archivePrefix={arXiv},
      primaryClass={cs.AI}
}

@misc{chen2025marsoptimizingdualsystemdeep,
      title={MARS: Optimizing Dual-System Deep Research via Multi-Agent Reinforcement Learning}, 
      author={Guoxin Chen and Zile Qiao and Wenqing Wang and Donglei Yu and Xuanzhong Chen and Hao Sun and Minpeng Liao and Kai Fan and Yong Jiang and Penguin Xie and Wayne Xin Zhao and Ruihua Song and Fei Huang},
      year={2025},
      eprint={2510.04935},
      archivePrefix={arXiv},
      primaryClass={cs.AI}
}

@misc{hu2025flowsearchadvancingdeepresearch,
      title={FlowSearch: Advancing deep research with dynamic structured knowledge flow}, 
      author={Yusong Hu and Runmin Ma and Yue Fan and Jinxin Shi and Zongsheng Cao and Yuhao Zhou and Jiakang Yuan and Xiangchao Yan and Wenlong Zhang and Lei Bai and Bo Zhang},
      year={2025},
      eprint={2510.08521},
      archivePrefix={arXiv},
      primaryClass={cs.AI}
}

@inproceedings{
	abaskohi2025drbenchrealisticbenchmarkenterprise,
	title={{DRB}ench: A Realistic Benchmark for Enterprise Deep Research},
	author={Amirhossein Abaskohi and Tianyi Chen and Miguel Mu{\~n}oz-M{\'a}rmol and Curtis Fox and Amrutha Varshini Ramesh and {\'E}tienne Marcotte and Xing Han L{\`u} and Nicolas Chapados and Spandana Gella and Christopher Pal and Alexandre Drouin and Issam H. Laradji},
	booktitle={The Fourteenth International Conference on Learning Representations},
	year={2026},
	url={https://openreview.net/forum?id=IGYQ4c92e2}
}

@inproceedings{
	wang2025liveresearchbenchlivebenchmarkusercentric,
	title={LiveResearchBench: Benchmarking Single- and Multi-Agent Systems for Citation-Grounded Deep Research},
	author={Jiayu Wang and Yifei Ming and Riya Dulepet and Qinglin Chen and Austin Xu and Zixuan Ke and Frederic Sala and Aws Albarghouthi and Caiming Xiong and Shafiq Joty},
	booktitle={The Fourteenth International Conference on Learning Representations},
	year={2026}
}

@misc{wang2025exploreevolvescalingevolved,
      title={Explore to Evolve: Scaling Evolved Aggregation Logic via Proactive Online Exploration for Deep Research Agents}, 
      author={Rui Wang and Ce Zhang and Jun-Yu Ma and Jianshu Zhang and Hongru Wang and Yi Chen and Boyang Xue and Tianqing Fang and Zhisong Zhang and Hongming Zhang and Haitao Mi and Dong Yu and Kam-Fai Wong},
      year={2025},
      eprint={2510.14438},
      archivePrefix={arXiv},
      primaryClass={cs.CL},
      url={https://arxiv.org/abs/2510.14438}, 
}

@misc{cong2025labosaixrcoscientistsees,
      title={LabOS: The AI-XR Co-Scientist That Sees and Works With Humans}, 
      author={Le Cong and Zaixi Zhang and Xiaotong Wang and Yin Di and Ruofan Jin and Michal Gerasimiuk and Yinkai Wang and Ravi K. Dinesh and David Smerkous and Alex Smerkous and Xuekun Wu and Shilong Liu and Peishan Li and Yi Zhu and Simran Serrao and Ning Zhao and Imran A. Mohammad and John B. Sunwoo and Joseph C. Wu and Mengdi Wang},
      year={2025},
      eprint={2510.14861},
      archivePrefix={arXiv},
      primaryClass={cs.AI}
}

@inproceedings{yang2025researstudiohumanintervenableframeworkbuilding,
	title = "{R}esear{S}tudio: A Human-intervenable Framework for Building Controllable Deep Research Agents",
	author = "Yang, Linyi  and
	Weng, Yixuan",
	booktitle = "Proceedings of the 2025 Conference on Empirical Methods in Natural Language Processing: System Demonstrations",
	month = nov,
	year = "2025",
	pages = "896--905"
}

@misc{wan2025pokeeresearcheffectivedeepresearch,
      title={PokeeResearch: Effective Deep Research via Reinforcement Learning from AI Feedback and Robust Reasoning Scaffold}, 
      author={Yi Wan and Jiuqi Wang and Liam Li and Jinsong Liu and Ruihao Zhu and Zheqing Zhu},
      year={2025},
      eprint={2510.15862},
      archivePrefix={arXiv},
      primaryClass={cs.AI}
}

@misc{zou2025realdeepresearchai,
      title={Real Deep Research for AI, Robotics and Beyond}, 
      author={Xueyan Zou and Jianglong Ye and Hao Zhang and Xiaoyu Xiang and Mingyu Ding and Zhaojing Yang and Yong Jae Lee and Zhuowen Tu and Sifei Liu and Xiaolong Wang},
      year={2025},
      eprint={2510.20809},
      archivePrefix={arXiv},
      primaryClass={cs.AI}
}

@misc{li2025inverseknowledgesearchverifiable,
      title={Inverse Knowledge Search over Verifiable Reasoning: Synthesizing a Scientific Encyclopedia from a Long Chains-of-Thought Knowledge Base}, 
      author={Yu Li and Yuan Huang and Tao Wang and Caiyu Fan and Xiansheng Cai and Sihan Hu and Xinzijian Liu and Cheng Shi and Mingjun Xu and Zhen Wang and Yan Wang and Xiangqi Jin and Tianhan Zhang and Linfeng Zhang and Lei Wang and Youjin Deng and Pan Zhang and Weijie Sun and Xingyu Li and Weinan E and Linfeng Zhang and Zhiyuan Yao and Kun Chen},
      year={2025},
      eprint={2510.26854},
      archivePrefix={arXiv},
      primaryClass={cs.AI}
}

@article{ding2025scitoolagent,
  title={SciToolAgent: a knowledge-graph-driven scientific agent for multitool integration},
  author={Ding, Keyan and Yu, Jing and Huang, Junjie and Yang, Yuchen and Zhang, Qiang and Chen, Huajun},
  journal={Nature Computational Science},
  pages={1--11},
  year={2025}
}

@misc{villaescusanavarro2025denarioprojectdeepknowledge,
      title={The Denario project: Deep knowledge AI agents for scientific discovery}, 
      author={Francisco Villaescusa-Navarro and Boris Bolliet and Pablo Villanueva-Domingo and Adrian E. Bayer and Aidan Acquah and Chetana Amancharla and Almog Barzilay-Siegal and Pablo Bermejo and Camille Bilodeau and Pablo Cárdenas Ramírez and Miles Cranmer and Urbano L. França and ChangHoon Hahn and Yan-Fei Jiang and Raul Jimenez and Jun-Young Lee and Antonio Lerario and Osman Mamun and Thomas Meier and Anupam A. Ojha and Pavlos Protopapas and Shimanto Roy and David N. Spergel and Pedro Tarancón-Álvarez and Ujjwal Tiwari and Matteo Viel and Digvijay Wadekar and Chi Wang and Bonny Y. Wang and Licong Xu and Yossi Yovel and Shuwen Yue and Wen-Han Zhou and Qiyao Zhu and Jiajun Zou and Íñigo Zubeldia},
      year={2025},
      eprint={2510.26887},
      archivePrefix={arXiv},
      primaryClass={cs.AI}
}

@misc{fan2025deepplannerscalingplanningcapability,
      title={DeepPlanner: Scaling Planning Capability for Deep Research Agents via Advantage Shaping}, 
      author={Wei Fan and Wenlin Yao and Zheng Li and Feng Yao and Xin Liu and Liang Qiu and Qingyu Yin and Yangqiu Song and Bing Yin},
      year={2025},
      eprint={2510.12979},
      archivePrefix={arXiv},
      primaryClass={cs.AI}
}

@inproceedings{
    gou2025mindweb,
    title={Mind2Web 2: Evaluating Agentic Search with Agent-as-a-Judge},
    author={Boyu Gou and Zanming Huang and Yuting Ning and Yu Gu and Michael Lin and Botao Yu and Andrei Kopanev and Weijian Qi and Yiheng Shu and Jiaman Wu and Chan Hee Song and Bernal Jimenez Gutierrez and Yifei Li and Zeyi Liao and Hanane Nour Moussa and TIANSHU ZHANG and Jian Xie and Tianci Xue and Shijie Chen and Boyuan Zheng and Kai Zhang and Zhaowei Cai and Viktor Rozgic and Morteza Ziyadi and Huan Sun and Yu Su},
    booktitle={The Thirty-ninth Annual Conference on Neural Information Processing Systems Datasets and Benchmarks Track},
    year={2025}
}

@misc{mitchener2025kosmosaiscientistautonomous,
      title={Kosmos: An AI Scientist for Autonomous Discovery}, 
      author={Ludovico Mitchener and Angela Yiu and Benjamin Chang and Mathieu Bourdenx and Tyler Nadolski and Arvis Sulovari and Eric C. Landsness and Daniel L. Barabasi and Siddharth Narayanan and Nicky Evans and Shriya Reddy and Martha Foiani and Aizad Kamal and Leah P. Shriver and Fang Cao and Asmamaw T. Wassie and Jon M. Laurent and Edwin Melville-Green and Mayk Caldas and Albert Bou and Kaleigh F. Roberts and Sladjana Zagorac and Timothy C. Orr and Miranda E. Orr and Kevin J. Zwezdaryk and Ali E. Ghareeb and Laurie McCoy and Bruna Gomes and Euan A. Ashley and Karen E. Duff and Tonio Buonassisi and Tom Rainforth and Randall J. Bateman and Michael Skarlinski and Samuel G. Rodriques and Michaela M. Hinks and Andrew D. White},
      year={2025},
      eprint={2511.02824},
      archivePrefix={arXiv},
      primaryClass={cs.AI}
}

@inproceedings{
	anonymous2025iterresearch,
	title={IterResearch: Rethinking Long-Horizon Agents via Markovian State Reconstruction},
	author={Guoxin Chen and Zile Qiao and Xuanzhong Chen and Donglei Yu and Haotian Xu and Xin Zhao and Ruihua Song and Wenbiao Yin and Huifeng Yin and Liwen Zhang and Kuan Li and Minpeng Liao and Yong Jiang and Pengjun Xie and Fei Huang and Jingren Zhou},
	booktitle={The Fourteenth International Conference on Learning Representations},
	year={2026}
}

@misc{li2025sciagentunifiedmultiagentgeneralistic,
      title={SciAgent: A Unified Multi-Agent System for Generalistic Scientific Reasoning}, 
      author={Xuchen Li and Ruitao Wu and Xuanbo Liu and Xukai Wang and Jinbo Hu and Zhixin Bai and Bohan Zeng and Hao Liang and Leheng Chen and Mingrui Chen and Haitian Zhong and Xuanlin Yang and Xu-Yao Zhang and Liu Liu and Jia Li and Kaiqi Huang and Jiahao Xu and Haitao Mi and Wentao Zhang and Bin Dong},
      year={2025},
      eprint={2511.08151},
      archivePrefix={arXiv},
      primaryClass={cs.AI}
}

@misc{pu2025piflowprincipleawarescientificdiscovery,
      title={PiFlow: Principle-aware Scientific Discovery with Multi-Agent Collaboration}, 
      author={Yingming Pu and Tao Lin and Hongyu Chen},
      year={2025},
      eprint={2505.15047},
      archivePrefix={arXiv},
      primaryClass={cs.LG}
}

@misc{lin2025comprehensivesurveyreinforcementlearningbased,
      title={A Comprehensive Survey on Reinforcement Learning-based Agentic Search: Foundations, Roles, Optimizations, Evaluations, and Applications}, 
      author={Minhua Lin and Zongyu Wu and Zhichao Xu and Hui Liu and Xianfeng Tang and Qi He and Charu Aggarwal and Hui Liu and Xiang Zhang and Suhang Wang},
      year={2025},
      eprint={2510.16724},
      archivePrefix={arXiv},
      primaryClass={cs.AI},
      url={https://arxiv.org/abs/2510.16724}, 
}

@inproceedings{ying-etal-2025-seedbench,
    title = "{S}eed{B}ench: A Multi-task Benchmark for Evaluating Large Language Models in Seed Science",
    author = "Ying, Jie  and
      Chen, Zihong  and
      Wang, Zhefan  and
      Jiang, Wanli  and
      Wang, Chenyang  and
      Yuan, Zhonghang  and
      Su, Haoyang  and
      Kong, Huanjun  and
      Yang, Fan  and
      Dong, Nanqing",
    booktitle = "Proceedings of the 63rd Annual Meeting of the Association for Computational Linguistics (Volume 1: Long Papers)",
    month = jul,
    year = "2025",
    pages = "31395--31449"
}

@inproceedings{su-etal-2025-many,
    title = "Many Heads Are Better Than One: Improved Scientific Idea Generation by A {LLM}-Based Multi-Agent System",
    author = "Su, Haoyang  and
      Chen, Renqi  and
      Tang, Shixiang  and
      Yin, Zhenfei  and
      Zheng, Xinzhe  and
      Li, Jinzhe  and
      Qi, Biqing  and
      Wu, Qi  and
      Li, Hui  and
      Ouyang, Wanli  and
      Torr, Philip  and
      Zhou, Bowen  and
      Dong, Nanqing",
    booktitle = "Proceedings of the 63rd Annual Meeting of the Association for Computational Linguistics (Volume 1: Long Papers)",
    month = jul,
    year = "2025",
    pages = "28201--28240"
}

@inproceedings{xiao-etal-2025-online,
    title = "Online Iterative Self-Alignment for Radiology Report Generation",
    author = "Xiao, Ting  and
      Shi, Lei  and
      Zhang, Yang  and
      Yang, HaoFeng  and
      Wang, Zhe  and
      Bai, Chenjia",
    booktitle = "Proceedings of the 63rd Annual Meeting of the Association for Computational Linguistics (Volume 1: Long Papers)",
    month = jul,
    year = "2025",
    pages = "27799--27814"
}

@inproceedings{wang-etal-2025-ormind,
    title = "{ORM}ind: A Cognitive-Inspired End-to-End Reasoning Framework for Operations Research",
    author = "Wang, Zhiyuan  and
      Chen, Bokui  and
      Huang, Yinya  and
      Cao, Qingxing  and
      He, Ming  and
      Fan, Jianping  and
      Liang, Xiaodan",
    booktitle = "Proceedings of the 63rd Annual Meeting of the Association for Computational Linguistics (Volume 6: Industry Track)",
    month = jul,
    year = "2025",
    pages = "104--131"
}

@misc{drtulu,
	title={DR Tulu: Reinforcement Learning with Evolving Rubrics for Deep Research}, 
	author={Rulin Shao and Akari Asai and Shannon Zejiang Shen and Hamish Ivison and Varsha Kishore and Jingming Zhuo and Xinran Zhao and Molly Park and Samuel G. Finlayson and David Sontag and Tyler Murray and Sewon Min and Pradeep Dasigi and Luca Soldaini and Faeze Brahman and Wen-tau Yih and Tongshuang Wu and Luke Zettlemoyer and Yoon Kim and Hannaneh Hajishirzi and Pang Wei Koh},
	year={2025},
	eprint={2511.19399},
	archivePrefix={arXiv},
	primaryClass={cs.CL}
}

@misc{hong2025multiagentdeepresearchtraining,
      title={Multi-Agent Deep Research: Training Multi-Agent Systems with M-GRPO}, 
      author={Haoyang Hong and Jiajun Yin and Yuan Wang and Jingnan Liu and Zhe Chen and Ailing Yu and Ji Li and Zhiling Ye and Hansong Xiao and Yefei Chen and Hualei Zhou and Yun Yue and Minghui Yang and Chunxiao Guo and Junwei Liu and Peng Wei and Jinjie Gu},
      year={2025},
      eprint={2511.13288},
      archivePrefix={arXiv},
      primaryClass={cs.AI}
}

@inproceedings{li-etal-2024-multimodal-arxiv,
    title = "Multimodal {A}r{X}iv: A Dataset for Improving Scientific Comprehension of Large Vision-Language Models",
    author = "Li, Lei  and
      Wang, Yuqi  and
      Xu, Runxin  and
      Wang, Peiyi  and
      Feng, Xiachong  and
      Kong, Lingpeng  and
      Liu, Qi",
    booktitle = "Proceedings of the 62nd Annual Meeting of the Association for Computational Linguistics (Volume 1: Long Papers)",
    month = aug,
    year = "2024",
    pages = "14369--14387"
}

@misc{zhang2025aixivnextgenerationopenaccess,
      title={aiXiv: A Next-Generation Open Access Ecosystem for Scientific Discovery Generated by AI Scientists}, 
      author={Pengsong Zhang and Xiang Hu and Guowei Huang and Yang Qi and Heng Zhang and Xiuxu Li and Jiaxing Song and Jiabin Luo and Yijiang Li and Shuo Yin and Chengxiao Dai and Eric Hanchen Jiang and Xiaoyan Zhou and Zhenfei Yin and Boqin Yuan and Jing Dong and Guinan Su and Guanren Qiao and Haiming Tang and Anghong Du and Lili Pan and Zhenzhong Lan and Xinyu Liu},
      year={2025},
      eprint={2508.15126},
      archivePrefix={arXiv},
      primaryClass={cs.AI}
}

@misc{wang2025scireasonerlayingscientificreasoning,
      title={SciReasoner: Laying the Scientific Reasoning Ground Across Disciplines}, 
      author={Yizhou Wang and Chen Tang and Han Deng and Jiabei Xiao and Jiaqi Liu and Jianyu Wu and Jun Yao and Pengze Li and Encheng Su and Lintao Wang and Guohang Zhuang and Yuchen Ren and Ben Fei and Ming Hu and Xin Chen and Dongzhan Zhou and Junjun He and Xiangyu Yue and Zhenfei Yin and Jiamin Wu and Qihao Zheng and Yuhao Zhou and Huihui Xu and Chenglong Ma and Yan Lu and Wenlong Zhang and Chunfeng Song and Philip Torr and Shixiang Tang and Xinzhu Ma and Wanli Ouyang and Lei Bai},
      year={2025},
      eprint={2509.21320},
      archivePrefix={arXiv},
      primaryClass={cs.CL}
}

@misc{bubeck2025earlyscienceaccelerationexperiments,
      title={Early science acceleration experiments with GPT-5}, 
      author={Sébastien Bubeck and Christian Coester and Ronen Eldan and Timothy Gowers and Yin Tat Lee and Alexandru Lupsasca and Mehtaab Sawhney and Robert Scherrer and Mark Sellke and Brian K. Spears and Derya Unutmaz and Kevin Weil and Steven Yin and Nikita Zhivotovskiy},
      year={2025},
      eprint={2511.16072},
      archivePrefix={arXiv},
      primaryClass={cs.CL},
      url={https://arxiv.org/abs/2511.16072}, 
}

@misc{liu2025atlashighdifficultymultidisciplinarybenchmark,
      title={ATLAS: A High-Difficulty, Multidisciplinary Benchmark for Frontier Scientific Reasoning}, 
      author={Hongwei Liu and Junnan Liu and Shudong Liu and Haodong Duan and Yuqiang Li and Mao Su and Xiaohong Liu and Guangtao Zhai and Xinyu Fang and Qianhong Ma and Taolin Zhang and Zihan Ma and Yufeng Zhao and Peiheng Zhou and Linchen Xiao and Wenlong Zhang and Shijie Zhou and Xingjian Ma and Siqi Sun and Jiaye Ge and Meng Li and Yuhong Liu and Jianxin Dong and Jiaying Li and Hui Wu and Hanwen Liang and Jintai Lin and Yanting Wang and Jie Dong and Tong Zhu and Tianfan Fu and Conghui He and Qi Zhang and Songyang Zhang and Lei Bai and Kai Chen},
      year={2025},
      eprint={2511.14366},
      archivePrefix={arXiv},
      primaryClass={cs.CL}
}

@misc{2025mirothinker,
    title={MiroFlow: A High-Performance Open-Source Research Agent Framework},
    author={AI Team MiroMind},
    howpublished={\url{https://github.com/MiroMindAI/MiroFlow}},
    year={2025}
}

@article{huang2025biomni,
  title={Biomni: A General-Purpose Biomedical AI Agent},
  author={Huang, Kexin and Zhang, Serena and Wang, Hanchen and Qu, Yuanhao and Lu, Yingzhou and Roohani, Yusuf and Li, Ryan and Qiu, Lin and Zhang, Junze and Di, Yin and others},
  journal={bioRxiv},
  pages={2025--05},
  year={2025}
}

@misc{grosskopf2025ursauniversalresearchscientific,
      title={URSA: The Universal Research and Scientific Agent}, 
      author={Michael Grosskopf and Russell Bent and Rahul Somasundaram and Isaac Michaud and Arthur Lui and Nathan Debardeleben and Earl Lawrence},
      year={2025},
      eprint={2506.22653},
      archivePrefix={arXiv},
      primaryClass={cs.AI}
}

@article{tinajero2025reac,
  title={Reac-Discovery: an artificial intelligence--driven platform for continuous-flow catalytic reactor discovery and optimization},
  author={Tinajero, Cristopher and Zanatta, Marcileia and S{\'a}nchez-Velandia, Juli{\'a}n E and Garc{\'\i}a-Verdugo, Eduardo and Sans, Victor},
  journal={Nature Communications},
  volume={16},
  number={1},
  pages={9062},
  year={2025}
}

@article{Preprints-dr,
	url = {https://doi.org/10.20944/preprints202511.2077.v1},
	year = 2025,
	month = {November},
	author = {Zhengliang Shi and Yiqun Chen and Haitao Li and Weiwei Sun and Shiyu Ni and Yougang Lyu and Run-Ze Fan and Bowen Jin and Yixuan Weng and Minjun Zhu and Qiujie Xie and Xinyu Guo and Qu Yang and Jiayi Wu and Jujia Zhao and Xiaqiang Tang and Xinbei Ma and Cunxiang Wang and Jiaxin Mao and Qingyao Ai and Jen-Tse Huang and Wenxuan Wang and Yue Zhang and Yiming Yang and Zhaopeng Tu and Zhaochun Ren},
	title = {Deep Research: A Systematic Survey},
	journal = {Preprints}
}

@article{han2025invdesflow,
  title={InvDesFlow-AL: active learning-based workflow for inverse design of functional materials},
  author={Han, Xiao-Qi and Guo, Peng-Jie and Gao, Ze-Feng and Sun, Hao and Lu, Zhong-Yi},
  journal={npj Computational Materials},
  year={2025}
}

@misc{jin2025finrptdatasetevaluationllmbased,
      title={FinRpt: Dataset, Evaluation System and LLM-based Multi-agent Framework for Equity Research Report Generation}, 
      author={Song Jin and Shuqi Li and Shukun Zhang and Rui Yan},
      year={2025},
      eprint={2511.07322},
      archivePrefix={arXiv},
      primaryClass={cs.CL}
}

@misc{zhang2025fargenuinelyusefuldeep,
      title={How Far Are We from Genuinely Useful Deep Research Agents?}, 
      author={Dingling Zhang and He Zhu and Jincheng Ren and Kangqi Song and Xinran Zhou and Boyu Feng and Shudong Liu and Jiabin Luo and Weihao Xie and Zhaohui Wang and Tianrui Qin and King Zhu and Yuqing Wang and Qianben Chen and Yuchen Eleanor Jiang and Wei Wang and Jiaheng Liu and Wangchunshu Zhou},
      year={2025},
      eprint={2512.01948},
      archivePrefix={arXiv},
      primaryClass={cs.CL}
}

@inproceedings{liomni,
  title={Omni-DNA: A Genomic Model Supporting Sequence Understanding, Long-context, and Textual Annotation},
  author={Li, Zehui and Subasri, Vallijah and Shen, Yifei and Li, Dongsheng and Gu, Wentao and Stan, Guy-Bart and Zhao, Yiren and Shan, Caihua},
  booktitle={The Thirty-ninth Annual Conference on Neural Information Processing Systems},
  year = {2025}
}

@article{mRNA-science,
author = {He Zhang  and Hailong Liu  and Yushan Xu  and Haoran Huang  and Yiming Liu  and Jia Wang  and Yan Qin  and Haiyan Wang  and Lili Ma  and Zhiyuan Xun  and Xuzhuang Hou  and Timothy K. Lu  and Jicong Cao },
title = {Deep generative models design mRNA sequences with enhanced translational capacity and stability},
journal = {Science},
volume = {390},
number = {6773},
pages = {eadr8470},
year = {2025}
}

@inproceedings{cao2025revivingdspadvancedtheorem,
      title={Reviving DSP for Advanced Theorem Proving in the Era of Reasoning Models}, 
      author={Chenrui Cao and Liangcheng Song and Zenan Li and Xinyi Le and Xian Zhang and Hui Xue and Fan Yang},
      booktitle={The Thirty-ninth Annual Conference on Neural Information Processing Systems},
      year = {2025}
}

@article{tom2024self,
  title={Self-driving laboratories for chemistry and materials science},
  author={Tom, Gary and Schmid, Stefan P and Baird, Sterling G and Cao, Yang and Darvish, Kourosh and Hao, Han and Lo, Stanley and Pablo-Garc{\'\i}a, Sergio and Rajaonson, Ella M and Skreta, Marta and others},
  journal={Chemical Reviews},
  volume={124},
  number={16},
  pages={9633--9732},
  year={2024}
}

@article{qiu2024towards,
  title={Towards building multilingual language model for medicine},
  author={Qiu, Pengcheng and Wu, Chaoyi and Zhang, Xiaoman and Lin, Weixiong and Wang, Haicheng and Zhang, Ya and Wang, Yanfeng and Xie, Weidi},
  journal={Nature Communications},
  volume={15},
  number={1},
  pages={8384},
  year={2024}
}

@article{zhang2024generalist,
  title={A generalist vision--language foundation model for diverse biomedical tasks},
  author={Zhang, Kai and Zhou, Rong and Adhikarla, Eashan and Yan, Zhiling and Liu, Yixin and Yu, Jun and Liu, Zhengliang and Chen, Xun and Davison, Brian D and Ren, Hui and others},
  journal={Nature Medicine},
  volume={30},
  number={11},
  pages={3129--3141},
  year={2024}
}

@article{singhal2025toward,
  title={Toward expert-level medical question answering with large language models},
  author={Singhal, Karan and Tu, Tao and Gottweis, Juraj and Sayres, Rory and Wulczyn, Ellery and Amin, Mohamed and Hou, Le and Clark, Kevin and Pfohl, Stephen R and Cole-Lewis, Heather and others},
  journal={Nature Medicine},
  volume={31},
  number={3},
  pages={943--950},
  year={2025}
}

@article{kang2024chatmof,
  title={ChatMOF: an artificial intelligence system for predicting and generating metal-organic frameworks using large language models},
  author={Kang, Yeonghun and Kim, Jihan},
  journal={Nature communications},
  volume={15},
  number={1},
  pages={4705},
  year={2024}
}

@article{zhang2025large,
  title={Large language models to accelerate organic chemistry synthesis},
  author={Zhang, Yu and Han, Yang and Chen, Shuai and Yu, Ruijie and Zhao, Xin and Liu, Xianbin and Zeng, Kaipeng and Yu, Mengdi and Tian, Jidong and Zhu, Feng and others},
  journal={Nature Machine Intelligence},
  pages={1--13},
  year={2025}
}

@article{ruan2024automatic,
  title={An automatic end-to-end chemical synthesis development platform powered by large language models},
  author={Ruan, Yixiang and Lu, Chenyin and Xu, Ning and He, Yuchen and Chen, Yixin and Zhang, Jian and Xuan, Jun and Pan, Jianzhang and Fang, Qun and Gao, Hanyu and others},
  journal={Nature communications},
  volume={15},
  number={1},
  pages={10160},
  year={2024}
}

@article{hao2024large,
  title={Large-scale foundation model on single-cell transcriptomics},
  author={Hao, Minsheng and Gong, Jing and Zeng, Xin and Liu, Chiming and Guo, Yucheng and Cheng, Xingyi and Wang, Taifeng and Ma, Jianzhu and Zhang, Xuegong and Song, Le},
  journal={Nature methods},
  volume={21},
  number={8},
  pages={1481--1491},
  year={2024}
}

@article{wang2023scientific,
  title={Scientific discovery in the age of artificial intelligence},
  author={Wang, Hanchen and Fu, Tianfan and Du, Yuanqi and Gao, Wenhao and Huang, Kexin and Liu, Ziming and Chandak, Payal and Liu, Shengchao and Van Katwyk, Peter and Deac, Andreea and others},
  journal={Nature},
  volume={620},
  number={7972},
  pages={47--60},
  year={2023}
}

@article{lu2024multimodal,
  title={A multimodal generative AI copilot for human pathology},
  author={Lu, Ming Y and Chen, Bowen and Williamson, Drew FK and Chen, Richard J and Zhao, Melissa and Chow, Aaron K and Ikemura, Kenji and Kim, Ahrong and Pouli, Dimitra and Patel, Ankush and others},
  journal={Nature},
  volume={634},
  number={8033},
  pages={466--473},
  year={2024}
}

@article{boiko2023autonomous,
  title={Autonomous chemical research with large language models},
  author={Boiko, Daniil A and MacKnight, Robert and Kline, Ben and Gomes, Gabe},
  journal={Nature},
  volume={624},
  number={7992},
  pages={570--578},
  year={2023}
}

@article{steyaert2023multimodal,
  title={Multimodal data fusion for cancer biomarker discovery with deep learning},
  author={Steyaert, Sandra and Pizurica, Marija and Nagaraj, Divya and Khandelwal, Priya and Hernandez-Boussard, Tina and Gentles, Andrew J and Gevaert, Olivier},
  journal={Nature machine intelligence},
  volume={5},
  number={4},
  pages={351--362},
  year={2023}
}

@article{he2025generalized,
  title={Generalized biological foundation model with unified nucleic acid and protein language},
  author={He, Yong and Fang, Pan and Shan, Yongtao and Pan, Yuanfei and Wei, Yanhong and Chen, Yichang and Chen, Yihao and Liu, Yi and Zeng, Zhenyu and Zhou, Zhan and others},
  journal={Nature Machine Intelligence},
  pages={1--12},
  year={2025}
}

@article{antunes2024crystal,
  title={Crystal structure generation with autoregressive large language modeling},
  author={Antunes, Luis M and Butler, Keith T and Grau-Crespo, Ricardo},
  journal={Nature Communications},
  volume={15},
  number={1},
  pages={10570},
  year={2024}
}

@article{de2025multimodal,
  title={A multimodal conversational agent for DNA, RNA and protein tasks},
  author={de Almeida, Bernardo P and Richard, Guillaume and Dalla-Torre, Hugo and Blum, Christopher and Hexemer, Lorenz and Pandey, Priyanka and Laurent, Stefan and Rajesh, Chandana and Lopez, Marie and Laterre, Alexandre and others},
  journal={Nature Machine Intelligence},
  pages={1--14},
  year={2025}
}

@article{cui2024scgpt,
  title={scGPT: toward building a foundation model for single-cell multi-omics using generative AI},
  author={Cui, Haotian and Wang, Chloe and Maan, Hassaan and Pang, Kuan and Luo, Fengning and Duan, Nan and Wang, Bo},
  journal={Nature methods},
  volume={21},
  number={8},
  pages={1470--1480},
  year={2024}
}

@article{m2024augmenting,
  title={Augmenting large language models with chemistry tools},
  author={M. Bran, Andres and Cox, Sam and Schilter, Oliver and Baldassari, Carlo and White, Andrew D and Schwaller, Philippe},
  journal={Nature Machine Intelligence},
  volume={6},
  number={5},
  pages={525--535},
  year={2024}
}

@article{wang2024multi,
  title={Multi-purpose RNA language modelling with motif-aware pretraining and type-guided fine-tuning},
  author={Wang, Ning and Bian, Jiang and Li, Yuchen and Li, Xuhong and Mumtaz, Shahid and Kong, Linghe and Xiong, Haoyi},
  journal={Nature Machine Intelligence},
  volume={6},
  number={5},
  pages={548--557},
  year={2024}
}

@article{lin2023evolutionary,
  title={Evolutionary-scale prediction of atomic-level protein structure with a language model},
  author={Lin, Zeming and Akin, Halil and Rao, Roshan and Hie, Brian and Zhu, Zhongkai and Lu, Wenting and Smetanin, Nikita and Verkuil, Robert and Kabeli, Ori and Shmueli, Yaniv and others},
  journal={Science},
  volume={379},
  number={6637},
  pages={1123--1130},
  year={2023}
}

@article{hayes2025simulating,
  title={Simulating 500 million years of evolution with a language model},
  author={Hayes, Thomas and Rao, Roshan and Akin, Halil and Sofroniew, Nicholas J and Oktay, Deniz and Lin, Zeming and Verkuil, Robert and Tran, Vincent Q and Deaton, Jonathan and Wiggert, Marius and others},
  journal={Science},
  volume={387},
  number={6736},
  pages={850--858},
  year={2025}
}

@article{nguyen2024sequence,
  title={Sequence modeling and design from molecular to genome scale with Evo},
  author={Nguyen, Eric and Poli, Michael and Durrant, Matthew G and Kang, Brian and Katrekar, Dhruva and Li, David B and Bartie, Liam J and Thomas, Armin W and King, Samuel H and Brixi, Garyk and others},
  journal={Science},
  volume={386},
  number={6723},
  pages={eado9336},
  year={2024}
}

@misc{wei2025aiscienceagenticscience,
      title={From AI for Science to Agentic Science: A Survey on Autonomous Scientific Discovery}, 
      author={Jiaqi Wei and Yuejin Yang and Xiang Zhang and Yuhan Chen and Xiang Zhuang and Zhangyang Gao and Dongzhan Zhou and Guangshuai Wang and Zhiqiang Gao and Juntai Cao and Zijie Qiu and Ming Hu and Chenglong Ma and Shixiang Tang and Junjun He and Chunfeng Song and Xuming He and Qiang Zhang and Chenyu You and Shuangjia Zheng and Ning Ding and Wanli Ouyang and Nanqing Dong and Yu Cheng and Siqi Sun and Lei Bai and Bowen Zhou},
      year={2025},
      eprint={2508.14111},
      archivePrefix={arXiv},
      primaryClass={cs.LG},
      url={https://arxiv.org/abs/2508.14111}, 
}

@misc{chen2025dingtalkdeepresearchunifiedmulti,
      title={Dingtalk DeepResearch: A Unified Multi Agent Framework for Adaptive Intelligence in Enterprise Environments}, 
      author={Mengyuan Chen and Chengjun Dai and Xinyang Dong and Chengzhe Feng and Kewei Fu and Jianshe Li and Zhihan Peng and Yongqi Tong and Junshao Zhang and Hong Zhu},
      year={2025},
      eprint={2510.24760},
      archivePrefix={arXiv},
      primaryClass={cs.CL}
}

@article{schaefer2025multimodal,
  title={Multimodal learning enables chat-based exploration of single-cell data},
  author={Schaefer, Moritz and Peneder, Peter and Malzl, Daniel and Lombardo, Salvo Danilo and Peycheva, Mihaela and Burton, Jake and Hakobyan, Anna and Sharma, Varun and Krausgruber, Thomas and Sin, Celine and others},
  journal={Nature Biotechnology},
  pages={1--11},
  year={2025}
}

@misc{hendrycks2025definitionagi,
      title={A Definition of AGI}, 
      author={Dan Hendrycks and Dawn Song and Christian Szegedy and Honglak Lee and Yarin Gal and Erik Brynjolfsson and Sharon Li and Andy Zou and Lionel Levine and Bo Han and Jie Fu and Ziwei Liu and Jinwoo Shin and Kimin Lee and Mantas Mazeika and Long Phan and George Ingebretsen and Adam Khoja and Cihang Xie and Olawale Salaudeen and Matthias Hein and Kevin Zhao and Alexander Pan and David Duvenaud and Bo Li and Steve Omohundro and Gabriel Alfour and Max Tegmark and Kevin McGrew and Gary Marcus and Jaan Tallinn and Eric Schmidt and Yoshua Bengio},
      year={2025},
      eprint={2510.18212},
      archivePrefix={arXiv},
      primaryClass={cs.AI},
      url={https://arxiv.org/abs/2510.18212}, 
}

@article{ai-science,
author = {Darío Gil  and Kathryn A. Moler},
title = {Accelerating science with AI},
journal = {Science},
volume = {390},
number = {6777},
pages = {965-965},
year = {2025}
}

@misc{simateam2025sima2generalistembodied,
      title={SIMA 2: A Generalist Embodied Agent for Virtual Worlds}, 
      author={Adrian Bolton and Alexander Lerchner and Alexandra Cordell and Alexandre Moufarek and Andrew Bolt and Andrew Lampinen and Anna Mitenkova and Arne Olav Hallingstad and Bojan Vujatovic and Bonnie Li and Cong Lu and Daan Wierstra and Daniel P. Sawyer and Daniel Slater and David Reichert and Davide Vercelli and Demis Hassabis and Drew A. Hudson and Duncan Williams and Ed Hirst and Fabio Pardo and Felix Hill and Frederic Besse and Hannah Openshaw and Harris Chan and Hubert Soyer and Jane X. Wang and Jeff Clune and John Agapiou and John Reid and Joseph Marino and Junkyung Kim and Karol Gregor and Kaustubh Sridhar and Kay McKinney and Laura Kampis and Lei M. Zhang and Loic Matthey and Luyu Wang and Maria Abi Raad and Maria Loks-Thompson and Martin Engelcke and Matija Kecman and Matthew Jackson and Maxime Gazeau and Ollie Purkiss and Oscar Knagg and Peter Stys and Piermaria Mendolicchio and Raia Hadsell and Rosemary Ke and Ryan Faulkner and Sarah Chakera and Satinder Singh Baveja and Shane Legg and Sheleem Kashem and Tayfun Terzi and Thomas Keck and Tim Harley and Tim Scholtes and Tyson Roberts and Volodymyr Mnih and Yulan Liu and Zhengdong Wang and Zoubin Ghahramani},
      year={2025},
      eprint={2512.04797},
      archivePrefix={arXiv},
      primaryClass={cs.AI},
      url={https://arxiv.org/abs/2512.04797}, 
}

@inproceedings{
	Si2025Can,
	title={Can {LLM}s Generate Novel Research Ideas? A Large-Scale Human Study with 100+ {NLP} Researchers},
	author={Chenglei Si and Diyi Yang and Tatsunori Hashimoto},
	booktitle={The Thirteenth International Conference on Learning Representations},
	year={2025}
}

@misc{Si2025Gap,
	title={The Ideation-Execution Gap: Execution Outcomes of LLM-Generated versus Human Research Ideas}, 
	author={Chenglei Si and Tatsunori Hashimoto and Diyi Yang},
	year={2025},
	eprint={2506.20803},
	archivePrefix={arXiv},
	primaryClass={cs.CL}
}

@misc{song2025evaluatinglargelanguagemodels,
	title={Evaluating Large Language Models in Scientific Discovery}, 
	author={Zhangde Song and Jieyu Lu and Yuanqi Du and Botao Yu and Thomas M. Pruyn and Yue Huang and Kehan Guo and Xiuzhe Luo and Yuanhao Qu and Yi Qu and Yinkai Wang and Haorui Wang and Jeff Guo and Jingru Gan and Parshin Shojaee and Di Luo and Andres M Bran and Gen Li and Qiyuan Zhao and Shao-Xiong Lennon Luo and Yuxuan Zhang and Xiang Zou and Wanru Zhao and Yifan F. Zhang and Wucheng Zhang and Shunan Zheng and Saiyang Zhang and Sartaaj Takrim Khan and Mahyar Rajabi-Kochi and Samantha Paradi-Maropakis and Tony Baltoiu and Fengyu Xie and Tianyang Chen and Kexin Huang and Weiliang Luo and Meijing Fang and Xin Yang and Lixue Cheng and Jiajun He and Soha Hassoun and Xiangliang Zhang and Wei Wang and Chandan K. Reddy and Chao Zhang and Zhiling Zheng and Mengdi Wang and Le Cong and Carla P. Gomes and Chang-Yu Hsieh and Aditya Nandy and Philippe Schwaller and Heather J. Kulik and Haojun Jia and Huan Sun and Seyed Mohamad Moosavi and Chenru Duan},
	year={2025},
	eprint={2512.15567},
	archivePrefix={arXiv},
	primaryClass={cs.AI},
	url={https://arxiv.org/abs/2512.15567}, 
}

@ARTICLE{Gao2024te,
	title    = "Quantifying the use and potential benefits of artificial
	intelligence in scientific research",
	author   = "Gao, Jian and Wang, Dashun",
	journal  = "Nature Human Behaviour",
	volume   =  8,
	number   =  12,
	pages    = "2281--2292",
	month    =  dec,
	year     =  2024
}

@ARTICLE{Merchant2025at,
	title    = "Semantic design of functional de novo genes from a genomic language model",
	author   = "Merchant, Aditi T and King, Samuel H and Nguyen, Eric and Hie, Brian L",
	journal  = "Nature",
	month    =  nov,
	year     =  2025
}

@article{CHUGUNOVA2026105381,
	title = {Who uses AI in research, and for what? Large-scale survey evidence from Germany},
	journal = {Research Policy},
	volume = {55},
	number = {2},
	pages = {105381},
	year = {2026},
	author = {Marina Chugunova and Dietmar Harhoff and Katharina Hölzle and Verena Kaschub and Sonal Malagimani and Ulrike Morgalla and Robert Rose}
}

@inproceedings{liu-etal-2025-vrope,
	title = "{VR}o{PE}: Rotary Position Embedding for Video Large Language Models",
	author = "Liu, Zikang  and
	Guo, Longteng  and
	Tang, Yepeng  and
	Yue, Tongtian  and
	Cai, Junxian  and
	Ma, Kai  and
	Liu, Qingbin  and
	Chen, Xi  and
	Liu, Jing",
	booktitle = "Proceedings of the 2025 Conference on Empirical Methods in Natural Language Processing",
	month = nov,
	year = "2025"
}

@article{RoFormer24,
	author = {Su, Jianlin and Ahmed, Murtadha and Lu, Yu and Pan, Shengfeng and Bo, Wen and Liu, Yunfeng},
	title = {RoFormer: Enhanced transformer with Rotary Position Embedding},
	year = {2024},
	issue_date = {Feb 2024},
	volume = {568},
	number = {C},
	journal = {Neurocomput.},
	month = feb,
	numpages = {12}
}

@inproceedings{dape24,
	author = {Zheng, Chuanyang and Gao, Yihang and Shi, Han and Huang, Minbin and Li, Jingyao and Xiong, Jing and Ren, Xiaozhe and Ng, Michael and Jiang, Xin and Li, Zhenguo and Li, Yu},
	title = {DAPE: data-adaptive positional encoding for length extrapolation},
	year = {2024},
	booktitle = {Proceedings of the 38th International Conference on Neural Information Processing Systems},
	articleno = {838},
	numpages = {42},
	series = {NIPS '24}
}

@misc{liu2025scientificreasoningmodelorganic,
	title={A Scientific Reasoning Model for Organic Synthesis Procedure Generation}, 
	author={Guoqing Liu and Junren Li and Zihan Zhao and Eray Inanc and Krzysztof Maziarz and Jose Garrido Torres and Victor Garcia Satorras and Shoko Ueda and Christopher M. Bishop and Marwin Segler},
	year={2025},
	eprint={2512.13668},
	archivePrefix={arXiv},
	primaryClass={cs.LG}
}

@inproceedings{kudo-2018-subword,
	title = "Subword Regularization: Improving Neural Network Translation Models with Multiple Subword Candidates",
	author = "Kudo, Taku",
	booktitle = "Proceedings of the 56th Annual Meeting of the Association for Computational Linguistics (Volume 1: Long Papers)",
	month = jul,
	year = "2018",
	pages = "66--75"
}

@inproceedings{NIPS2017_3f5ee243,
	author = {Vaswani, Ashish and Shazeer, Noam and Parmar, Niki and Uszkoreit, Jakob and Jones, Llion and Gomez, Aidan N and Kaiser, \L ukasz and Polosukhin, Illia},
	booktitle = {Advances in Neural Information Processing Systems},
	pages = {},
	title = {Attention is All you Need},
	volume = {30},
	year = {2017}
}

@inproceedings{devlin-etal-2019-bert,
	title = "{BERT}: Pre-training of Deep Bidirectional Transformers for Language Understanding",
	author = "Devlin, Jacob  and
	Chang, Ming-Wei  and
	Lee, Kenton  and
	Toutanova, Kristina",
	booktitle = "Proceedings of the 2019 Conference of the North {A}merican Chapter of the Association for Computational Linguistics: Human Language Technologies, Volume 1",
	month = jun,
	year = "2019",
	pages = "4171--4186"
}

@inproceedings{word2veciclr13,
	author       = {Tom{\'{a}}s Mikolov and
	Kai Chen and
	Greg Corrado and
	Jeffrey Dean},
	title        = {Efficient Estimation of Word Representations in Vector Space},
	booktitle    = {1st International Conference on Learning Representations, {ICLR} 2013},
	year         = {2013}
}

@article{radford2018improving,
	title={Improving language understanding by generative pre-training},
	author={Radford, Alec and Narasimhan, Karthik and Salimans, Tim and Sutskever, Ilya},
	year={2018},
	publisher={OpenAI},
	url={https://cdn.openai.com/research-covers/language-unsupervised/language_understanding_paper.pdf}
}

@inproceedings{sennrich-etal-2016-neural,
	title = "Neural Machine Translation of Rare Words with Subword Units",
	author = "Sennrich, Rico  and
	Haddow, Barry  and
	Birch, Alexandra",
	booktitle = "Proceedings of the 54th Annual Meeting of the Association for Computational Linguistics (Volume 1: Long Papers)",
	month = aug,
	year = "2016",
	pages = "1715--1725"
}

@inproceedings{gptrlhf,
	author = {Stiennon, Nisan and Ouyang, Long and Wu, Jeff and Ziegler, Daniel M. and Lowe, Ryan and Voss, Chelsea and Radford, Alec and Amodei, Dario and Christiano, Paul},
	title = {Learning to summarize from human feedback},
	year = {2020},
	booktitle = {Proceedings of the 34th International Conference on Neural Information Processing Systems},
	articleno = {253},
	numpages = {14},
	series = {NIPS '20}
}

@misc{kaplan2020scalinglawsneurallanguage,
	title={Scaling Laws for Neural Language Models}, 
	author={Jared Kaplan and Sam McCandlish and Tom Henighan and Tom B. Brown and Benjamin Chess and Rewon Child and Scott Gray and Alec Radford and Jeffrey Wu and Dario Amodei},
	year={2020},
	eprint={2001.08361},
	archivePrefix={arXiv},
	primaryClass={cs.LG}
}

@article{
	wei2022emergent,
	title={Emergent Abilities of Large Language Models},
	author={Jason Wei and Yi Tay and Rishi Bommasani and Colin Raffel and Barret Zoph and Sebastian Borgeaud and Dani Yogatama and Maarten Bosma and Denny Zhou and Donald Metzler and Ed H. Chi and Tatsunori Hashimoto and Oriol Vinyals and Percy Liang and Jeff Dean and William Fedus},
	journal={Transactions on Machine Learning Research},
	year={2022}
}

@inproceedings{
	hu2022lora,
	title={Lo{RA}: Low-Rank Adaptation of Large Language Models},
	author={Edward J Hu and Yelong Shen and Phillip Wallis and Zeyuan Allen-Zhu and Yuanzhi Li and Shean Wang and Lu Wang and Weizhu Chen},
	booktitle={International Conference on Learning Representations},
	year={2022}
}

@misc{yao2023reactsynergizingreasoningacting,
	title={ReAct: Synergizing Reasoning and Acting in Language Models}, 
	author={Shunyu Yao and Jeffrey Zhao and Dian Yu and Nan Du and Izhak Shafran and Karthik Narasimhan and Yuan Cao},
	year={2023},
	eprint={2210.03629},
	archivePrefix={arXiv},
	primaryClass={cs.CL}
}

@inproceedings{toolformer23,
	author = {Schick, Timo and Dwivedi-Yu, Jane and Dess\'{\i}, Roberto and Raileanu, Roberta and Lomeli, Maria and Hambro, Eric and Zettlemoyer, Luke and Cancedda, Nicola and Scialom, Thomas},
	title = {Toolformer: language models can teach themselves to use tools},
	year = {2023},
	booktitle = {Proceedings of the 37th International Conference on Neural Information Processing Systems},
	articleno = {2997},
	numpages = {13},
	series = {NIPS '23}
}

@inproceedings{ddpm20,
	author = {Ho, Jonathan and Jain, Ajay and Abbeel, Pieter},
	title = {Denoising diffusion probabilistic models},
	year = {2020},
	booktitle = {Proceedings of the 34th International Conference on Neural Information Processing Systems},
	articleno = {574},
	numpages = {12},
	series = {NIPS '20}
}

@InProceedings{Rombach_2022_CVPR,
	author    = {Rombach, Robin and Blattmann, Andreas and Lorenz, Dominik and Esser, Patrick and Ommer, Bj\"orn},
	title     = {High-Resolution Image Synthesis With Latent Diffusion Models},
	booktitle = {Proceedings of the IEEE/CVF Conference on Computer Vision and Pattern Recognition (CVPR)},
	month     = {June},
	year      = {2022},
	pages     = {10684-10695}
}

@inproceedings{sd324,
	author = {Esser, Patrick and Kulal, Sumith and Blattmann, Andreas and Entezari, Rahim and M\"{u}ller, Jonas and Saini, Harry and Levi, Yam and Lorenz, Dominik and Sauer, Axel and Boesel, Frederic and Podell, Dustin and Dockhorn, Tim and English, Zion and Rombach, Robin},
	title = {Scaling rectified flow transformers for high-resolution image synthesis},
	year = {2024},
	booktitle = {Proceedings of the 41st International Conference on Machine Learning},
	articleno = {503},
	numpages = {28},
	series = {ICML'24}
}

@inproceedings{peebles2023scalable,
	title={Scalable diffusion models with transformers},
	author={Peebles, William and Xie, Saining},
	booktitle={Proceedings of the IEEE/CVF international conference on computer vision},
	pages={4195--4205},
	year={2023}
}

@inproceedings{ronneberger2015u,
	title={U-net: Convolutional networks for biomedical image segmentation},
	author={Ronneberger, Olaf and Fischer, Philipp and Brox, Thomas},
	booktitle={International Conference on Medical image computing and computer-assisted intervention},
	pages={234--241},
	year={2015}
}

@INPROCEEDINGS{zhang2023adding,
	author={Zhang, Lvmin and Rao, Anyi and Agrawala, Maneesh},
	booktitle={2023 IEEE/CVF International Conference on Computer Vision (ICCV)}, 
	title={Adding Conditional Control to Text-to-Image Diffusion Models}, 
	year={2023},
	pages={3813-3824}
}

@article{wang2024instantid,
	title={InstantID: Zero-shot Identity-Preserving Generation in Seconds},
	author={Wang, Qixun and Bai, Xu and Wang, Haofan and Qin, Zekui and Chen, Anthony},
	journal={arXiv preprint arXiv:2401.07519},
	year={2024}
}

@inproceedings{ma2024sit,
	title={Sit: Exploring flow and diffusion-based generative models with scalable interpolant transformers},
	author={Ma, Nanye and Goldstein, Mark and Albergo, Michael S and Boffi, Nicholas M and Vanden-Eijnden, Eric and Xie, Saining},
	booktitle={European Conference on Computer Vision},
	pages={23--40},
	year={2024}
}

@article{guo2023animatediff,
	title={AnimateDiff: Animate Your Personalized Text-to-Image Diffusion Models without Specific Tuning},
	author={Guo, Yuwei and Yang, Ceyuan and Rao, Anyi and Liang, Zhengyang and Wang, Yaohui and Qiao, Yu and Agrawala, Maneesh and Lin, Dahua and Dai, Bo},
	journal={International Conference on Learning Representations},
	year={2024}
}

@misc{du2025acceleratingscientificdiscoveryautonomous,
	title={Accelerating Scientific Discovery with Autonomous Goal-evolving Agents}, 
	author={Yuanqi Du and Botao Yu and Tianyu Liu and Tony Shen and Junwu Chen and Jan G. Rittig and Kunyang Sun and Yikun Zhang and Zhangde Song and Bo Zhou and Cassandra Masschelein and Yingze Wang and Haorui Wang and Haojun Jia and Chao Zhang and Hongyu Zhao and Martin Ester and Teresa Head-Gordon and Carla P. Gomes and Huan Sun and Chenru Duan and Philippe Schwaller and Wengong Jin},
	year={2025},
	eprint={2512.21782},
	archivePrefix={arXiv},
	primaryClass={cs.AI},
	url={https://arxiv.org/abs/2512.21782}, 
}

@ARTICLE{Hafner2025-kx,
	title    = "Mastering diverse control tasks through world models",
	author   = "Hafner, Danijar and Pasukonis, Jurgis and Ba, Jimmy and
	Lillicrap, Timothy",
	journal  = "Nature",
	volume   =  640,
	number   =  8059,
	pages    = "647--653",
	month    =  apr,
	year     =  2025
}

@misc{goel2025trainingaicoscientistsusing,
	title={Training AI Co-Scientists Using Rubric Rewards}, 
	author={Shashwat Goel and Rishi Hazra and Dulhan Jayalath and Timon Willi and Parag Jain and William F. Shen and Ilias Leontiadis and Francesco Barbieri and Yoram Bachrach and Jonas Geiping and Chenxi Whitehouse},
	year={2025},
	eprint={2512.23707},
	archivePrefix={arXiv},
	primaryClass={cs.LG},
	url={https://arxiv.org/abs/2512.23707}, 
}

@article{t520,
	author = {Raffel, Colin and Shazeer, Noam and Roberts, Adam and Lee, Katherine and Narang, Sharan and Matena, Michael and Zhou, Yanqi and Li, Wei and Liu, Peter J.},
	title = {Exploring the limits of transfer learning with a unified text-to-text transformer},
	year = {2020},
	volume = {21},
	number = {1},
	journal = {J. Mach. Learn. Res.},
	month = jan,
	articleno = {140},
	numpages = {67}
}

@article{radford2019language,
	title={Language models are unsupervised multitask learners},
	author={Radford, Alec and Wu, Jeffrey and Child, Rewon and Luan, David and Amodei, Dario and Sutskever, Ilya and others},
	journal={OpenAI blog},
	volume={1},
	number={8},
	pages={9},
	year={2019},
	url={https://cdn.openai.com/better-language-models/language_models_are_unsupervised_multitask_learners.pdf}
}

@article{brown2020language,
	title={Language models are few-shot learners},
	author={Brown, Tom and Mann, Benjamin and Ryder, Nick and Subbiah, Melanie and Kaplan, Jared D and Dhariwal, Prafulla and Neelakantan, Arvind and Shyam, Pranav and Sastry, Girish and Askell, Amanda and others},
	journal={Advances in neural information processing systems},
	volume={33},
	pages={1877--1901},
	year={2020}
}

@misc{power2022grokking,
	title={Grokking: Generalization Beyond Overfitting on Small Algorithmic Datasets}, 
	author={Alethea Power and Yuri Burda and Harri Edwards and Igor Babuschkin and Vedant Misra},
	year={2022},
	eprint={2201.02177},
	archivePrefix={arXiv},
	primaryClass={cs.LG},
	url={https://arxiv.org/abs/2201.02177}, 
}

@misc{ge2025seedxmultimodalmodelsunified,
	title={SEED-X: Multimodal Models with Unified Multi-granularity Comprehension and Generation}, 
	author={Yuying Ge and Sijie Zhao and Jinguo Zhu and Yixiao Ge and Kun Yi and Lin Song and Chen Li and Xiaohan Ding and Ying Shan},
	year={2025},
	eprint={2404.14396},
	archivePrefix={arXiv},
	primaryClass={cs.CV},
}

@misc{wang2024emu3,
	title={Emu3: Next-Token Prediction is All You Need}, 
	author={Xinlong Wang and Xiaosong Zhang and Zhengxiong Luo and Quan Sun and Yufeng Cui and Jinsheng Wang and Fan Zhang and Yueze Wang and Zhen Li and Qiying Yu and Yingli Zhao and Yulong Ao and Xuebin Min and Tao Li and Boya Wu and Bo Zhao and Bowen Zhang and Liangdong Wang and Guang Liu and Zheqi He and Xi Yang and Jingjing Liu and Yonghua Lin and Tiejun Huang and Zhongyuan Wang},
	year={2024},
	eprint={2409.18869},
	archivePrefix={arXiv},
	primaryClass={cs.CV},
	url={https://arxiv.org/abs/2409.18869}, 
}

@misc{chen2025janusprounifiedmultimodalunderstanding,
	title={Janus-Pro: Unified Multimodal Understanding and Generation with Data and Model Scaling}, 
	author={Xiaokang Chen and Zhiyu Wu and Xingchao Liu and Zizheng Pan and Wen Liu and Zhenda Xie and Xingkai Yu and Chong Ruan},
	year={2025},
	eprint={2501.17811},
	archivePrefix={arXiv},
	primaryClass={cs.AI}
}

@article{Chen2024MultiModalGA,
	title={Multi-Modal Generative AI: Multi-modal LLM, Diffusion and Beyond},
	author={Hong Chen and Xin Wang and Yuwei Zhou and Bin Huang and Yipeng Zhang and Wei Feng and Houlun Chen and Zeyang Zhang and Siao Tang and Wenwu Zhu},
	journal={ArXiv},
	year={2024},
	volume={abs/2409.14993}
}

@inproceedings{wu2025janus,
	title={Janus: Decoupling visual encoding for unified multimodal understanding and generation},
	author={Wu, Chengyue and Chen, Xiaokang and Wu, Zhiyu and Ma, Yiyang and Liu, Xingchao and Pan, Zizheng and Liu, Wen and Xie, Zhenda and Yu, Xingkai and Ruan, Chong and others},
	booktitle={Proceedings of the Computer Vision and Pattern Recognition Conference},
	pages={12966--12977},
	year={2025}
}

@inproceedings{
	xie2025showo,
	title={Show-o: One Single Transformer to Unify Multimodal Understanding and Generation},
	author={Jinheng Xie and Weijia Mao and Zechen Bai and David Junhao Zhang and Weihao Wang and Kevin Qinghong Lin and Yuchao Gu and Zhijie Chen and Zhenheng Yang and Mike Zheng Shou},
	booktitle={The Thirteenth International Conference on Learning Representations},
	year={2025}
}

@inproceedings{ma2025janusflow,
	title={Janusflow: Harmonizing autoregression and rectified flow for unified multimodal understanding and generation},
	author={Ma, Yiyang and Liu, Xingchao and Chen, Xiaokang and Liu, Wen and Wu, Chengyue and Wu, Zhiyu and Pan, Zizheng and Xie, Zhenda and Zhang, Haowei and Yu, Xingkai and others},
	booktitle={Proceedings of the Computer Vision and Pattern Recognition Conference},
	pages={7739--7751},
	year={2025}
}

@inproceedings{cot22,
	author = {Wei, Jason and Wang, Xuezhi and Schuurmans, Dale and Bosma, Maarten and Ichter, Brian and Xia, Fei and Chi, Ed H. and Le, Quoc V. and Zhou, Denny},
	title = {Chain-of-thought prompting elicits reasoning in large language models},
	year = {2022},
	booktitle = {Proceedings of the 36th International Conference on Neural Information Processing Systems},
	articleno = {1800},
	numpages = {14},
	series = {NIPS '22}
}

@inproceedings{reflexion23,
	author = {Shinn, Noah and Cassano, Federico and Gopinath, Ashwin and Narasimhan, Karthik and Yao, Shunyu},
	title = {Reflexion: language agents with verbal reinforcement learning},
	year = {2023},
	booktitle = {Proceedings of the 37th International Conference on Neural Information Processing Systems},
	articleno = {377},
	numpages = {19},
	series = {NIPS '23}
}

@misc{nakano2022webgpt,
	title={WebGPT: Browser-assisted question-answering with human feedback}, 
	author={Reiichiro Nakano and Jacob Hilton and Suchir Balaji and Jeff Wu and Long Ouyang and Christina Kim and Christopher Hesse and Shantanu Jain and Vineet Kosaraju and William Saunders and Xu Jiang and Karl Cobbe and Tyna Eloundou and Gretchen Krueger and Kevin Button and Matthew Knight and Benjamin Chess and John Schulman},
	year={2022},
	eprint={2112.09332},
	archivePrefix={arXiv},
	primaryClass={cs.CL}
}

@inproceedings{tot23,
	author = {Yao, Shunyu and Yu, Dian and Zhao, Jeffrey and Shafran, Izhak and Griffiths, Thomas L. and Cao, Yuan and Narasimhan, Karthik},
	title = {Tree of thoughts: deliberate problem solving with large language models},
	year = {2023},
	booktitle = {Proceedings of the 37th International Conference on Neural Information Processing Systems},
	articleno = {517},
	numpages = {14},
	series = {NIPS '23}
}

@misc{yao2026oresearcheropenendeddeep,
	title={O-Researcher: An Open Ended Deep Research Model via Multi-Agent Distillation and Agentic RL}, 
	author={Yi Yao and He Zhu and Piaohong Wang and Jincheng Ren and Xinlong Yang and Qianben Chen and Xiaowan Li and Dingfeng Shi and Jiaxian Li and Qiexiang Wang and Sinuo Wang and Xinpeng Liu and Jiaqi Wu and Minghao Liu and Wangchunshu Zhou},
	year={2026},
	eprint={2601.03743},
	archivePrefix={arXiv},
	primaryClass={cs.CL}
}

@misc{trehan2026llmsarentscientistsyet,
	title={Why LLMs Aren't Scientists Yet: Lessons from Four Autonomous Research Attempts}, 
	author={Dhruv Trehan and Paras Chopra},
	year={2026},
	eprint={2601.03315},
	archivePrefix={arXiv},
	primaryClass={cs.LG}
}

@misc{wang2026deploymaster,
	title={Deploy-Master: Automating the Deployment of 50,000+ Agent-Ready Scientific Tools in One Day}, 
	author={Yi Wang and Zhenting Huang and Zhaohan Ding and Ruoxue Liao and Yuan Huang and Xinzijian Liu and Jiajun Xie and Siheng Chen and Linfeng Zhang},
	year={2026},
	eprint={2601.03513},
	archivePrefix={arXiv},
	primaryClass={cs.SE}
}

@misc{mukund2026marvelmultiagentbasedresearch,
	title={MARVEL: A Multi Agent-based Research Validator and Enabler using Large Language Models}, 
	author={Nikhil Mukund and Yifang Luo and Fan Zhang and Lisa Barsotti and Erik Katsavounidis},
	year={2026},
	eprint={2601.03436},
	archivePrefix={arXiv},
	primaryClass={astro-ph.IM}
}

@misc{touvron2023llama2openfoundation,
	title={Llama 2: Open Foundation and Fine-Tuned Chat Models}, 
	author={Hugo Touvron and Louis Martin and Kevin Stone and Peter Albert and Amjad Almahairi and Yasmine Babaei and Nikolay Bashlykov and Soumya Batra and Prajjwal Bhargava and Shruti Bhosale and Dan Bikel and Lukas Blecher and Cristian Canton Ferrer and Moya Chen and Guillem Cucurull and David Esiobu and Jude Fernandes and Jeremy Fu and Wenyin Fu and Brian Fuller and Cynthia Gao and Vedanuj Goswami and Naman Goyal and Anthony Hartshorn and Saghar Hosseini and Rui Hou and Hakan Inan and Marcin Kardas and Viktor Kerkez and Madian Khabsa and Isabel Kloumann and Artem Korenev and Punit Singh Koura and Marie-Anne Lachaux and Thibaut Lavril and Jenya Lee and Diana Liskovich and Yinghai Lu and Yuning Mao and Xavier Martinet and Todor Mihaylov and Pushkar Mishra and Igor Molybog and Yixin Nie and Andrew Poulton and Jeremy Reizenstein and Rashi Rungta and Kalyan Saladi and Alan Schelten and Ruan Silva and Eric Michael Smith and Ranjan Subramanian and Xiaoqing Ellen Tan and Binh Tang and Ross Taylor and Adina Williams and Jian Xiang Kuan and Puxin Xu and Zheng Yan and Iliyan Zarov and Yuchen Zhang and Angela Fan and Melanie Kambadur and Sharan Narang and Aurelien Rodriguez and Robert Stojnic and Sergey Edunov and Thomas Scialom},
	year={2023},
	eprint={2307.09288},
	archivePrefix={arXiv},
	primaryClass={cs.CL}
}

@Article{phan2025humanitysexam,
	author={Phan, Long
	and Gatti, Alice
	and Li, Nathaniel
	and Khoja, Adam
	and Kim, Ryan
	and Ren, Richard
	and Hausenloy, Jason
	and Zhang, Oliver
	and Mazeika, Mantas
	and Hendrycks, Dan
	and Han, Ziwen
	and Hu, Josephina
	and Zhang, Hugh
	and Zhang, Chen Bo Calvin
	and Shaaban, Mohamed
	and Ling, John
	and Shi, Sean
	and Choi, Michael
	and Agrawal, Anish
	and Chopra, Arnav
	and Nattanmai, Aakaash
	and McKellips, Gordon
	and Cheraku, Anish
	and Suhail, Asim
	and Luo, Ethan
	and Deng, Marvin
	and Luo, Jason
	and Zhang, Ashley
	and Jindel, Kavin
	and Paek, Jay
	and Halevy, Kasper
	and Baranov, Allen
	and Liu, Michael
	and Avadhanam, Advaith
	and Zhang, David
	and Cheng, Vincent
	and Ma, Brad
	and Fu, Evan
	and Do, Liam
	and Lass, Joshua
	and Yang, Hubert
	and Sunkari, Surya
	and Bharath, Vishruth
	and Ai, Violet
	and Leung, James
	and Agrawal, Rishit
	and Zhou, Alan
	and Chen, Kevin
	and Kalpathi, Tejas
	and Xu, Ziqi
	and Wang, Gavin
	and Xiao, Tyler
	and Maung, Erik
	and Lee, Sam
	and Yang, Ryan
	and Yue, Roy
	and Zhao, Ben
	and Yoon, Julia
	and Sun, Xiangwan
	and Singh, Aryan
	and Peng, Clark
	and Osbey, Tyler
	and Wang, Taozhi
	and Echeazu, Daryl
	and Wu, Timothy
	and Patel, Spandan
	and Kulkarni, Vidhi
	and Sundarapandiyan, Vijaykaarti
	and Le, Andrew
	and Nasim, Zafir
	and Yalam, Srikar
	and Kasamsetty, Ritesh
	and Samal, Soham
	and Sun, David
	and Shah, Nihar
	and Saha, Abhijeet
	and Zhang, Alex
	and Nguyen, Leon
	and Nagumalli, Laasya
	and Wang, Kaixin
	and Wu, Aidan
	and Telluri, Anwith
	and Yue, Summer
	and Wang, Alexandr
	and Dodonov, Dmitry
	and Nguyen, Tung
	and Lee, Jaeho
	and Anderson, Daron
	and Doroshenko, Mikhail
	and Stokes, Alun Cennyth
	and Mahmood, Mobeen
	and Pokutnyi, Oleksandr
	and Iskra, Oleg
	and Wang, Jessica P.
	and Levin, John-Clark
	and Kazakov, Mstyslav
	and Feng, Fiona
	and Feng, Steven Y.
	and Zhao, Haoran
	and Yu, Michael
	and Gangal, Varun
	and Zou, Chelsea
	and Wang, Zihan
	and Popov, Serguei
	and Gerbicz, Robert
	and Galgon, Geoff
	and Schmitt, Johannes
	and Yeadon, Will
	and Lee, Yongki
	and Sauers, Scott
	and Sanchez, Alvaro
	and Giska, Fabian
	and Roth, Marc
	and Riis, S{\o}ren
	and Utpala, Saiteja
	and Burns, Noah
	and Goshu, Gashaw M.
	and Naiya, Mohinder Maheshbhai
	and Agu, Chidozie
	and Giboney, Zachary
	and Cheatom, Antrell
	and Fournier-Facio, Francesco
	and Crowson, Sarah-Jane
	and Finke, Lennart
	and Cheng, Zerui
	and Zampese, Jennifer
	and Hoerr, Ryan G.
	and Nandor, Mark
	and Park, Hyunwoo
	and Gehrunger, Tim
	and Cai, Jiaqi
	and McCarty, Ben
	and Garretson, Alexis C.
	and Taylor, Edwin
	and Sileo, Damien
	and Ren, Qiuyu
	and Qazi, Usman
	and Li, Lianghui
	and Nam, Jungbae
	and Wydallis, John B.
	and Arkhipov, Pavel
	and Shi, Jack Wei Lun
	and Bacho, Aras
	and Willcocks, Chris G.
	and Cao, Hangrui
	and Motwani, Sumeet
	and de Oliveira Santos, Emily
	and Veith, Johannes
	and Vendrow, Edward
	and Cojoc, Doru
	and Zenitani, Kengo
	and Robinson, Joshua
	and Tang, Longke
	and Li, Yuqi
	and Vendrow, Joshua
	and Fraga, Natanael Wildner
	and Kuchkin, Vladyslav
	and Maksimov, Andrey Pupasov
	and Marion, Pierre
	and Efremov, Denis
	and Lynch, Jayson
	and Liang, Kaiqu
	and Mikov, Aleksandar
	and Gritsevskiy, Andrew
	and Guillod, Julien
	and Demir, G{\"o}zdenur
	and Martinez, Dakotah
	and Pageler, Ben
	and Zhou, Kevin
	and Soori, Saeed
	and Press, Ori
	and Tang, Henry
	and Rissone, Paolo
	and Green, Sean R.
	and Br{\"u}ssel, Lina
	and Twayana, Moon
	and Dieuleveut, Aymeric
	and Imperial, Joseph Marvin
	and Prabhu, Ameya
	and Yang, Jinzhou
	and Crispino, Nick
	and Rao, Arun
	and Zvonkine, Dimitri
	and Loiseau, Gabriel
	and Kalinin, Mikhail
	and Lukas, Marco
	and Manolescu, Ciprian
	and Stambaugh, Nate
	and Mishra, Subrata
	and Hogg, Tad
	and Bosio, Carlo
	and Coppola, Brian P.
	and Salazar, Julian
	and Jin, Jaehyeok
	and Sayous, Rafael
	and Ivanov, Stefan
	and Schwaller, Philippe
	and Senthilkumar, Shaipranesh
	and Bran, Andres M.
	and Algaba, Andres
	and Van den Houte, Kelsey
	and Van Der Sypt, Lynn
	and Verbeken, Brecht
	and Noever, David
	and Kopylov, Alexei
	and Myklebust, Benjamin
	and Li, Bikun
	and Schut, Lisa
	and Zheltonozhskii, Evgenii
	and Yuan, Qiaochu
	and Lim, Derek
	and Stanley, Richard
	and Yang, Tong
	and Maar, John
	and Wykowski, Julian
	and Oller, Mart
	and Sahu, Anmol
	and Ardito, Cesare Giulio
	and Hu, Yuzheng
	and Kamdoum, Ariel Ghislain Kemogne
	and Jin, Alvin
	and Vilchis, Tobias Garcia
	and Zu, Yuexuan
	and Lackner, Martin
	and Koppel, James
	and Sun, Gongbo
	and Antonenko, Daniil S.
	and Chern, Steffi
	and Zhao, Bingchen
	and Arsene, Pierrot
	and Cavanagh, Joseph M.
	and Li, Daofeng
	and Shen, Jiawei
	and Crisostomi, Donato
	and Zhang, Wenjin
	and Dehghan, Ali
	and Ivanov, Sergey
	and Perrella, David
	and Kaparov, Nurdin
	and Zang, Allen
	and Sucholutsky, Ilia
	and Kharlamova, Arina
	and Orel, Daniil
	and Poritski, Vladislav
	and Ben-David, Shalev
	and Berger, Zachary
	and Whitfill, Parker
	and Foster, Michael
	and Munro, Daniel
	and Ho, Linh
	and Sivarajan, Shankar
	and Hava, Dan Bar
	and Kuchkin, Aleksey
	and Holmes, David
	and Rodriguez-Romero, Alexandra
	and Sommerhage, Frank
	and Zhang, Anji
	and Moat, Richard
	and Schneider, Keith
	and Kazibwe, Zakayo
	and Clarke, Don
	and Kim, Dae Hyun
	and Dias, Felipe Meneguitti
	and Fish, Sara
	and Elser, Veit
	and Kreiman, Tobias
	and Vilchis, Victor Efren Guadarrama
	and Klose, Immo
	and Anantheswaran, Ujjwala
	and Zweiger, Adam
	and Rawal, Kaivalya
	and Li, Jeffery
	and Nguyen, Jeremy
	and Daans, Nicolas
	and Heidinger, Haline
	and Radionov, Maksim
	and Rozho{\v{n}}, V{\'a}clav
	and Ginis, Vincent
	and Stump, Christian
	and Cohen, Niv
	and Po{\'{s}}wiata, Rafa{\l}
	and Tkadlec, Josef
	and Goldfarb, Alan
	and Wang, Chenguang
	and Padlewski, Piotr
	and Barzowski, Stanislaw
	and Montgomery, Kyle
	and Stendall, Ryan
	and Tucker-Foltz, Jamie
	and Stade, Jack
	and Rogers, T. Ryan
	and Goertzen, Tom
	and Grabb, Declan
	and Shukla, Abhishek
	and Givr{\'e}, Alan
	and Ambay, John Arnold
	and Sen, Archan
	and for AI Safety, Center
	and AI, Scale
	and Consortium, HLE Contributors},
	title={A benchmark of expert-level academic questions to assess AI capabilities},
	journal={Nature},
	year={2026},
	month={Jan},
	day={01},
	volume={649},
	number={8099},
	pages={1139-1146}
}

@Article{Hao2026,
	author={Hao, Qianyue
	and Xu, Fengli
	and Li, Yong
	and Evans, James},
	title={Artificial intelligence tools expand scientists' impact but contract science's focus},
	journal={Nature},
	year={2026},
	month={Jan},
	day={14}
}

@Article{Liu2025,
	author={Liu, Linjing
	and Li, Wei
	and Wang, Fang
	and Li, Yiming
	and Huang, Long-Kai
	and Wong, Ka-Chun
	and Yang, Fan
	and Yao, Jianhua},
	title={A pre-trained large generative model for translating single-cell transcriptomes to proteomes},
	journal={Nature Biomedical Engineering},
	year={2025},
	month={Nov},
	day={05}
}

@Article{Ong2026,
	author={Ong, Jasmine Chiat Ling
	and Ning, Yilin
	and Yang, Rui
	and Bitterman, Danielle S.
	and Liu, Xiaoxuan
	and Tham, Yih Chung
	and Collins, Gary S.
	and Jim{\'e}nez de Tav{\'a}rez, Michelle Mar{\'i}a
	and Mateen, Bilal A.
	and Amissah-Arthur, Kwesi Nyan
	and Sheng, Bin
	and Tan, Iain Bee Huat
	and Hong, Chuan
	and Cheng, Lionel Tim-Ee
	and Goldstein, Benjamin Alan
	and Le, Phuoc V.
	and Liu, Yun
	and Tan, Hiang Khoon
	and Ong, Marcus Eng Hock
	and Wagner, Siegfried K.
	and Denniston, Alastair K.
	and Keane, Pearse A.
	and Car, Josip
	and Chapman, Wendy W.
	and Moons, Karel G. M.
	and Wong, Tien Yin
	and Topol, Eric J.
	and Liu, Nan},
	title={Large language models in global health},
	journal={Nature Health},
	year={2026},
	month={Jan},
	day={01},
	volume={1},
	number={1},
	pages={35-47}
}

@inproceedings{mialon2023gaia,
	title={Gaia: a benchmark for general ai assistants},
	author={Mialon, Gr{\'e}goire and Fourrier, Cl{\'e}mentine and Wolf, Thomas and LeCun, Yann and Scialom, Thomas},
	booktitle={The Twelfth International Conference on Learning Representations},
	year={2023}
}

@misc{wei2025browsecompsimplechallengingbenchmark,
	title={BrowseComp: A Simple Yet Challenging Benchmark for Browsing Agents}, 
	author={Jason Wei and Zhiqing Sun and Spencer Papay and Scott McKinney and Jeffrey Han and Isa Fulford and Hyung Won Chung and Alex Tachard Passos and William Fedus and Amelia Glaese},
	year={2025},
	eprint={2504.12516},
	archivePrefix={arXiv},
	primaryClass={cs.CL}
}

@misc{xu2023supercluecomprehensivechineselarge,
	title={SuperCLUE: A Comprehensive Chinese Large Language Model Benchmark}, 
	author={Liang Xu and Anqi Li and Lei Zhu and Hang Xue and Changtai Zhu and Kangkang Zhao and Haonan He and Xuanwei Zhang and Qiyue Kang and Zhenzhong Lan},
	year={2023},
	eprint={2307.15020},
	archivePrefix={arXiv},
	primaryClass={cs.CL}
}

@misc{chen2025xbenchtrackingagentsproductivity,
	title={xbench: Tracking Agents Productivity Scaling with Profession-Aligned Real-World Evaluations}, 
	author={Kaiyuan Chen and Yixin Ren and Yang Liu and Xiaobo Hu and Haotong Tian and Tianbao Xie and Fangfu Liu and Haoye Zhang and Hongzhang Liu and Yuan Gong and Chen Sun and Han Hou and Hui Yang and James Pan and Jianan Lou and Jiayi Mao and Jizheng Liu and Jinpeng Li and Kangyi Liu and Kenkun Liu and Rui Wang and Run Li and Tong Niu and Wenlong Zhang and Wenqi Yan and Xuanzheng Wang and Yuchen Zhang and Yi-Hsin Hung and Yuan Jiang and Zexuan Liu and Zihan Yin and Zijian Ma and Zhiwen Mo},
	year={2025},
	eprint={2506.13651},
	archivePrefix={arXiv},
	primaryClass={cs.LG}
}

@inproceedings{wu2025webwalker,
	title = "{W}eb{W}alker: Benchmarking {LLM}s in Web Traversal",
	author = "Wu, Jialong  and
	Yin, Wenbiao  and
	Jiang, Yong  and
	Wang, Zhenglin  and
	Xi, Zekun  and
	Fang, Runnan  and
	Zhang, Linhai  and
	He, Yulan  and
	Zhou, Deyu  and
	Xie, Pengjun  and
	Huang, Fei",
	booktitle = "Proceedings of the 63rd Annual Meeting of the Association for Computational Linguistics (Volume 1: Long Papers)",
	month = jul,
	year = "2025",
	pages = "10290--10305"
}

@misc{zhang2026evofsmcontrollableselfevolutiondeep,
	title={EvoFSM: Controllable Self-Evolution for Deep Research with Finite State Machines}, 
	author={Shuo Zhang and Chaofa Yuan and Ryan Guo and Xiaomin Yu and Rui Xu and Zhangquan Chen and Zinuo Li and Zhi Yang and Shuhao Guan and Zhenheng Tang and Sen Hu and Liwen Zhang and Ronghao Chen and Huacan Wang},
	year={2026},
	eprint={2601.09465},
	archivePrefix={arXiv},
	primaryClass={cs.AI}
}

@Article{Shmatko2025,
	author={Shmatko, Artem
	and Jung, Alexander Wolfgang
	and Gaurav, Kumar
	and Brunak, S{\o}ren
	and Mortensen, Laust Hvas
	and Birney, Ewan
	and Fitzgerald, Tom
	and Gerstung, Moritz},
	title={Learning the natural history of human disease with generative transformers},
	journal={Nature},
	year={2025},
	month={Nov},
	day={01},
	volume={647},
	number={8088},
	pages={248-256}
}

@misc{sharma2025researchrubricsbenchmarkpromptsrubrics,
	title={ResearchRubrics: A Benchmark of Prompts and Rubrics For Evaluating Deep Research Agents}, 
	author={Manasi Sharma and Chen Bo Calvin Zhang and Chaithanya Bandi and Clinton Wang and Ankit Aich and Huy Nghiem and Tahseen Rabbani and Ye Htet and Brian Jang and Sumana Basu and Aishwarya Balwani and Denis Peskoff and Marcos Ayestaran and Sean M. Hendryx and Brad Kenstler and Bing Liu},
	year={2025},
	eprint={2511.07685},
	archivePrefix={arXiv},
	primaryClass={cs.AI}
}

@misc{zhou2025scholarsearchbenchmarkingscholarsearching,
	title={ScholarSearch: Benchmarking Scholar Searching Ability of LLMs}, 
	author={Junting Zhou and Wang Li and Yiyan Liao and Nengyuan Zhang and Tingjia Miao and Zhihui Qi and Yuhan Wu and Tong Yang},
	year={2025},
	eprint={2506.13784},
	archivePrefix={arXiv},
	primaryClass={cs.IR}
}

@misc{xu2025researcherbenchevaluatingdeepai,
	title={ResearcherBench: Evaluating Deep AI Research Systems on the Frontiers of Scientific Inquiry}, 
	author={Tianze Xu and Pengrui Lu and Lyumanshan Ye and Xiangkun Hu and Pengfei Liu},
	year={2025},
	eprint={2507.16280},
	archivePrefix={arXiv},
	primaryClass={cs.AI}
}

@misc{patel2025deepscholarbenchlivebenchmarkautomated,
	title={DeepScholar-Bench: A Live Benchmark and Automated Evaluation for Generative Research Synthesis}, 
	author={Liana Patel and Negar Arabzadeh and Harshit Gupta and Ankita Sundar and Ion Stoica and Matei Zaharia and Carlos Guestrin},
	year={2025},
	eprint={2508.20033},
	archivePrefix={arXiv},
	primaryClass={cs.CL}
}

@misc{li2025reportbenchevaluatingdeepresearch,
	title={ReportBench: Evaluating Deep Research Agents via Academic Survey Tasks}, 
	author={Minghao Li and Ying Zeng and Zhihao Cheng and Cong Ma and Kai Jia},
	year={2025},
	eprint={2508.15804},
	archivePrefix={arXiv},
	primaryClass={cs.CL}
}

@inproceedings{
	du2025deepresearchbenchcomprehensivebenchmark,
	title={DeepResearch Bench: A Comprehensive Benchmark for Deep Research Agents},
	author={Mingxuan Du and Benfeng Xu and Chiwei Zhu and Licheng Zhang and Xiaorui Wang and Zhendong Mao},
	booktitle={The Fourteenth International Conference on Learning Representations},
	year={2026}
}

@inproceedings{
	ruan2025expertlongbenchbenchmarkinglanguagemodels,
	title={ExpertLongBench: Benchmarking Language Models on Expert-Level Long-Form Generation Tasks with Structured Checklists},
	author={Jie Ruan and Inderjeet Jayakumar Nair and Shuyang Cao and Amy Liu and Sheza Munir and Micah Pollens-Dempsey and Yune-Ting Tiffany Chiang and Lucy R. Kates and Nicholas David and Sihan Chen and Ruxin Yang and Yuqian Yang and Jihyun Jasmine Gump and Tessa Bialek and Vivek S Sankaran and Margo Schlanger and Lu Wang},
	booktitle={The Fourteenth International Conference on Learning Representations},
	year={2026},
	url={https://openreview.net/forum?id=nJvgBolRcR}
}

@misc{wan2025deepresearcharenaexamllms,
	title={DeepResearch Arena: The First Exam of LLMs' Research Abilities via Seminar-Grounded Tasks}, 
	author={Haiyuan Wan and Chen Yang and Junchi Yu and Meiqi Tu and Jiaxuan Lu and Di Yu and Jianbao Cao and Ben Gao and Jiaqing Xie and Aoran Wang and Wenlong Zhang and Philip Torr and Dongzhan Zhou},
	year={2025},
	eprint={2509.01396},
	archivePrefix={arXiv},
	primaryClass={cs.AI}
}

@misc{coelho2025deepresearchgymfreetransparentreproducible,
	title={DeepResearchGym: A Free, Transparent, and Reproducible Evaluation Sandbox for Deep Research}, 
	author={João Coelho and Jingjie Ning and Jingyuan He and Kangrui Mao and Abhijay Paladugu and Pranav Setlur and Jiahe Jin and Jamie Callan and João Magalhães and Bruno Martins and Chenyan Xiong},
	year={2025},
	eprint={2505.19253},
	archivePrefix={arXiv},
	primaryClass={cs.IR}
}

@misc{son2025aicoscientistsfailspota,
	title={When AI Co-Scientists Fail: SPOT-a Benchmark for Automated Verification of Scientific Research}, 
	author={Guijin Son and Jiwoo Hong and Honglu Fan and Heejeong Nam and Hyunwoo Ko and Seungwon Lim and Jinyeop Song and Jinha Choi and Gonçalo Paulo and Youngjae Yu and Stella Biderman},
	year={2025},
	eprint={2505.11855},
	archivePrefix={arXiv},
	primaryClass={cs.CL}
}

@misc{li2026deepresearchbenchiidiagnosing,
	title={DeepResearch Bench II: Diagnosing Deep Research Agents via Rubrics from Expert Report}, 
	author={Ruizhe Li and Mingxuan Du and Benfeng Xu and Chiwei Zhu and Xiaorui Wang and Zhendong Mao},
	year={2026},
	eprint={2601.08536},
	archivePrefix={arXiv},
	primaryClass={cs.CL}
}

@article{JMLRDLRL,
	author  = {Jiayi Weng and Huayu Chen and Dong Yan and Kaichao You and Alexis Duburcq and Minghao Zhang and Yi Su and Hang Su and Jun Zhu},
	title   = {Tianshou: A Highly Modularized Deep Reinforcement Learning Library},
	journal = {Journal of Machine Learning Research},
	year    = {2022},
	volume  = {23},
	number  = {267},
	pages   = {1--6}
}

@inproceedings{
	liang2025personalizeddeepresearchbenchmarks,
	title={Towards Personalized Deep Research: Benchmarks and Evaluations},
	author={Yuan Liang and Jiaxian Li and Yuqing Wang and WANG PIAOHONG and Motong Tian and Pai Liu and Shuofei Qiao and Runnan Fang and He Zhu and Ge Zhang and Minghao Liu and Yuchen Eleanor Jiang and Ningyu Zhang and Wangchunshu Zhou},
	booktitle={The Fourteenth International Conference on Learning Representations},
	year={2026}
}

@misc{li2025datasetresearchbenchmarkingagentsystems,
	title={DatasetResearch: Benchmarking Agent Systems for Demand-Driven Dataset Discovery}, 
	author={Keyu Li and Mohan Jiang and Dayuan Fu and Yunze Wu and Xiangkun Hu and Dequan Wang and Pengfei Liu},
	year={2025},
	eprint={2508.06960},
	archivePrefix={arXiv},
	primaryClass={cs.AI}
}

@inproceedings{
	bragg2025astabenchrigorousbenchmarkingai,
	title={AstaBench: Rigorous Benchmarking of {AI} Agents with a Scientific Research Suite},
	author={Jonathan Bragg and Mike D'Arcy and Nishant Balepur and Dan Bareket and Bhavana Dalvi Mishra and Sergey Feldman and Dany Haddad and Jena D. Hwang and Peter Jansen and Varsha Kishore and Bodhisattwa Prasad Majumder and Aakanksha Naik and Sigal Rahamimov and Kyle Richardson and Amanpreet Singh and Harshit Surana and Aryeh Tiktinsky and Rosni Vasu and Guy Wiener and Chloe Anastasiades and Stefanus Candra and Jason Dunkelberger and Daniel Emery and Rob Evans and Malachi Hamada and Regan Huff and Rodney Kinney and Matt Latzke and Jaron Lochner and Ruben Lozano-Aguilera and Ngoc-Uyen Nguyen and Smita Rao and Amber Tanaka and Brooke Vlahos and Peter Clark and Doug Downey and Yoav Goldberg and Ashish Sabharwal and Daniel S Weld},
	booktitle={The Fourteenth International Conference on Learning Representations},
	year={2026},
	url={https://openreview.net/forum?id=M7TNf5J26u}
}

@misc{zhou2025browsecompzhbenchmarkingwebbrowsing,
	title={BrowseComp-ZH: Benchmarking Web Browsing Ability of Large Language Models in Chinese}, 
	author={Peilin Zhou and Bruce Leon and Xiang Ying and Can Zhang and Yifan Shao and Qichen Ye and Dading Chong and Zhiling Jin and Chenxuan Xie and Meng Cao and Yuxin Gu and Sixin Hong and Jing Ren and Jian Chen and Chao Liu and Yining Hua},
	year={2025},
	eprint={2504.19314},
	archivePrefix={arXiv},
	primaryClass={cs.CL}
}

@inproceedings{krishna2025factfetchreasonunified,
	title = "Fact, Fetch, and Reason: A Unified Evaluation of Retrieval-Augmented Generation",
	author = "Krishna, Satyapriya  and
	Krishna, Kalpesh  and
	Mohananey, Anhad  and
	Schwarcz, Steven  and
	Stambler, Adam  and
	Upadhyay, Shyam  and
	Faruqui, Manaal",
	booktitle = "Proceedings of the 2025 Conference of the Nations of the Americas Chapter of the Association for Computational Linguistics: Human Language Technologies (Volume 1: Long Papers)",
	month = apr,
	year = "2025",
	pages = "4745--4759"
}

@inproceedings{
	pham2025sealqaraisingbarreasoning,
	title={Seal{QA}: Raising the Bar for Reasoning in Search-Augmented Language Models},
	author={Thinh Pham and Nguyen Phan Nguyen and Pratibha Zunjare and Weiyuan Chen and Yu-Min Tseng and Tu Vu},
	booktitle={The Fourteenth International Conference on Learning Representations},
	year={2026}
}

@misc{gao2026drarenaautomatedevaluationframework,
	title={DR-Arena: an Automated Evaluation Framework for Deep Research Agents}, 
	author={Yiwen Gao and Ruochen Zhao and Yang Deng and Wenxuan Zhang},
	year={2026},
	eprint={2601.10504},
	archivePrefix={arXiv},
	primaryClass={cs.CL}
}

@misc{anonymous2025livenewsbench,
	title={LiveNewsBench: Evaluating LLM Web Search Capabilities with Freshly Curated News}, 
	author={Yunfan Zhang and Kathleen McKeown and Smaranda Muresan},
	year={2026},
	eprint={2602.13543},
	archivePrefix={arXiv},
	primaryClass={cs.IR}
}

@misc{zhou2025livesearchbenchautomaticallyconstructedbenchmark,
	title={LiveSearchBench: An Automatically Constructed Benchmark for Retrieval and Reasoning over Dynamic Knowledge}, 
	author={Heng Zhou and Ao Yu and Yuchen Fan and Jianing Shi and Li Kang and Hejia Geng and Yongting Zhang and Yutao Fan and Yuhao Wu and Tiancheng He and Yiran Qin and Lei Bai and Zhenfei Yin},
	year={2025},
	eprint={2511.01409},
	archivePrefix={arXiv},
	primaryClass={cs.CL}
}

@inproceedings{rein2024gpqa,
	title={{GPQA}: A Graduate-Level Google-Proof Q\&A Benchmark},
	author={David Rein and Betty Li Hou and Asa Cooper Stickland and Jackson Petty and Richard Yuanzhe Pang and Julien Dirani and Julian Michael and Samuel R. Bowman},
	booktitle={First Conference on Language Modeling},
	year={2024}
}

@misc{lu2026statictoolstesttimetool,
	title={Beyond Static Tools: Test-Time Tool Evolution for Scientific Reasoning}, 
	author={Jiaxuan Lu and Ziyu Kong and Yemin Wang and Rong Fu and Haiyuan Wan and Cheng Yang and Wenjie Lou and Haoran Sun and Lilong Wang and Yankai Jiang and Xiaosong Wang and Xiao Sun and Dongzhan Zhou},
	year={2026},
	eprint={2601.07641},
	archivePrefix={arXiv},
	primaryClass={cs.AI}
}

@Article{Li2026,
	author={Li, Haote
	and Sarkar, Sumon
	and Lu, Wenxin
	and Loftus, Patrick O.
	and Qiu, Tianyin
	and Shee, Yu
	and Cuomo, Abbigayle E.
	and Webster, John-Paul
	and Kelly, H. Ray
	and Manee, Vidhyadhar
	and Sreekumar, Sanil
	and Buono, Frederic G.
	and Crabtree, Robert H.
	and Newhouse, Timothy R.
	and Batista, Victor S.},
	title={Collective intelligence for AI-assisted chemical synthesis},
	journal={Nature},
	year={2026},
	month={Jan},
	day={19}
}

@misc{ju2026aimathematicsprogresschallenges,
	title={AI for Mathematics: Progress, Challenges, and Prospects}, 
	author={Haocheng Ju and Bin Dong},
	year={2026},
	eprint={2601.13209},
	archivePrefix={arXiv},
	primaryClass={math.HO}
}

@misc{internagentteam2025,
	title={InternAgent: When Agent Becomes the Scientist -- Building Closed-Loop System from Hypothesis to Verification}, 
	author={Team InternAgent and Bo Zhang and Shiyang Feng and Xiangchao Yan and Jiakang Yuan and Runmin Ma and Yusong Hu and Zhiyin Yu and Xiaohan He and Songtao Huang and Shaowei Hou and Zheng Nie and Zhilong Wang and Jinyao Liu and Tianshuo Peng and Peng Ye and Dongzhan Zhou and Shufei Zhang and Xiaosong Wang and Yilan Zhang and Meng Li and Zhongying Tu and Xiangyu Yue and Wangli Ouyang and Bowen Zhou and Lei Bai},
	year={2025},
	eprint={2505.16938},
	archivePrefix={arXiv},
	primaryClass={cs.AI}
}

@misc{deep_ideation_2025,
	title={Deep Ideation: Designing LLM Agents to Generate Novel Research Ideas on Scientific Concept Network}, 
	author={Keyu Zhao and Weiquan Lin and Qirui Zheng and Fengli Xu and Yong Li},
	year={2025},
	eprint={2511.02238},
	archivePrefix={arXiv},
	primaryClass={cs.AI}
}

@inproceedings{nipsword2vec13,
	author = {Mikolov, Tomas and Sutskever, Ilya and Chen, Kai and Corrado, Greg and Dean, Jeffrey},
	title = {Distributed representations of words and phrases and their compositionality},
	year = {2013},
	booktitle = {Proceedings of the 27th International Conference on Neural Information Processing Systems - Volume 2},
	pages = {3111–3119},
	numpages = {9},
	series = {NIPS'13}
}

@book{mitchell1997machine,
	title={Machine Learning},
	author={Mitchell, T.M.},
	url={https://books.google.com/books?id=EoYBngEACAAJ},
	year={1997},
	publisher={McGraw-Hill}
}

%\appendix
%
%\section{Example Appendix}
%\label{sec:appendix}
%
%This is an appendix.
%
%\textbf{Responsible NLP Checklist}

\end{document}